\documentclass[final,onefignum,onetabnum,letterpaper]{siamonline190516}
\usepackage[scale=0.8]{geometry} 

\usepackage{verbatim}
\usepackage{lipsum}
\usepackage{amsfonts}
\usepackage{epstopdf}
\ifpdf
  \DeclareGraphicsExtensions{.eps,.pdf,.png,.jpg}
\else
  \DeclareGraphicsExtensions{.eps}
\fi

\usepackage{datetime}
\newdateformat{monthyeardate}{%
	\monthname[\THEMONTH] \THEDAY, \THEYEAR}

\usepackage{academicons}
\usepackage{xcolor}
\renewcommand{\orcid}[1]{\href{https://orcid.org/#1}{\textcolor[HTML]{A6CE39}{orcid.org/#1}}}

\usepackage{amsmath}
\allowdisplaybreaks
\usepackage{amssymb}
\usepackage{commath}
\usepackage{mathtools}
\usepackage{bbm}
\usepackage{bm}

\usepackage{color}
\usepackage{graphicx}
\usepackage[small]{caption}
\usepackage{subcaption}

\usepackage{relsize}
\usepackage{adjustbox}
\usepackage{algorithmic}
\usepackage{booktabs}
\usepackage{tikz}
\usepackage{pifont}
\usetikzlibrary{bayesnet} 

\usepackage{enumitem}
\setlist[enumerate]{leftmargin=.5in}
\setlist[itemize]{leftmargin=.5in}

\newsiamremark{remark}{Remark}
\newsiamremark{example}{Example}
\newsiamremark{algo}{Algorithm}
\newsiamremark{hypothesis}{Hypothesis}
\crefname{hypothesis}{Hypothesis}{Hypotheses}
\newsiamthm{claim}{Claim}

\headers{Joint hierarchical Bayesian learning (JHBL)}{Y.\ Xiao and J.\ Glaubitz}

\title{Sequential image recovery using joint hierarchical Bayesian learning
\thanks{\monthyeardate\today 
\correspondings{Jan Glaubitz} 
}}

\author{ 
Yao Xiao\thanks{Department of Mathematics, Dartmouth College, Hanover, NH 03755, USA (\email{yao.xiao.gr.dartmouth@gmail.com}, \orcid{0000-0002-0850-9624})}
\and 
Jan Glaubitz\thanks{Department of Aeronautics and Astronautics, MIT, Cambridge, MA 02139, USA (\email{glaubitz@mit.edu}, \orcid{0000-0002-3434-5563})}
}

\usepackage{amsopn}

\usepackage[normalem]{ulem}

\DeclareMathOperator{\diag}{diag}

\DeclareMathOperator*{\argmax}{arg\,max}

\newcommand{\intd}{\, \mathrm{d}}

\newcommand{\R}{\mathbb{R}}

\usepackage{lineno}

\begin{document}

\maketitle

\begin{abstract}
Recovering temporal image sequences (videos) based on indirect, noisy, or incomplete data is an essential yet challenging task. 
We specifically consider the case where each data set is missing vital information, which prevents the accurate recovery of the individual images. 
Although some recent (variational) methods have demonstrated high-resolution image recovery based on jointly recovering sequential images, there remain robustness issues due to parameter tuning and restrictions on the type of sequential images. 
Here, we present a method based on hierarchical Bayesian learning for the joint recovery of sequential images that incorporates prior intra- and inter-image information. 
Our method restores the missing information in each image by ``borrowing" it from the other images. 
More precisely, we couple sequential images by penalizing their pixel-wise difference. 
The corresponding penalty terms (one for each pixel and pair of subsequent images) are treated as weakly-informative random variables that favor small pixel-wise differences but allow occasional outliers. 
As a result, \emph{all} of the individual reconstructions yield improved accuracy.
Our method can be used for various data acquisitions and allows for uncertainty quantification. 
Some preliminary results indicate its potential use for sequential deblurring and magnetic resonance imaging. 
\end{abstract}



\begin{DOI}
    \url{https://doi.org/10.1007/s10915-023-02234-1}
\end{DOI}

\section{Introduction} 
\label{sec:introduction} 

Many applications rely on recovering temporal image sequences from noisy, indirect, and incomplete data \cite{hu2007semantic,benfold2011stable,yang2020ground,xiao2022sequential}. 
We can often formulate this task as a sequence of \emph{linear inverse problems}, 
\begin{equation}\label{eq:data_model}
    \mathbf{y}^{(j)} = F^{(j)} \mathbf{x}^{(j)} + \mathbf{e}^{(j)}, 
    \quad j=1,\dots,J,
\end{equation}
where $\mathbf{y}^{(j)}$ is a given data vector, $\mathbf{x}^{(j)}$ is the (unknown) vectorized image, $F^{(j)}$ is a known linear forward operator, and $\mathbf{e}^{(j)}$ corresponds to unknown noise. 
The individual recovery of each image by separately solving the linear inverse problems \eqref{eq:data_model} is a well-studied, although challenging problem by itself \cite{groetsch1993inverse,vogel2002computational,hansen2010discrete}. 
A prominent approach is to replace \eqref{eq:data_model} with a nearby regularized inverse problem that promotes some prior belief about the unknown image. 
In imaging applications, it is often reasonable to assume that some linear transform of the unknown image $\mathbf{x}$, say $R \mathbf{x}$, is sparse. 
This prior belief yields the \emph{$\ell^1$-regularized inverse problems} 
\begin{equation}\label{eq:l1_RIP}
	\min_{\mathbf{x}^{(j)}} \left\{ \| F^{(j)} \mathbf{x}^{(j)} - \mathbf{y}^{(j)} \|_2^2 + \lambda_j \| R \mathbf{x}^{(j)} \|_1 \right\}, \quad j=1,\dots,J,
\end{equation}
where $R$ is the regularization operator and $\lambda_j > 0$ are regularization parameters. 
Usual choices for $R$ are discrete (high-order) total variation (TV) \cite{rudin1992nonlinear}, total generalized/directional variation \cite{bredies2010total}/\cite{parisotto2020higher,parisotto2020higher2}, and polynomial annihilation \cite{archibald2005polynomial,archibald2016image,glaubitz2019high} operators. 
The rationale behind considering \eqref{eq:l1_RIP} is that the $\ell^1$-norm, $\|\cdot||_1$, serves as a convex surrogate for the $\ell^0$-``norm", $\|\cdot\|_0$; an observation that lies in the heart of compressed sensing \cite{donoho2006compressed,eldar2012compressed,foucart2017mathematical}. 
Another prominent approach, closely related to regularization, is Bayesian inverse problems \cite{kaipio2006statistical,calvetti2007introduction,stuart2010inverse}. 
In this setting, we express our lack of information about some of the quantities in \eqref{eq:data_model} by modeling them as random variables, which (relations) are characterized by certain density functions. 
The fidelity and regularization term corresponds to the negative logarithm of the likelihood and prior density, respectively, and the regularized solution \eqref{eq:l1_RIP} corresponds to the maximizer of the posterior density. 
The class of conditionally Gaussian priors \cite{tipping2001sparse,calvetti2007gaussian,calvetti2008hypermodels,calvetti2019hierachical,calvetti2020sparsity} is particularly suited to promote sparsity. 

\begin{figure}[tb]
    \centering
    \begin{subfigure}[b]{.325\textwidth}
    \includegraphics[width=\textwidth]{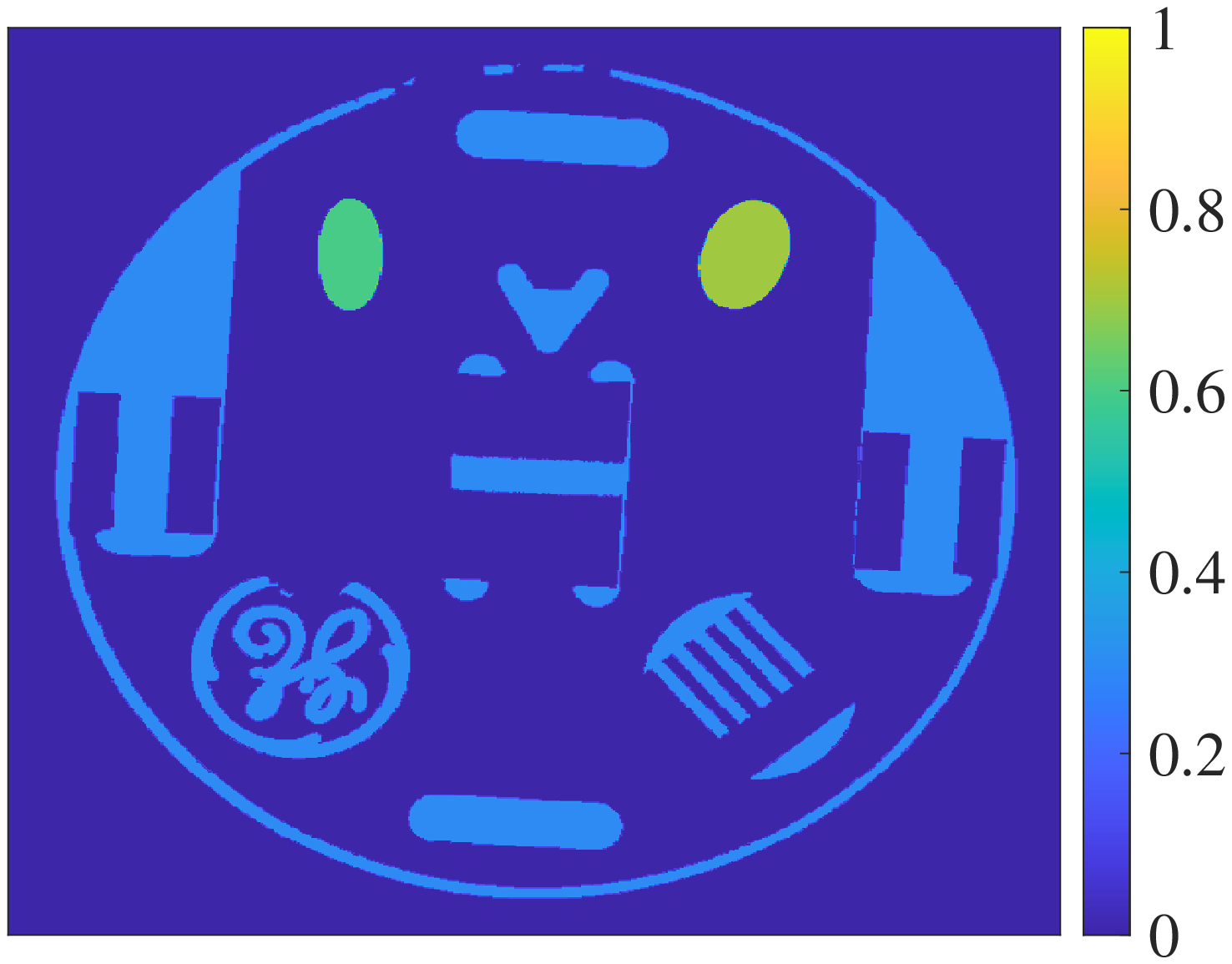}
    \caption{Reference image 1}
    \end{subfigure}
    \begin{subfigure}[b]{.325\textwidth}
    \includegraphics[width=\textwidth]{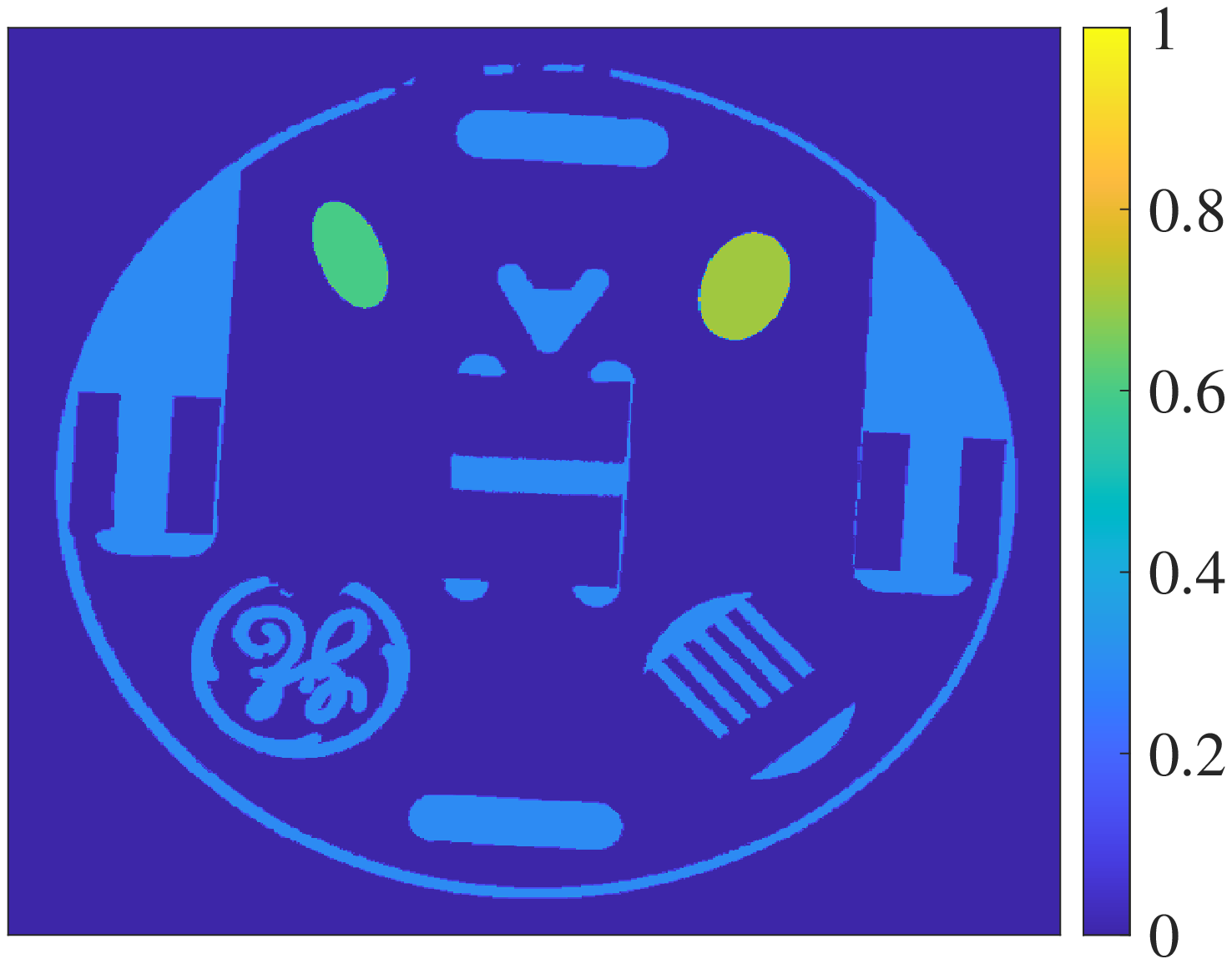}
    \caption{Reference image 2}
    \end{subfigure}
    \begin{subfigure}[b]{.325\textwidth}
    \includegraphics[width=\textwidth]{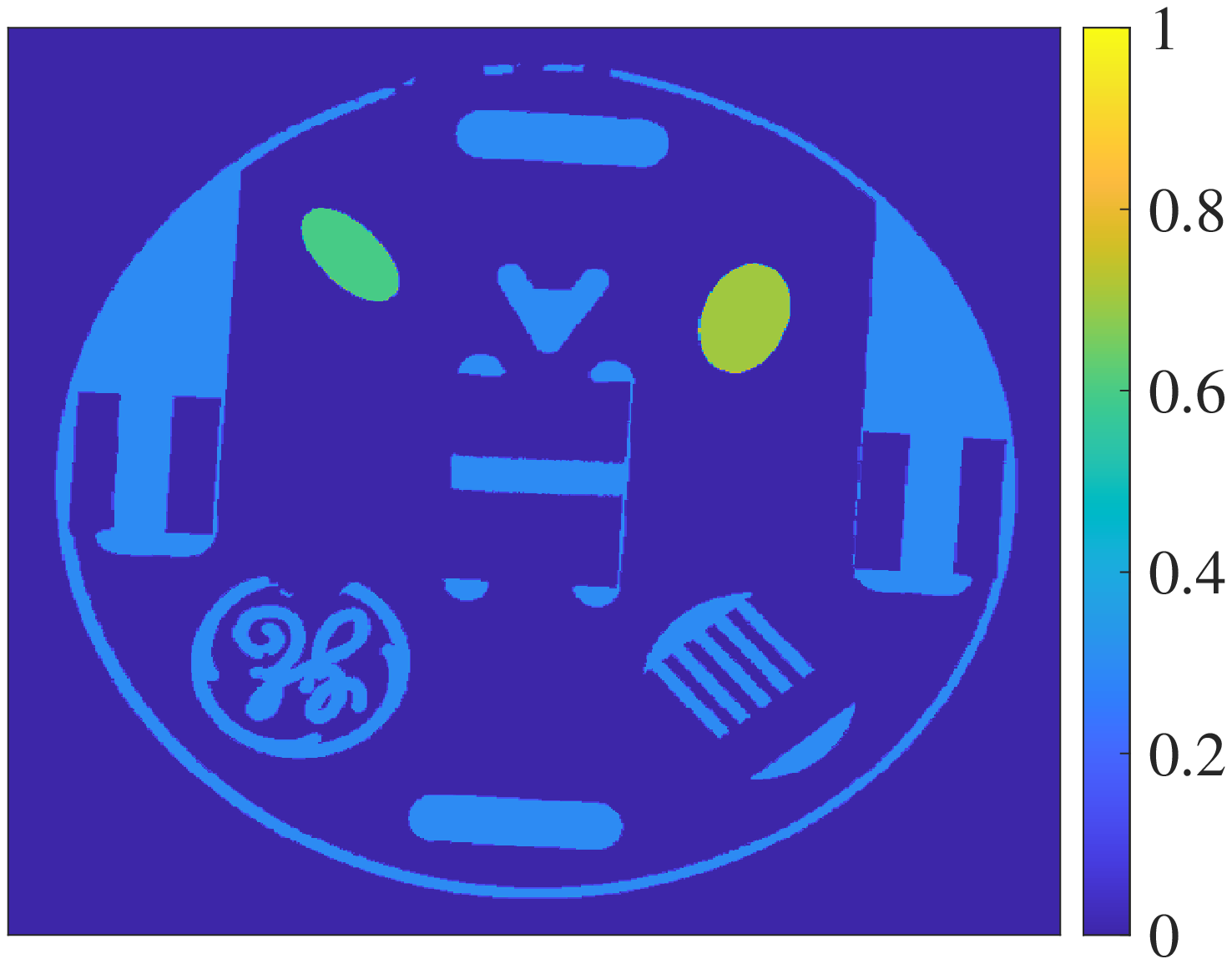}
    \caption{Reference image 3}
    \end{subfigure}
    \\
    \begin{subfigure}[b]{.325\textwidth}
    \includegraphics[width=\textwidth]{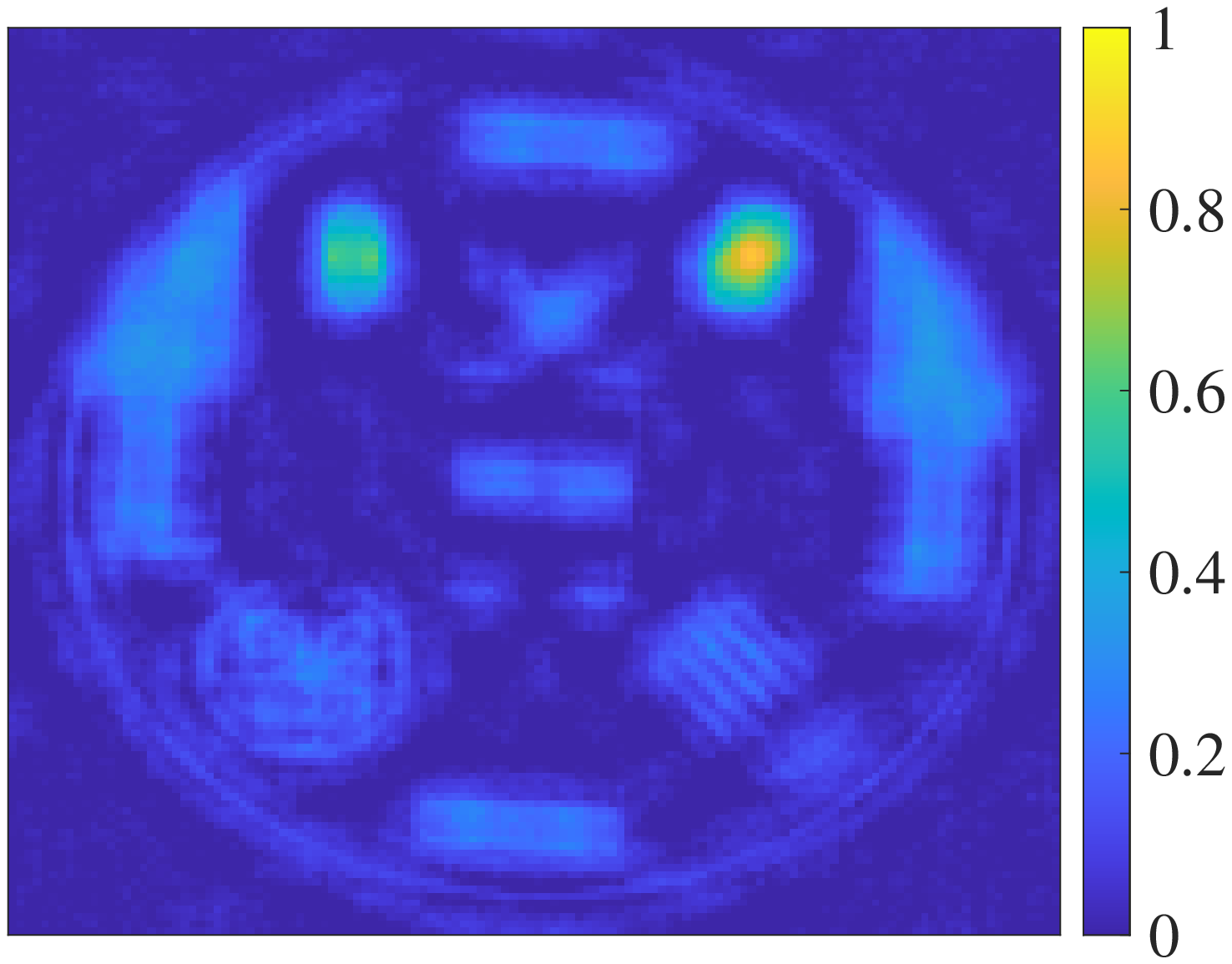}
    \caption{Sep.\ recovered image 1}
    \end{subfigure} 
    \begin{subfigure}[b]{.325\textwidth}
    \includegraphics[width=\textwidth]{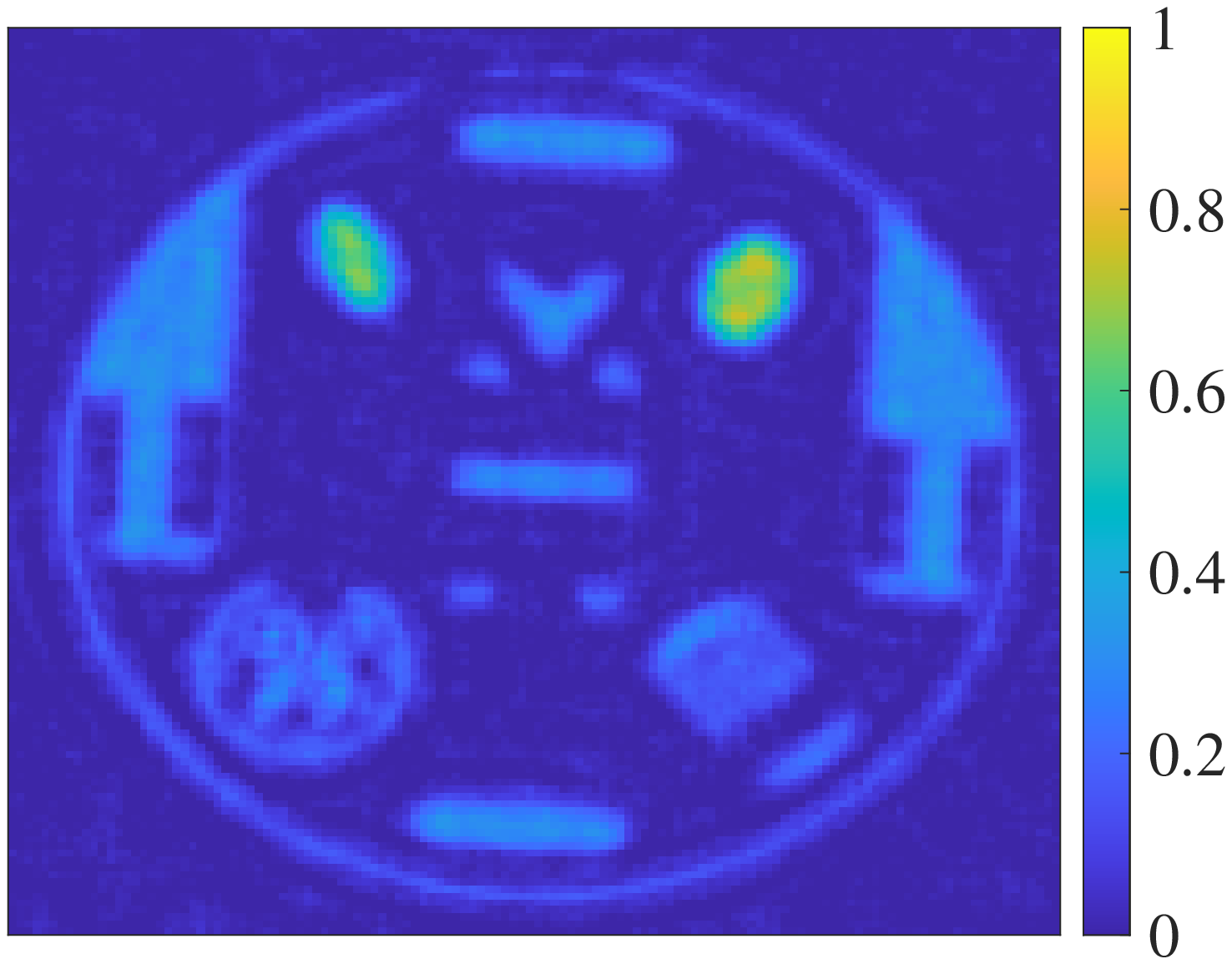}
    \caption{Sep.\ recovered image 2}
    \end{subfigure} 
    \begin{subfigure}[b]{.325\textwidth}
    \includegraphics[width=\textwidth]{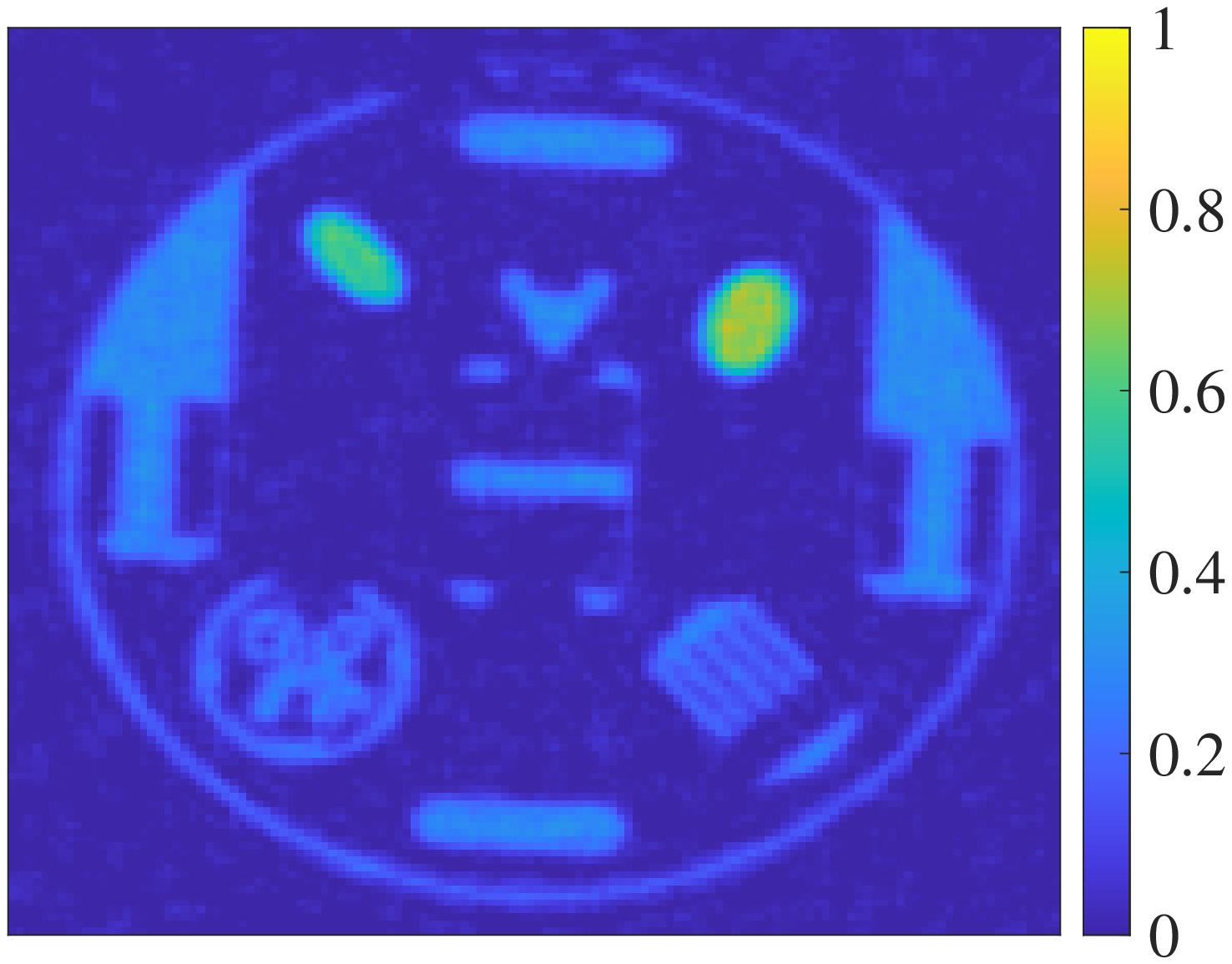}
    \caption{Sep.\ recovered image 3}
    \end{subfigure} 
    \\
    \begin{subfigure}[b]{.325\textwidth}
    \includegraphics[width=\textwidth]{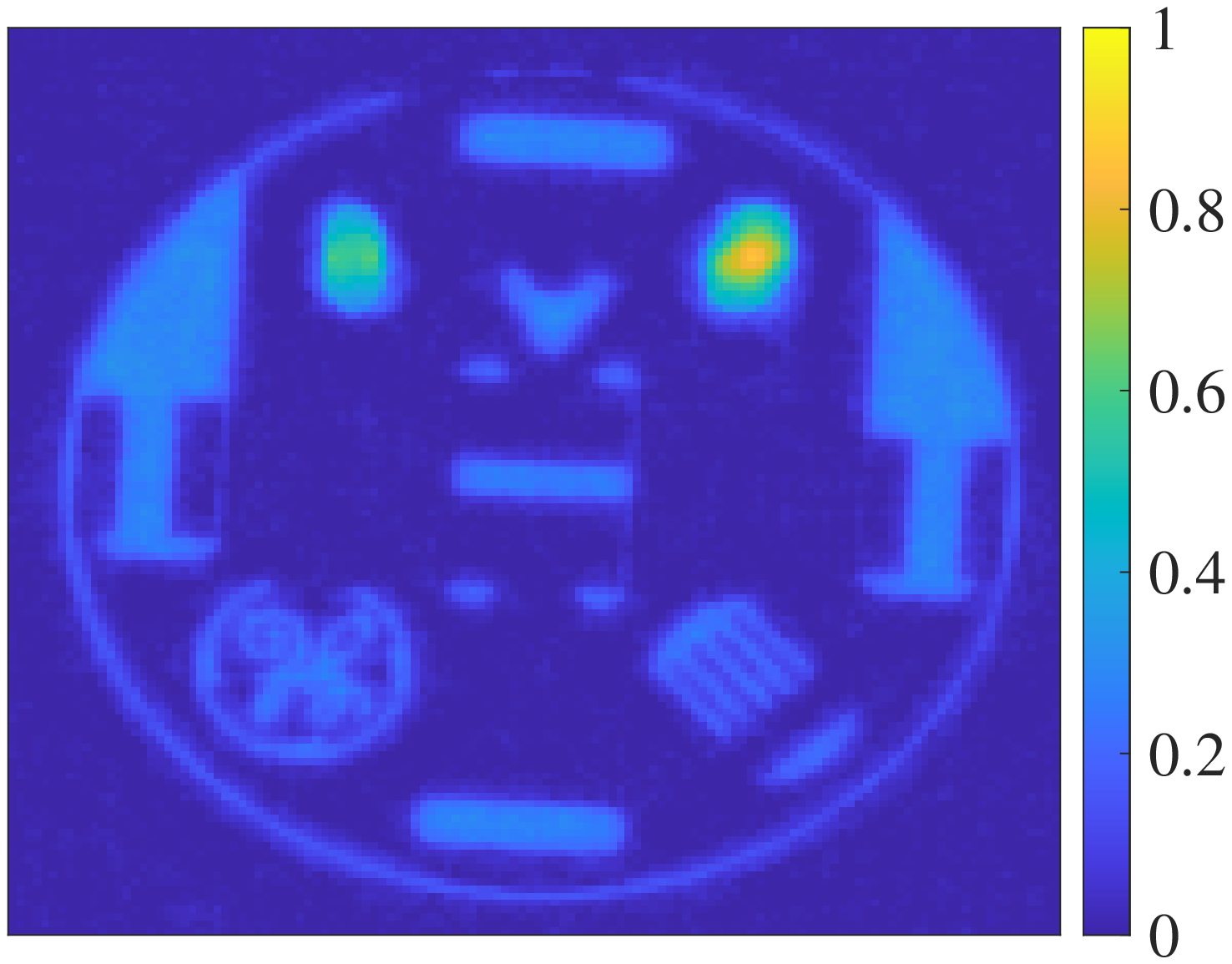}
    \caption{Jointly recovered image 1}
    \end{subfigure}
    \begin{subfigure}[b]{.325\textwidth}
    \includegraphics[width=\textwidth]{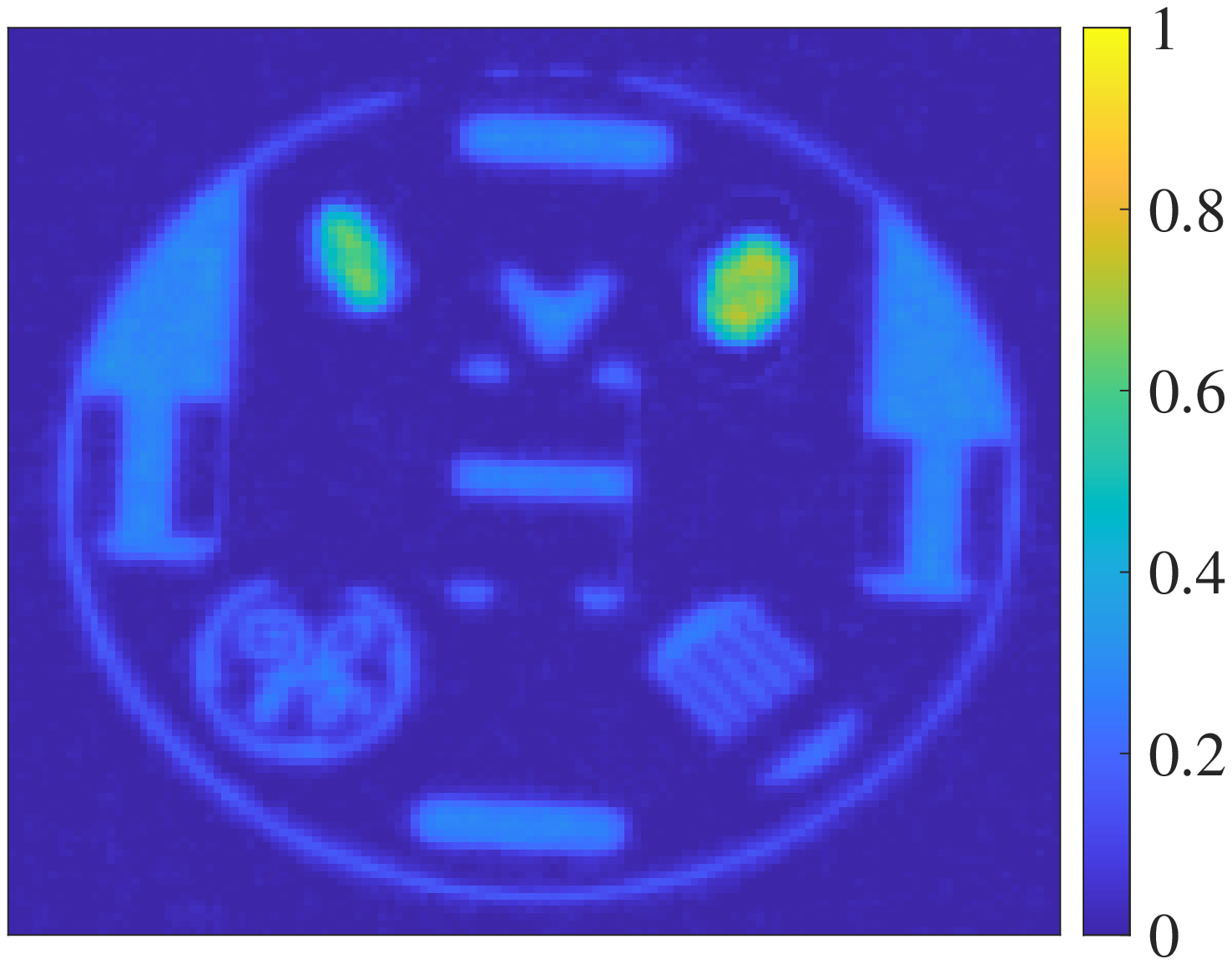}
    \caption{Jointly recovered image 2}
    \end{subfigure}
    \begin{subfigure}[b]{.325\textwidth}
    \includegraphics[width=\textwidth]{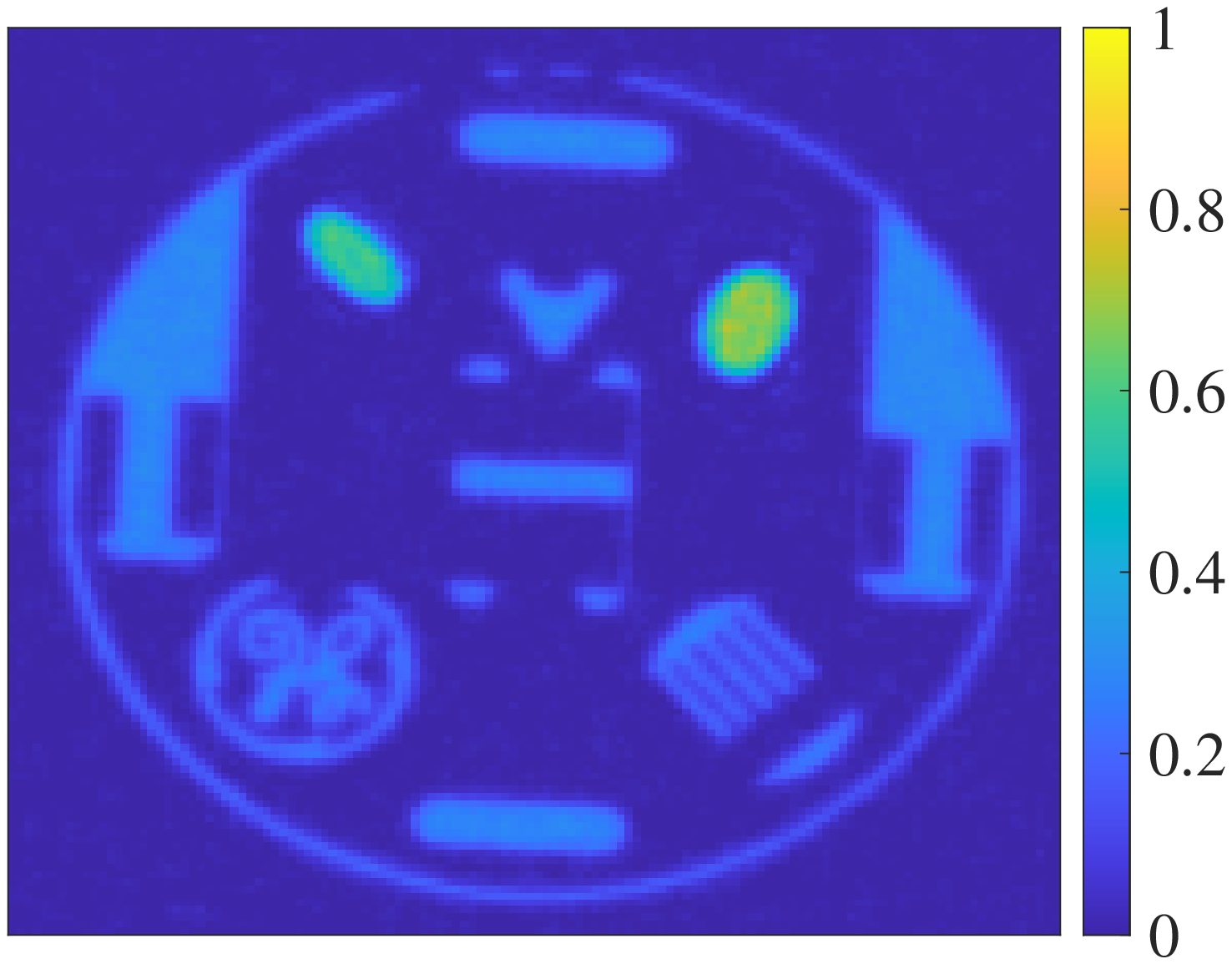}
    \caption{Jointly recovered image 3}
    \end{subfigure}
    \caption{
    Three images of a temporal magnetic resonance image sequence with a rotating left (green) and a down-moving right (yellow) ellipse. 
    The first row shows the reference images, the second row the separately recovered images using \eqref{eq:l1_RIP}, and the third row the jointly recovered images using \eqref{eq:joint_RIP}. 
    The data sets are noisy and miss different Fourier samples. 
    }
    \label{fig:intro}
\end{figure}

Here, we consider the case that each data set $\mathbf{y}^{(j)}$ in \eqref{eq:data_model} is missing vital information, which prevents the accurate individual recovery of the images $\mathbf{x}^{(j)}$. 
A common strategy is to restore the missing information in each image by ``borrowing" it from the other images. 
\cite{cotter2005sparse,ehrhardt2014joint,adcock2019joint} and \cite{wipf2007empirical,zhang2011sparse} considered deterministic and Bayesian methods of such a flavor that rely on a \emph{common sparsity assumption}, meaning that $R \mathbf{x}^{(1)},\dots,R \mathbf{x}^{(J)}$ have the same support. 
Unfortunately, temporally changing image sequences often violate the common sparsity assumption.
Recently, \cite{xiao2022sequential} addressed this problem by locally coupling the images in no-change regions while decoupling them in change regions. 
This was done by first computing a diagonal change mask $C^{(j-1,j)}$ directly from the consecutive data sets $\mathbf{y}^{(j-1)}, \mathbf{y}^{(j)}$ and using this change mask to penalize any difference between $\mathbf{x}^{(j-1)}$ and $\mathbf{x}^{(j)}$ in no-change regions. 
Let $[C^{(j-1,j)}]_{n,n} = 0$ in change regions and $[C^{(j-1,j)}]_{n,n} = 1$ in no-change regions. 
The \emph{joint $\ell^1$-regularized inverse problem} used in \cite{xiao2022sequential} is  
\begin{equation}\label{eq:joint_RIP} 
\resizebox{.92\textwidth}{!}{$\displaystyle 
	\min_{\mathbf{x}^{(1)},\dots,\mathbf{x}^{(J)}} \left\{ \sum_{j=1}^J \| F^{(j)} \mathbf{x}^{(j)} - \mathbf{y}^{(j)} \|_2^2 + \sum_{j=1}^J \lambda_j \| R \mathbf{x}^{(j)} \|_1 + \sum_{j=2}^J \mu_j \| C^{(j-1,j)} ( \mathbf{x}^{(j-1)} - \mathbf{x}^{(j)} ) \|_2^2 \right\}, 
$}
\end{equation}
where the $\mu_j \geq 0$ are fixed coupling parameters. 
\eqref{eq:joint_RIP} balances data fidelity, intra-image regularization (sparsity of $R \mathbf{x}^{(j)}$), and inter-image regularization (coupling in no-change regions). 
Roughly speaking, the last term in \eqref{eq:joint_RIP},
\begin{equation} 
	\| C^{(j-1,j)} ( \mathbf{x}^{(j-1)} - \mathbf{x}^{(j)} ) \|_2^2 
		= \sum_{n=1}^N [C^{(j-1,j)}]_{n,n} \left| x^{(j-1)}_n - x^{(j)}_n \right|^2,
\end{equation}
penalizes any change between the two subsequent images $\mathbf{x}^{(j-1)}$ and $\mathbf{x}^{(j)}$ in no-change regions, where $[C^{(j-1,j)}]_{n,n} = 1$, since the value of the objective function in \eqref{eq:joint_RIP} increases with $|x^{(j-1)}_n - x^{(j)}_n|$ in this case. 
At the same time, in change regions, the difference $|x^{(j-1)}_n - x^{(j)}_n|$ does not influence the value of the objective function since $[C^{(j-1,j)}]_{n,n} = 0$ in this case. 
Figure \ref{fig:intro} illustrates the advantage of jointly recovering a temporal sequence of magnetic resonance images from noisy and under-sampled Fourier data (see \S\ref{sec:tests} for more details).

While the joint recovery of changing images using \eqref{eq:joint_RIP} can yield an improved accuracy, there remain issues with robustness and the range of application: 
(I1) Selecting appropriate regularization and coupling parameters is a non-trivial task, and their choice can critically influence the quality of the recovered images. 
Although (re-)weighted \cite{candes2008enhancing,adcock2019joint,gelb2019reducing} $\ell^1$-regularization can increase the robustness w.\,r.\,t.\ the intra-image regularization parameters $\lambda_j$, selecting suitable inter-image coupling parameters $\mu_j$ remains an open problem. 
(I2) The method proposed in \cite[Section 3]{xiao2022sequential} to pre-compute the change masks uses Fourier data and assumes that the sequential images only include objects with closed boundaries. 
This assumption prevents the application to problems with other types of data acquisition. 
Finally, one has to tune some problem-dependent free parameters by hand on a case-by-case basis.

\subsection*{Our contribution} 

We propose a \emph{joint hierarchical Bayesian learning (JHBL)} method for sequential image recovery. 
The method is easy to implement, efficient, and simple to parallelize. 
Our method avoids issues (I1) and (I2) by reinterpreting all involved parameters---including the change mask---as random variables, which we then estimate together with the recovered images. 
In particular, for the random variables responsible for the inter-image coupling, we use weakly-informative gamma distributions that favor small pixel-wise differences (in no-change regions) but allow for occasional outliers (in change regions). 
Our approach does not rely on Fourier data or images showing objects with closed boundaries. 
We demonstrate that our JHBL method improves the accuracy of all individual images. 
Another advantage is that our method quantifies the uncertainty in the recovered images, which is often desirable. 
Some preliminary results for magnetic resonance imaging and road-traffic monitoring indicate the advantage of JHBL for sequential image recovery.

\subsection*{Outline} 

In \S\ref{sec:model}, we present the JHBL model that promotes intra-image sparsity and inter-image coupling.  
In \S\ref{sec:inference}, we propose an efficient method for Bayesian inference.  
In \S\ref{sec:tests}, we demonstrate the performance of the resulting JHBL method for test cases from sequential deblurring and magnetic resonance imaging. 
\S\ref{sec:summary} offers some concluding thoughts.  
\section{The joint hierarchical Bayesian model} 
\label{sec:model} 

We now describe the hierarchical Bayesian model we use to develop our JHBL method.

\subsection{The likelihood} 
\label{sub:likelihood}

\begin{figure}[tb]
    \centering
    \begin{subfigure}[b]{.45\textwidth}
    \includegraphics[width=\textwidth]{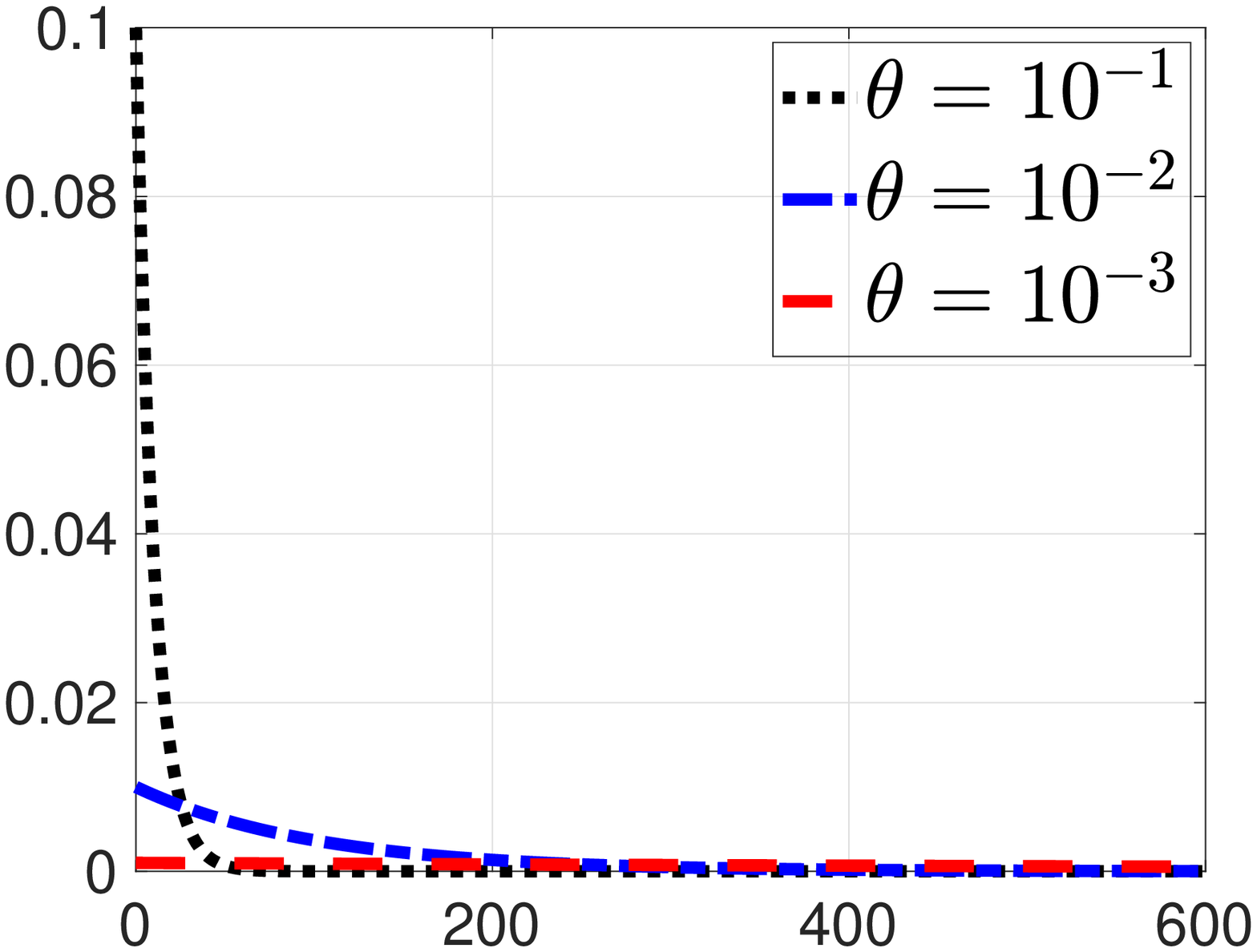}
    \caption{Shape parameter $\eta=1$}
    \label{fig:gamma_eta1}
    \end{subfigure}
    \begin{subfigure}[b]{.45\textwidth}
    \includegraphics[width=\textwidth]{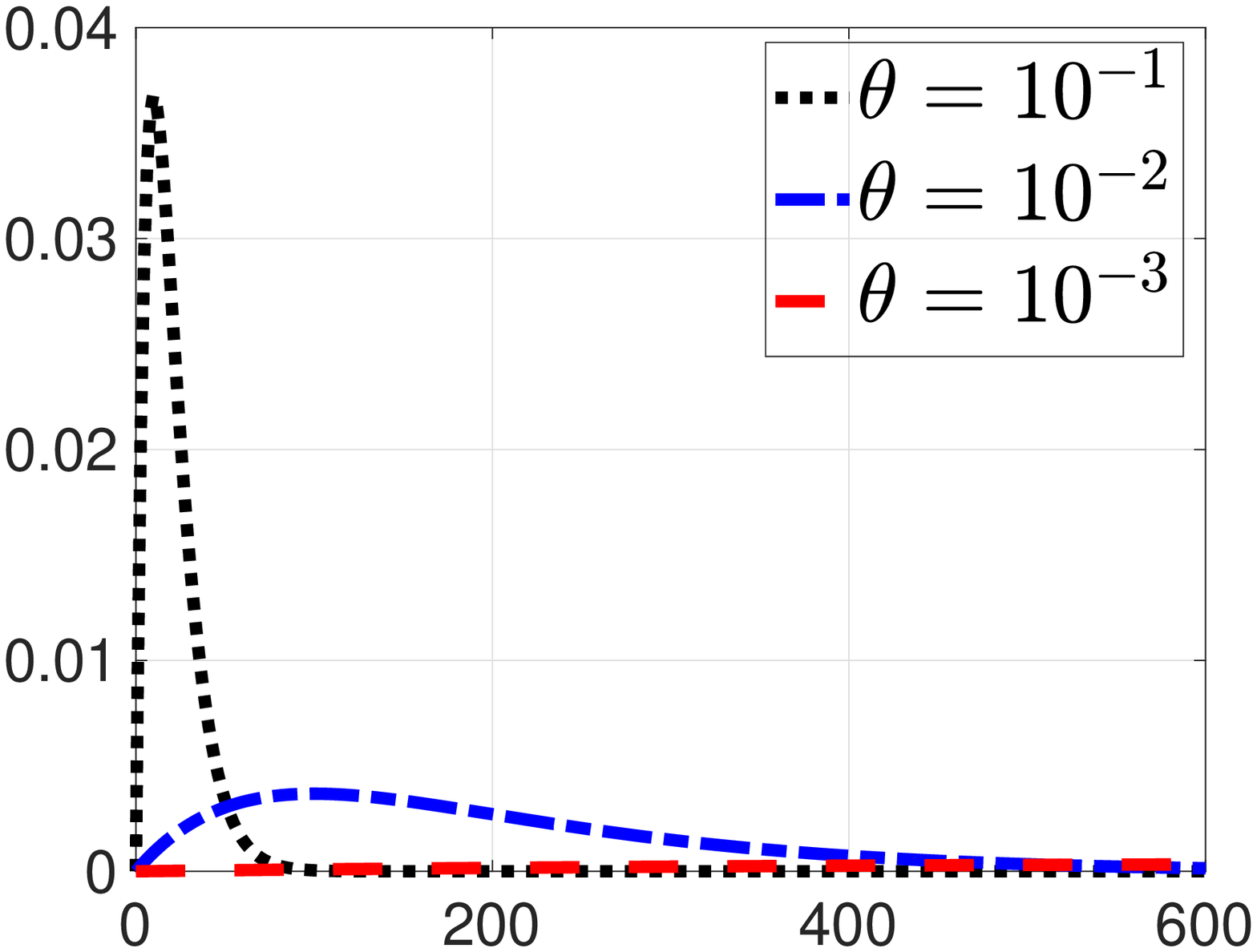}
    \caption{Shape parameter $\eta=2$} 
    \label{fig:gamma_eta2}
    \end{subfigure}
    \caption{
    Gamma density functions for different shape and rate parameters $\eta, \theta$. 
    Recall that the mode, expected value, and variance of a gamma distribution $\Gamma(\eta,\theta)$ are $(\eta-1)/\theta$, $\eta/\theta$, and $\eta/\theta^2$, respectively.
    In particular, decreasing the rate $\theta$ makes the density flatter, while $\eta$ influences its peak. 
    }
    \label{fig:gamma}
\end{figure}

Consider the data model \eqref{eq:data_model} and assume that $\mathbf{y}^{(j)} \in \R^{M_j}$, $F \in \R^{M_j \times N}$, $\mathbf{x}^{(j)} \in \R^N$, and that $\mathbf{e}^{(j)} \in \R^{M_j}$ is independent and identically distributed (i.i.d.) zero-mean normal noise, $e_m^{(j)} \sim \mathcal{N}(0,\alpha_j)$ for $m=1,\dots,M_j$, with noise precision $\alpha_j > 0$.\footnote{%
We can also use \eqref{eq:data_model} for complex (Fourier) data. 
}  
The $j$th likelihood function, which is the conditional probability density of $\mathbf{y}^{(j)}$ given $\mathbf{x}^{(j)}$ and $\alpha_j$, is 
\begin{equation}\label{eq:likelihood}
    p(\mathbf{y}^{(j)} | \mathbf{x}^{(j)}, \alpha_j) 
        \propto \alpha_j^{M_j/2} \exp\left\{ -\frac{\alpha_j}{2} \| F^{(j)} \mathbf{x}^{(j)} - \mathbf{y}^{(j)} \|_2^2 \right\}, 
\end{equation}
where ''$\propto$" means that the two sides are equal up to a multiplicative constant. 
We also treat the noise precision $\alpha_j$ as a random variable that is learned together with the images $\mathbf{x}^{(j)}$.\footnote{Even when the exact noise precision is known, using it as a fixed value for $\alpha_j$ can yield suboptimal reconstructions \cite{zhang2011clarify}.}
We assume that $\alpha_j$ is gamma distributed, 
\begin{equation}\label{eq:hyperprior_alpha}
    p(\alpha_j) 
        = \Gamma(\alpha_j|\eta_{\alpha_j},\theta_{\alpha_j}) 
        \propto \alpha_j^{\eta_{\alpha_j}-1} \exp\{-\theta_{\alpha_j} \alpha_j\}, 
        \quad j=1,\dots,J.
\end{equation} 
For simplicity, we use the same shape and rate parameter for all noise precisions, denoted by $\eta_{\alpha}$ and $\theta_{\alpha}$. 
We use the gamma distribution because it is conditionally conjugate to the normal distribution, which is convenient for Bayesian inference (see \S\ref{sec:inference}). 
Moreover, we can make the noise hyper-priors \eqref{eq:hyperprior_alpha} flat and therefore uninformative by choosing $\theta_{\alpha} \approx 0$. 
Figure \ref{fig:gamma} illustrates the gamma density function for different shape and rate parameters. 
Remark \ref{rem:informative_likelihood} addresses the possibility of using informative hyper-priors.

If we denote the collection of all images by $\mathbf{x} = [\mathbf{x}^{(1)};\dots;\mathbf{x}^{(J)}]$, of all data sets by $\mathbf{y} = [\mathbf{y}^{(1)};\dots;\mathbf{y}^{(J)}]$, and of all noise precisions by $\boldsymbol{\alpha} = [\alpha_1;\dots;\alpha_J]$, then the \emph{joint likelihood function} is 
\begin{equation}\label{eq:likelihood_joint}
    p( \mathbf{y} | \mathbf{x}, \boldsymbol{\alpha} ) 
        = \prod_{j=1}^J p(\mathbf{y}^{(j)} | \mathbf{x}^{(j)}, \alpha_j) 
        \propto \prod_{j=1}^J \alpha_j^{M_j/2} \exp\left\{ -\frac{\alpha_j}{2} \| F^{(j)} \mathbf{x}^{(j)} - \mathbf{y}^{(j)} \|_2^2 \right\}, 
\end{equation}
assuming that the data sets are conditionally independent. 
A few remarks are in order. 

\begin{remark}\label{rem:informative_likelihood}
    For simplicity we use the same hyper-prior $\Gamma(\cdot|\eta_{\alpha},\theta_{\alpha})$ and parameters $\eta_{\alpha},\theta_{\alpha}$ for all components of $\boldsymbol{\alpha}$. 
    We do this since we assume no prior knowledge about the noise variances of the different measurement vectors. 
    However, if one has a reasonable a priori notion of the underlying noise variances, the choice for hyper-prior could be modified accordingly \cite{bardsley2012mcmc,calvetti2019hierachical}. 
\end{remark} 

\begin{remark}\label{rem:indep_likelihood}
    The data sets being conditionally independent means that if we know the images $\mathbf{x}$ and the noise precisions $\boldsymbol{\alpha}$, then knowledge of one data set $\mathbf{y}^{(j)}$ provides no additional information about the likelihood of another data set $\mathbf{y}^{(i)}$ with $i \neq j$. 
\end{remark}

\subsection{The intra-image prior} 
\label{sub:prior_intra}

We assume that some linear transform of the images, say $R \mathbf{x}^{(j)}$ with $R \in \R^{K \times N}$, is sparse. 
One can model sparsity by various priors, including TV  \cite{kaipio2000statistical,babacan2008parameter}, mixture-of-Gaussian  \cite{fergus2006removing}, Laplace \cite{figueiredo2007majorization}, and hyper-Laplace \cite{levin2007image,krishnan2009fast} priors. 
Here, we use conditionally Gaussian priors, which are particularly suited to promote sparsity and allow for efficient Bayesian inference \cite{calvetti2019hierachical,calvetti2020sparsity,glaubitz2022generalized}. 
The $j$th intra-image prior is 
\begin{equation}\label{eq:intra_prior}
    p( \mathbf{x}^{(j)} | \boldsymbol{\beta}^{(j)} ) 
        \propto \det\left( B^{(j)} \right)^{1/2} \exp\left\{ - \frac{1}{2} ( \mathbf{x}^{(j)} )^T R^T B^{(j)} R \mathbf{x}^{(j)} \right\}, 
\end{equation}
where $B^{(j)} = \diag(\boldsymbol{\beta}^{(j)})$ and $\boldsymbol{\beta}^{(j)} = [\beta^{(j)}_1,\dots,\beta^{(j)}_K]$ is treated as a random vector with i.\,i.\,d.\ gamma distributed components, 
\begin{equation}\label{eq:hyperprior_beta}
    p(\beta^{(j)}_k) 
        = \Gamma(\beta^{(j)}_k|\eta_{\beta^{(j)}_k},\theta_{\beta^{(j)}_k}), \quad k=1,\dots,K.
\end{equation} 
For simplicity, we use the same shape and rate parameter for all prior precisions, denoted by $\eta_{\beta}$ and $\theta_{\beta}$. 
We choose $\theta_{\beta} \approx 0$ to make the hyper-prior flat and thus uninformative. 
Again, if one has a reasonable a priori notion of the support of $R \mathbf{x}^{(j)}$, the choice for the hyper-prior \eqref{eq:hyperprior_beta} and the parameters $\eta_{\beta^{(j)}_k},\theta_{\beta^{(j)}_k}$ could be modified correspondingly. 
If we denote the collection of all precision vectors by $\boldsymbol{\beta} = [\boldsymbol{\beta}^{(1)};\dots;\boldsymbol{\beta}^{(J)}]$, then the \emph{joint intra-image prior} is
\begin{equation}\label{eq:intra_prior_joint} 
\begin{aligned}
    p( \mathbf{x} | \boldsymbol{\beta} ) 
        = \prod_{j=1}^J p( \mathbf{x}^{(j)} | \boldsymbol{\beta}^{(j)} )  
        \propto \prod_{j=1}^J \det\left( B^{(j)} \right)^{1/2} \exp\left\{ - \frac{1}{2}( \mathbf{x}^{(j)} )^T R^T B^{(j)} R \mathbf{x}^{(j)} \right\}, 
\end{aligned}
\end{equation}
assuming that the images are conditionally independent.

\begin{remark}\label{rem:indep_intra}
    The images being conditionally independent means that if we know the parameters $\boldsymbol{\beta}$, then knowledge of one image $\mathbf{x}^{(j)}$ provides no additional information about the likelihood of another image $\mathbf{x}^{(i)}$ with $i \neq j$. 
    Although we assume that the images form a temporal sequence with only parts of the images changing, we treat this as qualitative information; 
    We neither want to quantify the exact location nor the size of the (no-)change regions, making us assume that the images are conditionally independent. 
\end{remark}

\subsection{The inter-image prior} 
\label{sub:prior_inter}

Assume for the moment that we had a pre-computed diagonal change mask $C^{(j-1,j)}$ with $[C^{(j-1,j)}]_{n,n} = 0$ in change regions and $[C^{(j-1,j)}]_{n,n} = 1$ in no-change regions\footnote{%
For instance, Figures \ref{fig:MRI_change_det1} and \ref{fig:MRI_change_det2} in \S\ref{sub:tests_MRI} illustrate the pre-computed binary change mask used to jointly recover the images in Figure \ref{fig:intro}. 
} 
as in \cite{xiao2022sequential}.
We could then translate the coupling term in \eqref{eq:joint_RIP} into the empirical conditionally Gaussian prior 
\begin{equation}\label{eq:inter_prior_direct} 
\begin{aligned}
    p( \mathbf{x} | \mu_2,\dots,\mu_J ) 
        & \propto \prod_{j=2}^J \mu_j^{N/2} \exp\left\{ - \frac{\mu_j}{2} \| C^{(j-1,j)} ( \mathbf{x}^{(j-1)} - \mathbf{x}^{(j)} ) \|_2^2 \right\} \\ 
        & = \prod_{j=2}^J \mu_j^{N/2} \exp\left\{ - \frac{\mu_j}{2} ( \mathbf{x}^{(j-1)} - \mathbf{x}^{(j)} )^T C^{(j-1,j)} ( \mathbf{x}^{(j-1)} - \mathbf{x}^{(j)} ) \right\},
\end{aligned}
\end{equation}
where we have used that $(C^{(j-1,j)})^2 = C^{(j-1,j)}$.  
As mentioned before, the method proposed in \cite[Section 3]{xiao2022sequential} to pre-compute the change masks uses Fourier data and assumes that the sequential images only include objects with closed boundaries, however. 
We overcome these restrictions by replacing the pre-computed diagonal elements of the change mask with random variables, which we then estimate together with the other parameters and images. 
We propose to use the \emph{inter-image prior} 
\begin{equation}\label{eq:inter_prior_joint} 
    p( \mathbf{x} | \boldsymbol{\gamma} ) 
        \propto \prod_{j=2}^J \, \det\left( C^{(j-1,j)} \right)^{1/2} \exp\left\{ - \frac{1}{2} \left( \mathbf{x}^{(j-1)} - \mathbf{x}^{(j)} \right)^T C^{(j-1,j)} \left( \mathbf{x}^{(j-1)} - \mathbf{x}^{(j)} \right) \right\}
\end{equation}
with $C^{(j-1,j)} = \diag(\boldsymbol{\gamma}^{(j-1,j)})$ and $\boldsymbol{\gamma}^{(j-1,j)} = [\gamma^{(j-1,j)}_1, \dots, \gamma^{(j-1,j)}_N]^T$. 
Furthermore, $\boldsymbol{\gamma}$ in \eqref{eq:inter_prior_joint} denotes the collection of all coupling vectors $\boldsymbol{\gamma}^{(1,2)},\dots,\boldsymbol{\gamma}^{(J-1,J)}$. 
The random variables $\gamma^{(j-1,j)}_n$ introduce an adaptive and spatially varying weighting in the change mask. 
Another advantage of \eqref{eq:inter_prior_joint} is that we stay within the class of conditionally Gaussian densities, which is convenient for Bayesian inference. 
This further motivates us to assume that the elements of $\boldsymbol{\gamma}^{(j-1,j)}$ are i.\,i.\,d.\ gamma distributed, 
\begin{equation}\label{eq:hyperprior_gamma}
    p(\gamma^{(j-1,j)}_n) 
        = \Gamma( \gamma^{(j-1,j)}_n | \eta_{\gamma^{(j-1,j)}_n}, \theta_{\gamma^{(j-1,j)}_n} ), \quad n=1,\dots,N.
\end{equation} 
For simplicity, we use the same shape and rate parameter for all elements, denoted by $\eta_{\gamma}$ and $\theta_{\gamma}$. 
To make the hyper-prior flat and therefore uninformative again, we choose $\theta_{\gamma} \approx 0$. 
As will be discussed in greater detail in \S \ref{sec:tests}, the choice of $\eta_{\gamma}$ is influenced by the magnitude of change we expect to occur between consecutive pairs of images. 
Moreover, if one has a reasonable a priori notion of the location or amount of change between subsequent images, the choice for the hyper-prior \eqref{eq:hyperprior_gamma} and the parameters $\eta_{\gamma^{(j-1,j)}_n}, \theta_{\gamma^{(j-1,j)}_n}$ could be modified correspondingly. 
We will investigate such informative hyper-priors in future works.

\subsection{The combined prior} 
\label{sub:prior} 

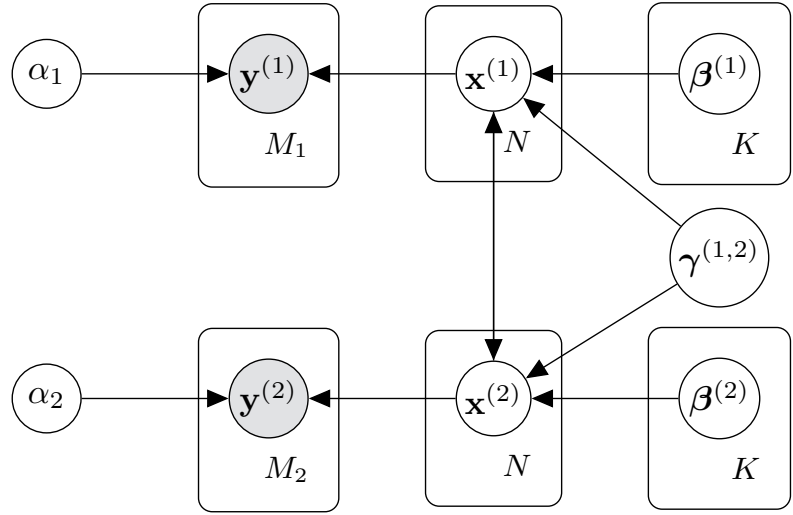
\begin{figure}[tb]
\centering
\resizebox{0.6\textwidth}{!}{%
\begin{tikzpicture}
    \node[obs] (y1) {$\mathbf{y}^{(1)}$}; %
    \node[latent, left=1.5 of y1] (alpha1) {$\alpha_1$}; %
	\node[latent, right=1.5 of y1] (x1) {$\mathbf{x}^{(1)}$} ; %
	\node[latent, right=1.5 of x1] (beta1) {$\boldsymbol{\beta}^{(1)}$} ; %
	\node[latent, below=.95 of beta1] (lambda2) {$\boldsymbol{\gamma}^{(1,2)}$} ; %
	\node[obs, below=2.5 of y1] (y2) {$\mathbf{y}^{(2)}$}; %
	\node[latent, left=1.5 of y2] (alpha2) {$\alpha_2$}; %
	\node[latent, right=1.5 of y2] (x2) {$\mathbf{x}^{(2)}$} ; %
	\node[latent, right=1.5 of x2] (beta2) {$\boldsymbol{\beta}^{(2)}$} ; %
	%
	\edge {alpha1} {y1}; %
	\edge {x1} {y1}; %
	\edge {beta1} {x1}; %
	\edge {alpha2} {y2}; %
	\edge {x2} {y2}; %
	\edge {beta2} {x2}; %
	\edge {lambda2} {x1}; %
	\edge {lambda2} {x2}; %
	\edge {x1} {x2}; %
	\edge {x2} {x1}; %
	\plate[inner sep=0.3cm, xshift=0cm, yshift=0cm] {plate_y1} {(y1)} {$M_1$}; %
	\plate[inner sep=0.3cm, xshift=0cm, yshift=0cm] {plate_x} {(x1)} {$N$}; %
	\plate[inner sep=0.3cm, xshift=0cm, yshift=0cm] {plate_beta} {(beta1)} {$K$}; %
	\plate[inner sep=0.3cm, xshift=0cm, yshift=0cm] {plate_y2} {(y2)} {$M_2$}; %
	\plate[inner sep=0.3cm, xshift=0cm, yshift=0cm] {plate_x} {(x2)} {$N$}; %
	\plate[inner sep=0.3cm, xshift=0cm, yshift=0cm] {plate_beta} {(beta2)} {$K$}; %
\end{tikzpicture} 
}%
\caption{ 
    Graphical representation of the JHBL model for two images. 
  	Shaded and plain circles represent observed and unobserved random variables, respectively. 
	The arrows indicate how the random variables influence each other. 
	More specifically, the noise precisions $\alpha_1, \alpha_2$ and images $\mathbf{x}^{(1)},\mathbf{x}^{(2)}$ are connected to the data vectors $\mathbf{y}^{(1)},\mathbf{y}^{(2)}$ via the likelihood \eqref{eq:likelihood}; 
	The hyper-parameters $\boldsymbol{\beta}^{(1)},\boldsymbol{\beta}^{(2)}$ are connected to the images $\mathbf{x}^{(1)},\mathbf{x}^{(2)}$ via the intra-image prior \eqref{eq:intra_prior_joint}; 
	The images $\mathbf{x}^{(1)}$ and $\mathbf{x}^{(2)}$ are connected to each and the hyper-parameter $\boldsymbol{\gamma}^{(1,2)}$ via the inter-image prior \eqref{eq:inter_prior_joint}, where $\boldsymbol{\gamma}^{(1,2)}$ encodes the (change and no-change) regions and the amount of coupling.  
}
\label{fig:graphical_model}
\end{figure}

We now discuss how promoting intra-image sparsity as in \S \ref{sub:prior_intra} and inter-image coupling as in \S \ref{sub:prior_inter} can be combined in a single prior. 
To this end, we can re-write the joint intra-image prior \eqref{eq:intra_prior_joint} more compactly as 
\begin{equation}\label{eq:intra_prior_joint2} 
    p( \mathbf{x} | \boldsymbol{\beta} ) 
        \propto \det( B )^{1/2} \exp\left\{ - \frac{1}{2} \mathbf{x}^T \tilde{R}^T B \tilde{R} \mathbf{x} \right\}, 
\end{equation} 
where $B = \diag(B^{(1)},\dots,B^{(J)})$ and $\tilde{R} = \diag(R,\dots,R)$ are diagonal matrices, and $\mathbf{x} = [\mathbf{x}^{(1)};\dots;\mathbf{x}^{(J)}]$ again denotes the collection of all images. 
We can now note that the joint intra-image prior \eqref{eq:intra_prior_joint2} encodes the assumption that 
\begin{equation}\label{eq:intra_prior_joint3}  
	\tilde{R} \mathbf{x} \sim \mathcal{N}(\mathbf{0},B^{-1}),
\end{equation}
i.e., $\tilde{R} \mathbf{x}$ is zero-mean normal distributed with diagonal precision matrix $B$. 
At the same time, we can re-write the inter-image prior \eqref{eq:inter_prior_joint} more compactly as 
\begin{equation}\label{eq:inter_prior_joint2} 
    p( \mathbf{x} | \boldsymbol{\gamma} ) 
        \propto \det( C )^{1/2} \exp\left\{ - \frac{1}{2} \mathbf{x}^T S^T C S \mathbf{x} \right\}, 
\end{equation} 
where $C = \diag(C^{(1,2)},\dots,C^{(J-1,J)})$ is a diagonal matrix and 
\begin{equation} 
	S = 
	\begin{bmatrix} 
		I & -I & & \\ 
		  & \ddots & \ddots & \\  
		  & & I & -I
	\end{bmatrix}
\end{equation} 
with $I$ denoting the $N \times N$ identity matrix. 
Basically, $S$ transforms the collection of images $\mathbf{x} = [\mathbf{x}^{(1)};\dots;\mathbf{x}^{(J)}]$ into the collection of differences of sequential images $[\mathbf{x}^{(1)} - \mathbf{x}^{(2)};\dots;\mathbf{x}^{(J-1)} - \mathbf{x}^{(J)}]$. 
Similar to before, we can therefore note that the inter-image prior \eqref{eq:inter_prior_joint2} encodes the assumption that 
\begin{equation}\label{eq:inter_prior_joint3}  
	S \mathbf{x} \sim \mathcal{N}(\mathbf{0},C^{-1}),
\end{equation}
Having \eqref{eq:intra_prior_joint3} and \eqref{eq:inter_prior_joint3} at hand is convenient since it positions us to combine these two assumptions in a single prior. 
To this end, note that \eqref{eq:intra_prior_joint3} and \eqref{eq:inter_prior_joint3} holding simultaneously is equivalent to 
\begin{equation}\label{eq:combined_prior}  
	\begin{bmatrix} \tilde{R} \\ S \end{bmatrix} \mathbf{x} 
	\sim 
	\mathcal{N}\left(\mathbf{0}, \begin{bmatrix} B & 0 \\ 0 & C \end{bmatrix}^{-1} \right),
\end{equation} 
which corresponds to the \emph{combined prior} 
\begin{equation}\label{eq:combined_prior15} 
    p( \mathbf{x} | \boldsymbol{\beta}, \boldsymbol{\gamma} ) 
        \propto \det( \tilde{R}^T B \tilde{R} + S^T C S )^{1/2} \exp\left\{ - \frac{1}{2} \mathbf{x}^T ( \tilde{R}^T B \tilde{R} + S^T C S ) \mathbf{x} \right\}. 
\end{equation} 
The combined prior \eqref{eq:combined_prior15} is a desirable model, but it can have issues when the covariance matrix $\tilde{R}^T B \tilde{R} + S^T C S$ is non-invertible\footnote{%
For example, consider the discrete gradient operator $R$ with $[R \mathbf{x}^{(j)}]k = x^{(j)}{k+1} - x^{(j)}_k$. 
If $\mathbf{x}^{(1)} = \dots = \mathbf{x}^{(J)}$ is a constant sequence of constant images, where $\mathbf{x}^{(j)} = C \cdot [1,\dots,1]^T$ and $C \in \mathbb{R}$, then $\tilde{R}^T B \tilde{R} + S^T C S$ maps it to zero. 
This illustrates that the kernel of $\tilde{R}^T B \tilde{R} + S^T C S$ is not trivial and, as a result, $\tilde{R}^T B \tilde{R} + S^T C S$ is non-invertible.
}, 
rendering the normalizing constant $\det( \tilde{R}^T B \tilde{R} + S^T C S )^{1/2}$ zero. 
In this case, one may consider making the covariance matrix invertible by adding a small multiple of the identity to it. 
This approximation would allow some variability in the directions that collapse without changing in any substantial manner the variance along to dominant directions.
Still, even when the covariance matrix is invertible, Bayesian inference can become computationally intractable due to the complicated relationship between the eigenvalues of $\tilde{R}^T B \tilde{R} + S^T C S$ and the hyper-parameters $\boldsymbol{\beta}$ and $\boldsymbol{\gamma}$. 
To overcome these challenges, we propose the following modification to the combined prior \eqref{eq:combined_prior15}: 
\begin{equation}\label{eq:combined_prior2} 
    p( \mathbf{x} | \boldsymbol{\beta}, \boldsymbol{\gamma} ) 
        \propto \det( B )^{1/2} \det( C )^{1/2} \exp\left\{ - \frac{1}{2} \mathbf{x}^T ( \tilde{R}^T B \tilde{R} + S^T C S ) \mathbf{x} \right\}. 
\end{equation} 
The modified combined prior \eqref{eq:combined_prior2} offers computational benefits and produces satisfactory numerical results. 
We chose this particular form as the resulting fully conditional distributions for the hyper-parameters $\boldsymbol{\beta}$ and $\boldsymbol{\gamma}$ are gamma distributions, making Bayesian maximum a posteriori estimation computationally efficient. 
Further details on the proposed model's inference are provided in \S\ref{sec:inference}. 
We acknowledge that this modification is ad-hoc, and future research could explore other potential modifications.
Overall, the modified combined prior strikes a balance between modeling considerations and computational efficiency. 
Figure \ref{fig:graphical_model} provides a graphical summary of the JHBL model described in this section. 

\begin{remark}
	The joint intra-image and inter-image priors, \eqref{eq:intra_prior_joint2} and \eqref{eq:inter_prior_joint2}, are not necessarily consistent. 
	That is, the marginal distributions of $\mathbf{x}$ derived from the conditional distributions of $\mathbf{x}|\boldsymbol{\beta}$ and $\mathbf{x}|\boldsymbol{\gamma}$, respectively, can differ. 
	For this reason, we advise against performing Bayesian inference using the marginal prior. 
	Instead, our Bayesian inference algorithm outlined in \S\ref{sec:inference} will utilize the hierarchical structure of the Bayesian model. 
\end{remark}
\section{Bayesian inference} 
\label{sec:inference}

Assume that we are interested in the most probable value (mode) of the posterior $p( \mathbf{x}, \boldsymbol{\alpha}, \boldsymbol{\beta}, \boldsymbol{\gamma} | \mathbf{y} )$, i.\,e., a combination of images and parameters for which the given data sets are most likely.
To this end, we adopt the Bayesian coordinate descent (BCD) algorithm \cite{glaubitz2022generalized}, which efficiently approximates the posterior mode by alternatingly updating the images and parameters based on the mode of the fully conditional distributions.

\subsection{The fully conditional distributions} 
\label{sub:cond}

We chose conditionally Gaussian priors and gamma hyper-priors because they are conditionally conjugate. 
This allows us to derive analytical expressions for the fully conditional densities, which is convenient for Bayesian inference.  
Bayes' theorem states that 
\begin{equation}\label{eq:posterior} 
\begin{aligned}
    p( \mathbf{x}, \boldsymbol{\alpha}, \boldsymbol{\beta}, \boldsymbol{\gamma} | \mathbf{y} ) 
        & \propto p( \mathbf{y} | \mathbf{x}, \boldsymbol{\alpha}, \boldsymbol{\beta}, \boldsymbol{\gamma} ) p( \mathbf{x}, \boldsymbol{\alpha}, \boldsymbol{\beta}, \boldsymbol{\gamma} ) \\
        & = p( \mathbf{y} | \mathbf{x}, \boldsymbol{\alpha} ) p( \mathbf{x} | \boldsymbol{\beta}, \boldsymbol{\gamma} ) p( \boldsymbol{\alpha} ) p( \boldsymbol{\beta} ) p( \boldsymbol{\gamma} ), 
\end{aligned}
\end{equation} 
where $p( \mathbf{y} | \mathbf{x}, \boldsymbol{\alpha} )$ is the likelihood, $p( \mathbf{x} | \boldsymbol{\beta}, \boldsymbol{\gamma} )$ is the prior, and $p( \boldsymbol{\alpha} ), p( \boldsymbol{\beta} ), p( \boldsymbol{\gamma} )$ are the hyper-priors. 
If we substitute the joint likelihood \eqref{eq:likelihood_joint} and combined prior \eqref{eq:combined_prior2} together with the hyper-priors \eqref{eq:hyperprior_alpha}, \eqref{eq:hyperprior_beta}, \eqref{eq:hyperprior_gamma} into \eqref{eq:posterior}, we get 
\begin{equation}\label{eq:posterior2} 
\begin{aligned}
    & p( \mathbf{x}, \boldsymbol{\alpha}, \boldsymbol{\beta}, \boldsymbol{\gamma} | \mathbf{y} ) \\
        \propto & \left( \prod_{j=1}^J \alpha_j^{M_j/2} \exp\left\{ -\frac{\alpha_j}{2} \| F^{(j)} \mathbf{x}^{(j)} - \mathbf{y}^{(j)} \|_2^2 \right\} \right) \cdot \\ 
        & \left( \prod_{j=1}^J \det\left( B^{(j)} \right)^{1/2} \exp\left\{ -\frac{1}{2} ( \mathbf{x}^{(j)} )^T R^T B^{(j)} R \mathbf{x}^{(j)} \right\} \right) \cdot \\ 
        & \left( \prod_{j=2}^J \, \det\left( C^{(j-1,j)} \right)^{1/2} \exp\left\{ -\frac{1}{2} \left( \mathbf{x}^{(j-1)} - \mathbf{x}^{(j)} \right)^T C^{(j-1,j)} \left( \mathbf{x}^{(j-1)} - \mathbf{x}^{(j)} \right) \right\} \right) \cdot \\ 
        & \left( \prod_{j=1}^J \alpha_j^{\eta_{\alpha}-1} \exp\{ -\theta_{\alpha} \alpha_j \} \right)  
        \left( \prod_{j=1}^J \beta_j^{\eta_{\beta}-1} \exp\{ -\theta_{\beta} \beta_j \} \right) 
        \left( \prod_{j=2}^J \gamma_j^{\eta_{\gamma}-1} \exp\{ -\theta_{\gamma} \gamma_j \} \right). 
\end{aligned}
\end{equation}
We can note from \eqref{eq:posterior2} that 
\begin{equation} 
\begin{aligned} 
    p\left( \alpha_j | \mathbf{x}^{(j)}, \mathbf{y}^{(j)} \right) 
        & \propto \alpha_j^{M_j/2 + \eta_\alpha - 1} \exp\left\{ - \left( \| F^{(j)} \mathbf{x}^{(j)} - \mathbf{y}^{(j)} \|_2^2/2 + \theta_\alpha \right) \alpha_j \right\}, \\ 
    p\left( \beta^{(j)}_k | \mathbf{x}^{(j)} \right) 
        & \propto \left( \beta^{(j)}_k \right)^{1/2 + \eta_\beta - 1} \exp\left\{ - \left( [R\mathbf{x}^{(j)}]_k^2/2 + \theta_\beta \right) \beta^{(j)}_k \right\}, \\ 
    p\left( \gamma_n^{(j-1,j)} | \mathbf{x}^{(j-1)}, \mathbf{x}^{(j)} \right) 
        & \propto \left( \gamma_n^{(j-1,j)} \right)^{1/2 + \eta_\gamma - 1} \exp\left\{ - \left( [ \mathbf{x}^{(j-1)} - \mathbf{x}^{(j)} ]_n^2/2 + \theta_\gamma \right) \gamma_n^{(j-1,j)} \right\}, 
\end{aligned}
\end{equation}
and thus 
\begin{align}
    p\left( \alpha_j | \mathbf{x}^{(j)}, \mathbf{y}^{(j)} \right) 
        & \propto \Gamma\left( \alpha_j | \eta_{\alpha} + M_j/2, \theta_{\alpha} + \| F^{(j)} \mathbf{x}^{(j)} - \mathbf{y}^{(j)} \|_2^2/2 \right), \label{eq:cond_post_alpha} \\ 
    p\left( \beta^{(j)}_k | \mathbf{x}^{(j)} \right) 
        & \propto \Gamma\left( \beta^{(j)}_k | \eta_{\beta} + 1/2, \theta_{\beta} + [R\mathbf{x}^{(j)}]_k^2/2 \right), \label{eq:cond_post_beta} \\ 
    p\left( \gamma_n^{(j-1,j)} | \mathbf{x}^{(j-1)}, \mathbf{x}^{(j)} \right) 
        & \propto \Gamma\left( \gamma_n^{(j-1,j)} | \eta_{\gamma} + 1/2, \theta_{\gamma} + [ \mathbf{x}^{(j-1)} - \mathbf{x}^{(j)} ]_n^2/2 \right), \label{eq:cond_post_gamma}
\end{align} 
where \eqref{eq:cond_post_alpha}, \eqref{eq:cond_post_beta} hold for $j=1,\dots,J$ and $k=1,\dots,K$, while \eqref{eq:cond_post_gamma} holds for $j=2,\dots,J$ and $n=1,\dots,N$. 
Similarly, for $j=2,\dots,J-1$, we have 
\begin{equation}\label{eq:cond_post_X}
\begin{aligned}
    p\left( \mathbf{x}^{(1)} | \mathbf{x}^{(2)}, \mathbf{y}^{(1)}, \alpha_1, \boldsymbol{\beta}^{(1)}, \boldsymbol{\gamma}^{(1,2)} \right) 
        & \propto \mathcal{N}\left( \mathbf{x}^{(1)} | \boldsymbol{\mu}^{(1)}, \Sigma^{(1)} \right), \\ 
    p\left( \mathbf{x}^{(j)} | \mathbf{x}^{(j-1)}, \mathbf{x}^{(j+1)}, \mathbf{y}^{(j)}, \alpha_j, \boldsymbol{\beta}^{(j)}, \boldsymbol{\gamma}^{(j-1,j)}, \boldsymbol{\gamma}^{(j,j+1)} \right)  
        & \propto \mathcal{N}\left( \mathbf{x}^{(j)} | \boldsymbol{\mu}^{(j)}, \Sigma^{(j)} \right), \\ 
    p\left( \mathbf{x}^{(J)} | \mathbf{x}^{(J-1)}, \mathbf{y}^{(J)}, \alpha_J, \boldsymbol{\beta}^{(J)}, \boldsymbol{\gamma}^{(J-1,J)} \right)  
        & \propto \mathcal{N}\left( \mathbf{x}^{(N)} | \boldsymbol{\mu}^{(N)}, \Sigma^{(N)} \right),
\end{aligned}
\end{equation}
with covariance matrices 
\begin{equation}\label{eq:covariances}
\begin{aligned}
    \Sigma^{(1)} & = \left( \alpha_1 \left( F^{(1)} \right)^T F^{(1)} + R^T B^{(1)} R + C^{(1,2)} \right)^{-1}, \\ 
    \Sigma^{(j)} & = \left( \alpha_j \left( F^{(j)} \right)^T F^{(j)} + R^T B^{(j)} R + C^{(j-1,j)} + C^{(j,j+1)} \right)^{-1}, \\ 
    \Sigma^{(J)} & = \left( \alpha_J \left( F^{(J)} \right)^T F^{(J)} + R^T B^{(J)} R + C^{(J-1,J)} \right)^{-1}, \\ 
\end{aligned}
\end{equation}
and means
\begin{equation}\label{eq:means}
\begin{aligned}
    \boldsymbol{\mu}^{(1)} & = \Sigma^{(1)} \left( \alpha_1 \left( F^{(1)} \right)^T \mathbf{y}^{(1)} + C^{(1,2)} \mathbf{x}^{(2)} \right), \\ 
    \boldsymbol{\mu}^{(j)} & = \Sigma^{(j)} \left( \alpha_j \left( F^{(j)} \right)^T \mathbf{y}^{(j)} + C^{(j-1,j)} \mathbf{x}^{(j-1)} + C^{(j,j+1)} \mathbf{x}^{(j+1)} \right), \\ 
    \boldsymbol{\mu}^{(J)} & = \Sigma^{(J)} \left( \alpha_J \left( F^{(J)} \right)^T \mathbf{y}^{(J)} + C^{(J-1,J)} \mathbf{x}^{(J-1)} \right). 
\end{aligned}
\end{equation} 
Since $\alpha_j$, $\beta^{(j)}_k$, and $\gamma^{(j-1,j)}_n$ are gamma distributed and only take on positive values, the covariance matrices in \eqref{eq:covariances} are symmetric and positive definite (SPD).

\subsection{Proposed method: Joint hierarchical Bayesian learning} 
\label{sub:BCD}

We are now positioned to formulate our JHBL method by adapting the BCD algorithm \cite{glaubitz2022generalized} to our model in \S\ref{sec:model}. 
The BCD algorithm is described in Algorithm \ref{algo:BCD} and approximates the mode (or mean) of the posterior $p( \mathbf{x}, \boldsymbol{\alpha}, \boldsymbol{\beta}, \boldsymbol{\gamma} | \mathbf{y} )$ by alternatingly updating the images and parameters based on the mode (or mean) of the fully conditional distributions \eqref{eq:cond_post_alpha}, \eqref{eq:cond_post_beta}, \eqref{eq:cond_post_gamma}, \eqref{eq:cond_post_X}.  

\begin{algorithm}[h!]
\caption{BCD algorithm}\label{algo:BCD} 
\begin{algorithmic}[1]
    \STATE{Initialize $\mathbf{x}^{0}$ and set $l=0$}
    \REPEAT
        \STATE{Update $\boldsymbol{\alpha}$ by setting $\boldsymbol{\alpha}^{l+1} = \argmax_{\boldsymbol{\alpha}} p(\boldsymbol{\alpha}|\mathbf{x}^{l})$}
    	\STATE{Update $\boldsymbol{\beta}$ by setting $\boldsymbol{\beta}^{l+1} = \argmax_{\boldsymbol{\beta}} p(\boldsymbol{\beta}|\mathbf{x}^{l})$}
    	\STATE{Update $\boldsymbol{\gamma}$ by setting $\boldsymbol{\gamma}^{l+1} = \argmax_{\boldsymbol{\gamma}} p(\boldsymbol{\gamma}|\mathbf{x}^{l})$}
    	\STATE{Update $\mathbf{x}$ by setting $\mathbf{x}^{l+1} = \argmax_{\mathbf{x}} p(\mathbf{x}|\mathbf{x}^{l},\boldsymbol{\alpha}^{l+1},\boldsymbol{\beta}^{l+1},\boldsymbol{\gamma}^{l+1},\mathbf{y})$}
    	\STATE{Increase $l \to l+1$}
    \UNTIL{convergence or maximum number of iterations is reached}
\end{algorithmic}
\end{algorithm} 

Algorithm \ref{algo:BCD} is efficient and straightforward to implement because of the analytical expressions for the fully conditional distributions we derived in \eqref{eq:cond_post_alpha}, \eqref{eq:cond_post_beta}, \eqref{eq:cond_post_gamma}, \eqref{eq:cond_post_X}. 
The mode of a gamma distribution $\Gamma(\eta,\theta)$ with $\eta,\theta > 0$ is $\max\{0,(\eta-1)/\theta\}$. 
Thus, \eqref{eq:cond_post_alpha}, \eqref{eq:cond_post_beta}, \eqref{eq:cond_post_gamma} imply that the $\boldsymbol{\alpha}$-, $\boldsymbol{\beta}$-, and $\boldsymbol{\gamma}$-updates in Algorithm \ref{algo:BCD} are equivalent to 
\begin{alignat}{2}
    \left\{ \alpha_j \right\}^{l+1} 
        & = \frac{ \eta_{\alpha} + M_j/2 - 1}{ \theta_{\alpha} + \| F^{(j)}  \{ \mathbf{x}^{(j)} \}^{l} - \mathbf{y}^{(j)} \|_2^2/2}, \quad 
        && j=1,\dots,J, \label{eq:update_alpha} \\ 
    \left\{ \beta^{(j)}_k \right\}^{l+1} 
        & = \frac{ \eta_{\beta} - 1/2 }{ \theta_{\beta} + [ R \{ \mathbf{x}^{(j)} \}^{l} ]_k^2/2 }, \quad 
        && j=1,\dots,J,\ k=1,\dots,K, \label{eq:update_beta} \\ 
    \left\{ \gamma^{(j-1,j)}_n \right\}^{l} 
        & = \frac{ \eta_{\gamma} - 1/2 }{ \theta_{\gamma} + [ \{ \mathbf{x}^{(j-1)} \}^{l} - \{ \mathbf{x}^{(j)} \}^{l} ]_n^2/2}, \quad 
        && j=2,\dots,J,\ n=1,\dots,N, \label{eq:update_gamma}
\end{alignat}
assuming nonnegative numerators, where $\{ \ \}^{l}$ denotes the $l$-th iteration of the term inside the curved brackets. 
Further, \eqref{eq:cond_post_X} implies that the $\mathbf{x}$-update in Algorithm \ref{algo:BCD} is equivalent to solving the linear systems 
\begin{equation}\label{eq:update_x}
    \left\{ G^{(j)} \right\}^{l+1} \left\{ \mathbf{x}^{(j)} \right\}^{l+1} 
        = \left\{ \mathbf{b}^{(j)} \right\}^{l+1}, \quad j=1,\dots,J,  
\end{equation}
with SPD coefficient matrices 
\begin{equation}\label{eq:matrix_G} 
\begin{aligned}
    \left\{ G^{(1)} \right\}^{l+1} 
        = & \ \{ \alpha_1 \}^{l+1} (F^{(1)})^T F^{(1)} + R^T \{ B^{(1)} \}^{l+1} R + \{ C^{(1,2)} \}^{l+1}, \\ 
    \left\{ G^{(j)} \right\}^{l+1} 
        = & \ \{ \alpha_j \}^{l+1} (F^{(j)})^T F^{(j)} + R^T \{ B^{(j)} \}^{l+1} R + \{ C^{(j-1,j)} \}^{l+1} \\ 
        & + \{ C^{(j,j+1)} \}^{l+1}, \\ 
    \left\{ G^{(J)} \right\}^{l+1} 
        = & \ \{ \alpha_J \}^{l+1} (F^{(J)})^T F^{(J)} + R^T \{ B^{(J)} \}^{l+1} R + \{ C^{(J-1,J)} \}^{l+1},
\end{aligned} 
\end{equation}
and right-hand sides 
\begin{equation}\label{eq:RHS_b}
\begin{aligned}
    \left\{ \mathbf{b}^{(1)} \right\}^{l+1} 
        = & \ \{ \alpha_1 \}^{l+1} ( F^{(1)} )^T \mathbf{y}^{(1)} + \{ C^{(1,2)} \}^{l+1} \{ \mathbf{x}^{(2)} \}^{l}, \\ 
    \left\{ \mathbf{b}^{(j)} \right\}^{l+1} 
        = & \ \{ \alpha_j \}^{l+1} ( F^{(1)} )^T \mathbf{y}^{(j)} + \{ C^{(j-1,j)} \}^{l+1} \{ \mathbf{x}^{(j-1)} \}^{l} \\ 
        & + \{ C^{(j,j+1)} \}^{l+1} \{ \mathbf{x}^{(j+1)} \}^{l}, \\
    \left\{ \mathbf{b}^{(J)} \right\}^{l+1} 
        = & \ \{ \alpha_J \}^{l+1} ( F^{(J)} )^T \mathbf{y}^{(J)} + \{ C^{(J-1,J)} \}^{l+1} \{ \mathbf{x}^{(J-1)} \}^{l}, 
\end{aligned}
\end{equation}
for $j=2,\dots,J-1$.
Notably, we compute the $(l+1)$-th iteration of the $\mathbf{x}^{(j)}$'s using the $l$-th iteration of the neighboring images in \eqref{eq:RHS_b}. 
This allows us to parallelize the $\mathbf{x}^{(j)}$-updates \eqref{eq:update_x}, making our method efficient even for large image sequences.
We can now summarize our JHBL method for sequential image recovery as in Algorithm \ref{algo:JHBL}. 

\begin{algorithm}[h]
\caption{Joint hierarchical Bayesian learning (JHBL) algorithm}\label{algo:JHBL} 
\begin{algorithmic}[1]
    \STATE{Initialize the images $\{ \mathbf{x}^{(1)} \}^{0},\dots,\{ \mathbf{x}^{(J)} \}^{0}$ and set $l=0$}
    \REPEAT
        \STATE{Update the noise parameters $\alpha_1,\dots,\alpha_J$ according to \eqref{eq:update_alpha}}
        \STATE{Update the intra-image parameters $\boldsymbol{\beta}^{(1)},\dots,\boldsymbol{\beta}^{(J)}$ according to \eqref{eq:update_beta}}
        \STATE{Update the inter-image parameters $\boldsymbol{\gamma}^{(1,2)},\dots,\boldsymbol{\gamma}^{(J-1,J)}$ according to \eqref{eq:update_gamma}} 
        \STATE{Update the images $\mathbf{x}^{(1)},\dots,\mathbf{x}^{(J)}$ according to \eqref{eq:update_x}}
    	\STATE{Increase $l \to l+1$}
    \UNTIL{convergence or maximum number of iterations is reached}
\end{algorithmic}
\end{algorithm}

\subsection{Separate recovery as a special case} 
\label{sub:seperate}

We can recover the generalized sparse Bayesian learning (GSBL) method \cite{glaubitz2022generalized} from our JHBL procedure as a special/limit case. 
If $\eta_{\gamma} \leq 1/2$ or $\theta_{\gamma} \to \infty$, then the $\boldsymbol{\gamma}^{(j)}$-update \eqref{eq:update_gamma} becomes 
\begin{equation}
    \left\{ \boldsymbol{\gamma}^{(j-1,j)} \right\}^{l} 
        = 0, \quad j=2,\dots,J,
\end{equation}
and the $\mathbf{x}^{(j)}$-update \eqref{eq:update_x} reduces to 
\begin{equation}
    \left( \{ \alpha_j \}^{l+1} (F^{(j)})^T F^{(j)} + R^T \{ B^{(j)} \}^{l+1} R \right) \left\{ \mathbf{x}^{(j)} \right\}^{l+1} 
        = \{ \alpha_j \}^{l+1} ( F^{(j)} )^T \mathbf{y}^{(j)},
\end{equation}
for $j=1,\dots,J$.
In this case, Algorithm \ref{algo:JHBL} corresponds to using the GSBL algorithm \cite{glaubitz2022generalized} to recover the images separately.

\subsection{Efficient implementation} 
\label{sub:implementation}

A few remarks on Algorithm \ref{algo:JHBL} are in order. 

\begin{remark}[Initialization]
    We initialize the images $\{ \mathbf{x}^{(1)} \}^{0},\dots,\{ \mathbf{x}^{(J)} \}^{0}$ in Algorithm \ref{algo:JHBL} as the separately recovered images using the GSBL algorithm \cite{glaubitz2022generalized}, which we efficiently implemented in parallel. 
\end{remark}

\begin{remark}[Parallelization]
    The different $\mathbf{x}^{j}$-, $\alpha_j$-, $\boldsymbol{\beta}^{(j)}$-, and $\boldsymbol{\gamma}^{(j)}$-updates can be easily parallelized. 
    This makes Algorithm \ref{algo:JHBL} efficient even for large image sequences. 
\end{remark}

\begin{remark}[Efficient $\mathbf{x}^{(j)}$-updates]\label{rem:grad_desc}
    The coefficient matrices in the $\mathbf{x}^{(j)}$-updates \eqref{eq:update_x} can become prohibitively large in imaging applications. 
    To avoid storage and efficiency issues, we identify the solutions of \eqref{eq:update_x} with the unique minimizers of the quadratic functionals 
    \begin{equation}\label{eq:const_function}
        L^{(j)}(\mathbf{x}) = \mathbf{x}^T G^{(j)} \mathbf{x} - 2 \mathbf{x}^T \mathbf{b}^{(j)}, \quad j=1,\dots,J,
    \end{equation} 
    which is possible since the $G^{(j)}$'s are (almost surely) SPD. 
    We then efficiently solve for the minimizers of \eqref{eq:const_function} using a gradient descent method \cite{glaubitz2022generalized}. 
    In our implementation, we use five gradient descent steps for every iteration of the $\mathbf{x}^{(j)}$-updates. 
\end{remark}

\begin{remark}[Stopping criterion]\label{rem:stopping}
    We stop the iterations in Algorithm \ref{algo:JHBL} if the average relative and absolute change between two subsequent image sequence iterations w.\,r.\,t.\ the $\|\cdot\|_2$-norm are less than $10^{-3}$ or if a maximum number of $10^{3}$ iterations is reached. 
\end{remark}

\begin{remark}[Convexity and convergence]
    A detailed analysis regarding the convexity of the cost function, $-\log p(\mathbf{x},\boldsymbol{\alpha},\boldsymbol{\beta},\boldsymbol{\gamma}| \mathbf{y})$, and the convergence of Algorithm \ref{algo:JHBL} exceeds the scope of this paper and will be addressed in future works. 
\end{remark}
\section{Numerical tests} 
\label{sec:tests}

We consider two numerical tests to demonstrate the performance of our JHBL method.

\subsection{Sequential magnetic resonance imaging} 
\label{sub:tests_MRI} 

\begin{figure}[tb]
    \centering
    \begin{subfigure}[b]{.32\textwidth}
    \includegraphics[width=\textwidth]{figures/true_x1_GE.eps}
    \caption{Reference image 1}
    \label{fig:MRI_separate_ref1}
    \end{subfigure}
    \begin{subfigure}[b]{.32\textwidth}
    \includegraphics[width=\textwidth]{figures/true_x2_GE.eps}
    \caption{Reference image 2}
    \label{fig:MRI_separate_ref2}
    \end{subfigure}
    \begin{subfigure}[b]{.32\textwidth}
    \includegraphics[width=\textwidth]{figures/true_x3_GE.eps}
    \caption{Reference image 3}
    \label{fig:MRI_separate_ref3}
    \end{subfigure}
    \\
    \begin{subfigure}[b]{.32\textwidth}
    \includegraphics[width=\textwidth]{figures/sADMM_x1_GE.eps}
    \caption{Separately recovered 1,\\ deterministic}
    \label{fig:MRI_ADMM_sep1}
    \end{subfigure}
    \begin{subfigure}[b]{.32\textwidth}
    \includegraphics[width=\textwidth]{figures/sADMM_x2_GE.eps}
    \caption{Separately recovered 2,\\ deterministic}
    \label{fig:MRI_ADMM_sep2}
    \end{subfigure}
    \begin{subfigure}[b]{.32\textwidth}
    \includegraphics[width=\textwidth]{figures/sADMM_x3_GE.eps}
    \caption{Separately recovered 3,\\ deterministic}
    \label{fig:MRI_ADMM_sep3}
    \end{subfigure}
    \\ 
    \begin{subfigure}[b]{.32\textwidth}
    \includegraphics[width=\textwidth]{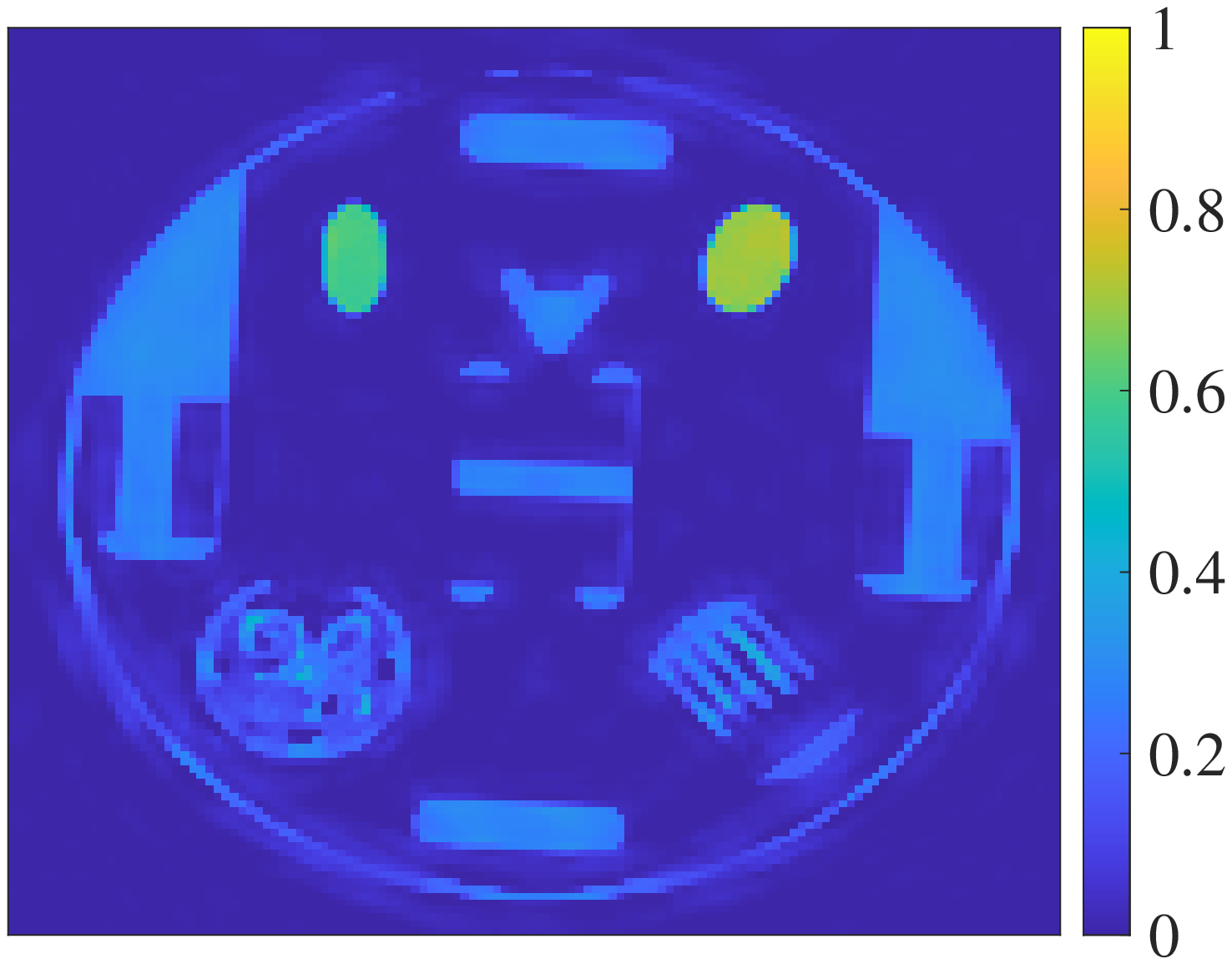}
    \caption{Separately recovered 1,\\ Bayesian}
    \label{fig:MRI_GSBL1} 
    \end{subfigure}
    \begin{subfigure}[b]{.32\textwidth}
    \includegraphics[width=\textwidth]{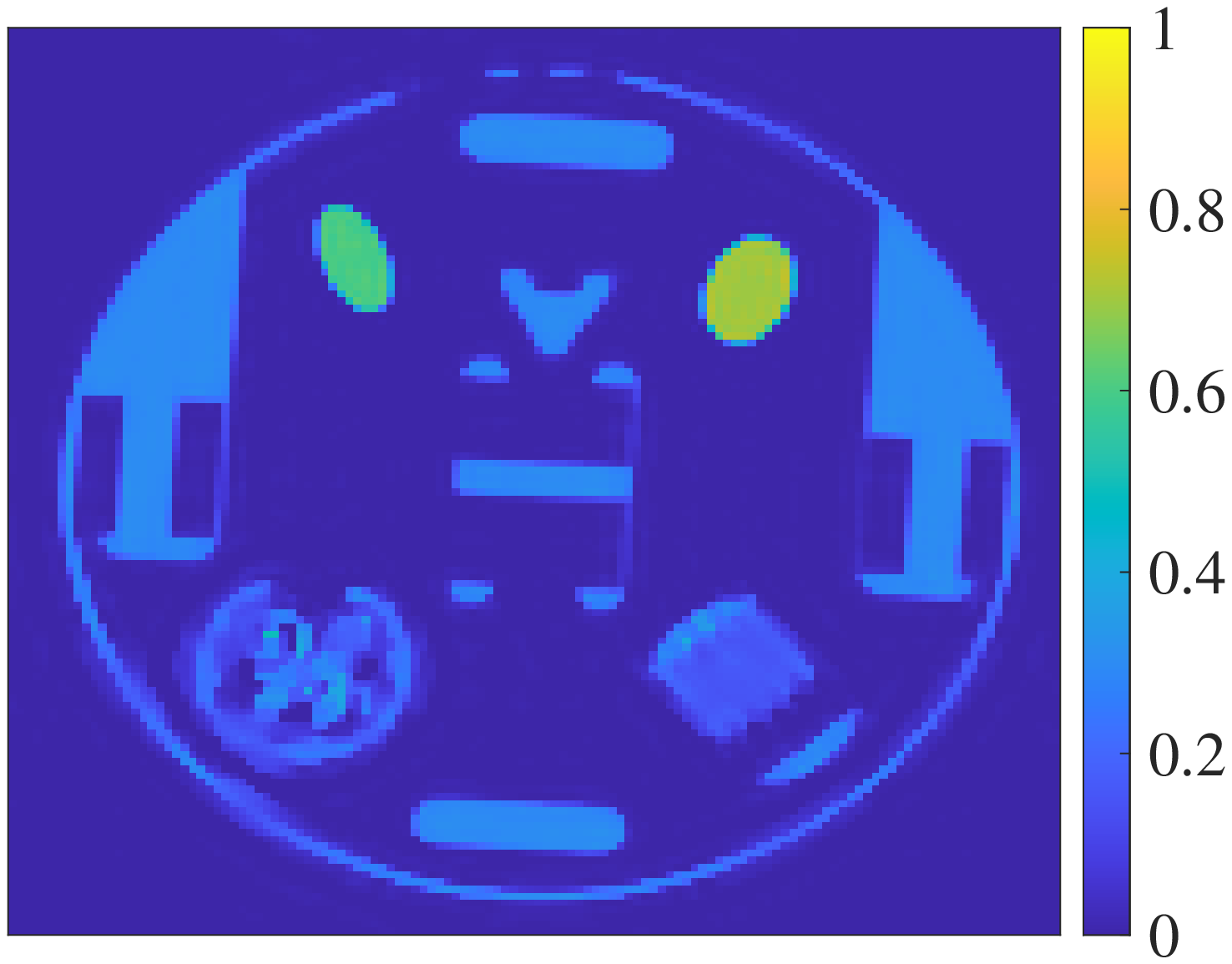}
    \caption{Separately recovered 2,\\ Bayesian} 
    \label{fig:MRI_GSBL2} 
    \end{subfigure}
    \begin{subfigure}[b]{.32\textwidth}
    \includegraphics[width=\textwidth]{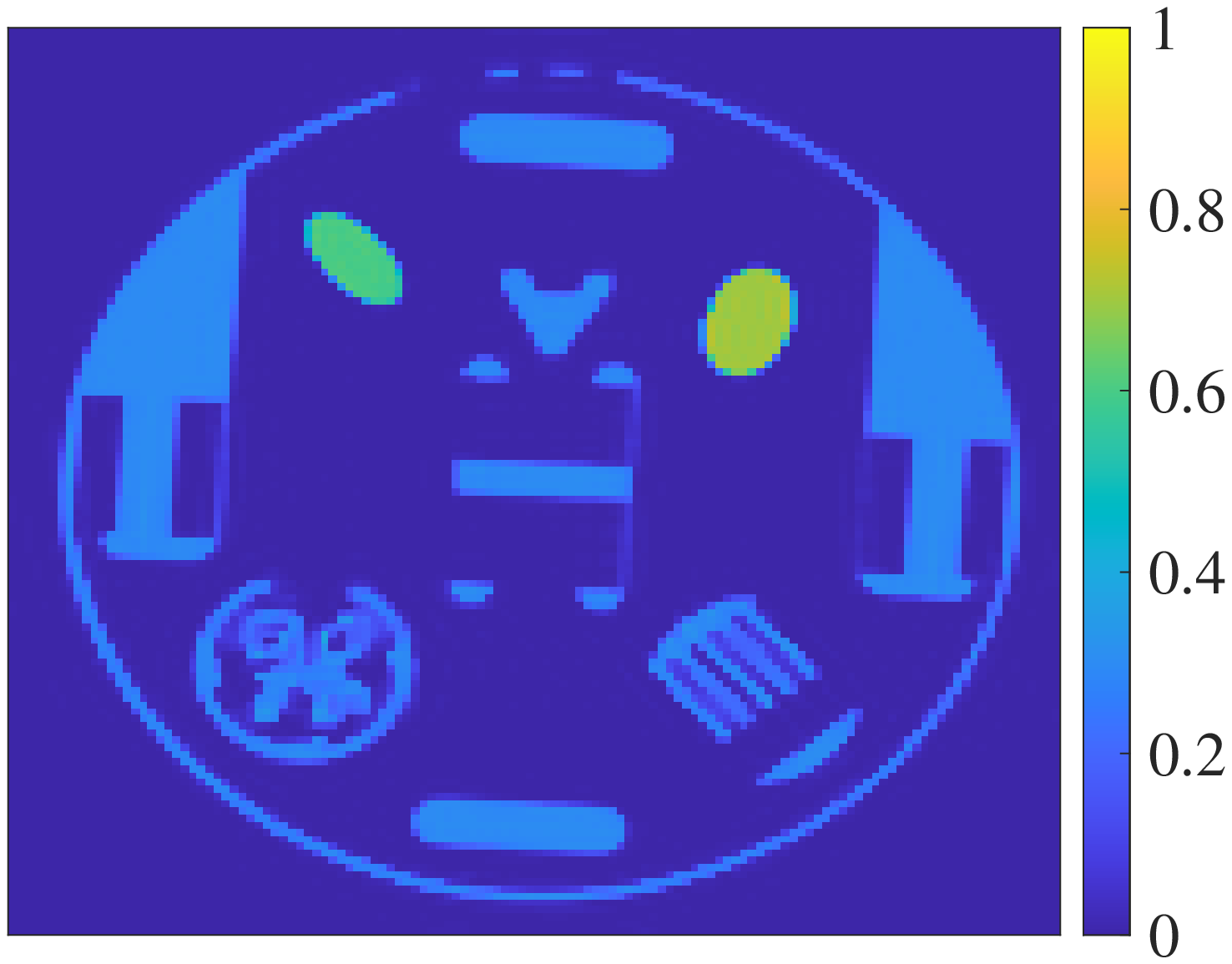}
    \caption{Separately recovered 3,\\ Bayesian} 
    \label{fig:MRI_GSBL3} 
    \end{subfigure}
    \caption{ 
    Three temporal phantom images by a GE HTXT 1.5T clinical MRI scanner \cite{Lalwanietal} (first row) and their separate reconstructions from noisy and under-sampled Fourier data solving the deterministic $\ell^1$-regularized inverse problems \eqref{eq:l1_RIP} (second row) and the Bayesian GSBL algorithm (third row) 
    }
    \label{fig:MRI_sep}
\end{figure}

Given is a temporal sequence of six $128 \times 128$ phantom images obtained by a GE HTXT 1.5T clinical magnetic resonance imaging (MRI) scanner \cite{Lalwanietal}, from which we generate under-sampled and indirect Fourier data. 
Figures \ref{fig:MRI_separate_ref1}, \ref{fig:MRI_separate_ref2}, and \ref{fig:MRI_separate_ref3} show the first three reference images, where the change consists of the rotating left (green) ellipse and the down-moving right (yellow) ellipse. 
The Fourier samples contained in the data sets are 
\begin{equation}\label{eq:Fourier_samples}
    y^{(j)}_{k,l} = \int^1_0\int^1_0 x^{(j)}(s,t)e^{-i2\pi (ks+lt)}\intd s \intd t, 
	\quad - \left\lceil \frac{N_1}{2} \right\rceil \le k,l < \left\lceil \frac{N_1}{2} \right\rceil, 
\end{equation} 
for $j = 1,\dots,J$, where $N = N_1^2$ and $x^{(j)}$ is a function descibing the $j$th image. 
We use the discrete Fourier transform as a linear forward operator, thereby introducing model discrepancy and avoiding the inverse crime \cite{kaipio2007statistical}. 
Further, each data set is missing Fourier samples for the symmetric bands 
\begin{equation}\label{eq:MRI_bands_removed}
    \mathcal{K}_j = \left[ \pm (10j + 1), \pm (10j+10) \right]^2, \quad j=1,\dots,6,
\end{equation} 
and the remaining Fourier samples contain additive i.\,i.\,d.\ zero-mean normal noise. 
The amount of noise is measured using the signal-to-noise ratio (SNR) 
\begin{equation}\label{eq:snr}
    {\rm SNR}^{(j)} = 10 \log_{10} \left( \alpha_j y^{(j)}_{0,0} \right), \quad j=1,\dots,J,
\end{equation}
where $y^{(j)}_{0,0}$ is the average of the $j$th image. 
The SNR for all images in Figure \ref{fig:MRI_sep} is $2$.  

Figures \ref{fig:MRI_ADMM_sep1}, \ref{fig:MRI_ADMM_sep2}, and \ref{fig:MRI_ADMM_sep3} illustrate the separately recovered images from the noisy and under-sampled Fourier data by solving the (weighted) $\ell^1$-regularized inverse problems \eqref{eq:l1_RIP} using the alternating directions method of multipliers (ADMM) \cite{boyd2011distributed,xiao2022sequential}. 
Figures \ref{fig:MRI_GSBL1}, \ref{fig:MRI_GSBL2}, and \ref{fig:MRI_GSBL3} visualizes the separately recovered images from the same data sets using the GSBL algorithm \cite{glaubitz2022generalized} with hyper-parameters $\eta_{\alpha} = \eta_{\beta} = 1$ and $\theta_{\alpha} = \theta_{\beta} = 10^{-3}$. 
In both cases, we used an anisotropic first-order TV regularization operator
\begin{equation}\label{eq:MRI_reg}
    R = \begin{bmatrix} I \otimes D \\ D \otimes I \end{bmatrix} 
    \quad \text{with} \quad 
    D = 
    \begin{bmatrix}
        -1 & 1 & & \\ 
         & \ddots & \ddots & \\ 
         & & -1 & 1 
    \end{bmatrix} 
    \in \R^{(N_1-1) \times N_1},  
\end{equation} 
to promote the images being piecewise constant. 
Figure \ref{fig:MRI_sep} demonstrates that the Bayesian GSBL algorithm separately recovers the images more accurately than the deterministic algorithm, although we still observe some smeared features in Figures \ref{fig:MRI_GSBL1} and \ref{fig:MRI_GSBL2}.  

\begin{figure}[tb]
    \centering
    \begin{subfigure}[b]{.32\textwidth}
    \includegraphics[width=\textwidth]{figures/true_x1_GE.eps}
    \caption{Reference image 1}
    \end{subfigure}
    \begin{subfigure}[b]{.32\textwidth}
    \includegraphics[width=\textwidth]{figures/true_x2_GE.eps}
    \caption{Reference image 2}
    \end{subfigure}
    \begin{subfigure}[b]{.32\textwidth}
    \includegraphics[width=\textwidth]{figures/true_x3_GE.eps}
    \caption{Reference image 3}
    \end{subfigure}
    \\
    \begin{subfigure}[b]{.32\textwidth}
    \includegraphics[width=\textwidth]{figures/cADMM_x1_GE.eps}
    \caption{Jointly recovered 1,\\ deterministic}
    \label{fig:MRI_ADMM_joint1}
    \end{subfigure}
    \begin{subfigure}[b]{.32\textwidth}
    \includegraphics[width=\textwidth]{figures/cADMM_x2_GE.eps}
    \caption{Jointly recovered 2,\\ deterministic}
    \label{fig:MRI_ADMM_joint2}
    \end{subfigure}
    \begin{subfigure}[b]{.32\textwidth}
    \includegraphics[width=\textwidth]{figures/cADMM_x3_GE.eps}
    \caption{Jointly recovered 3,\\ deterministic}
    \label{fig:MRI_ADMM_joint3}
    \end{subfigure}
    \\
    \begin{subfigure}[b]{.32\textwidth}
    \includegraphics[width=\textwidth]{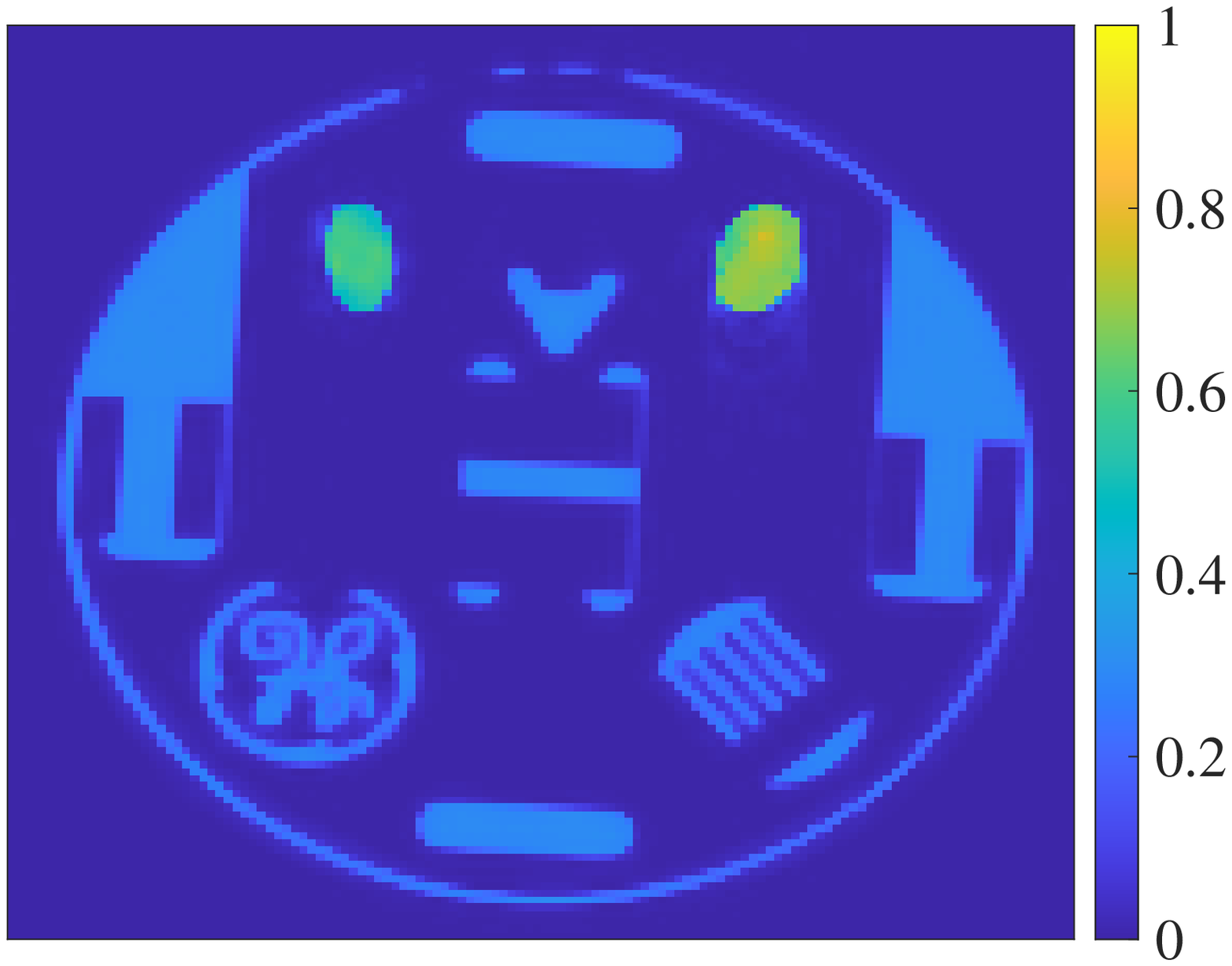}
    \caption{Jointly recovered 1,\\ Bayesian} 
    \label{fig:MRI_JHBL1} 
    \end{subfigure}
    \begin{subfigure}[b]{.32\textwidth}
    \includegraphics[width=\textwidth]{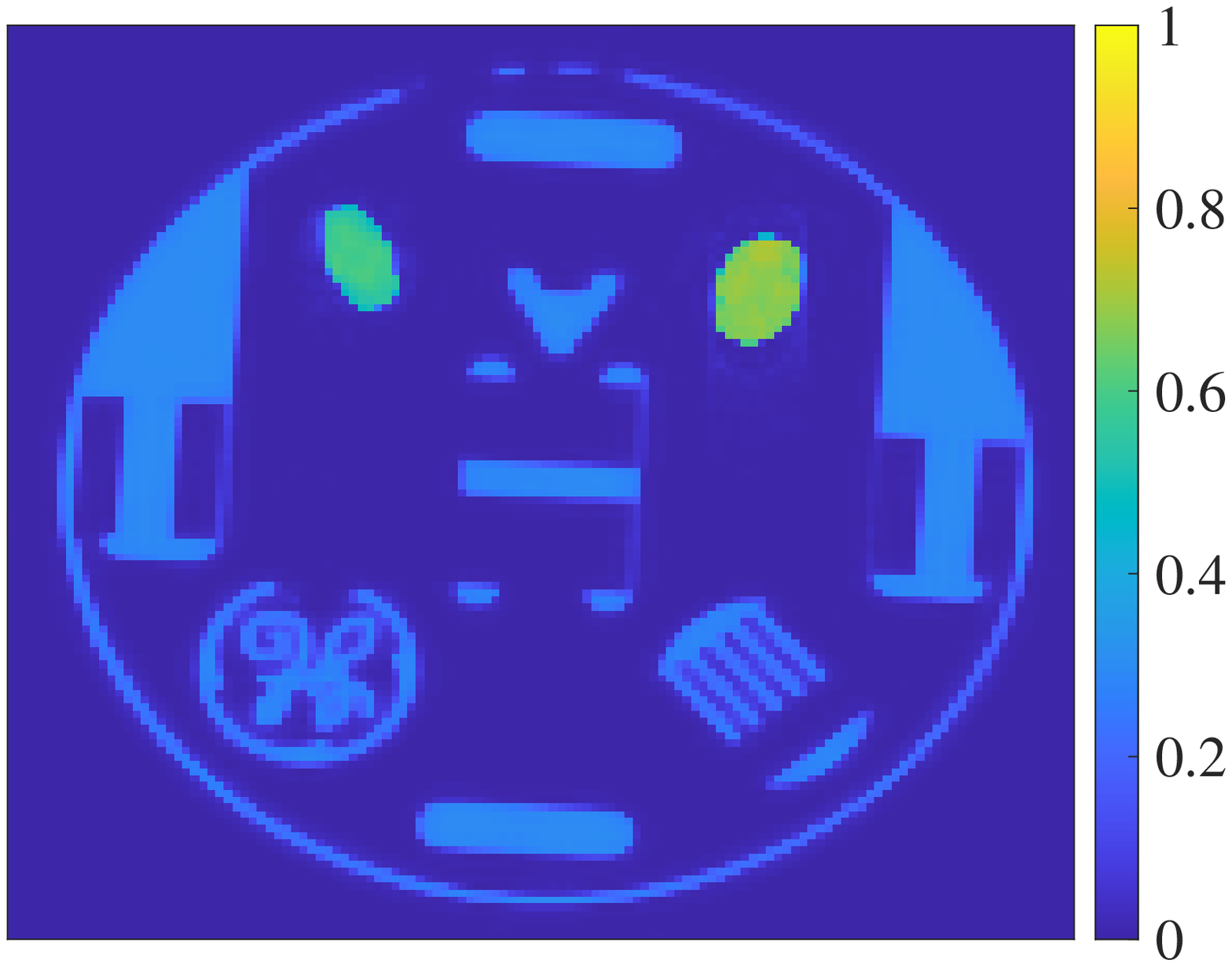}
    \caption{Jointly recovered 2,\\ Bayesian} 
    \label{fig:MRI_JHBL2} 
    \end{subfigure}
    \begin{subfigure}[b]{.32\textwidth}
    \includegraphics[width=\textwidth]{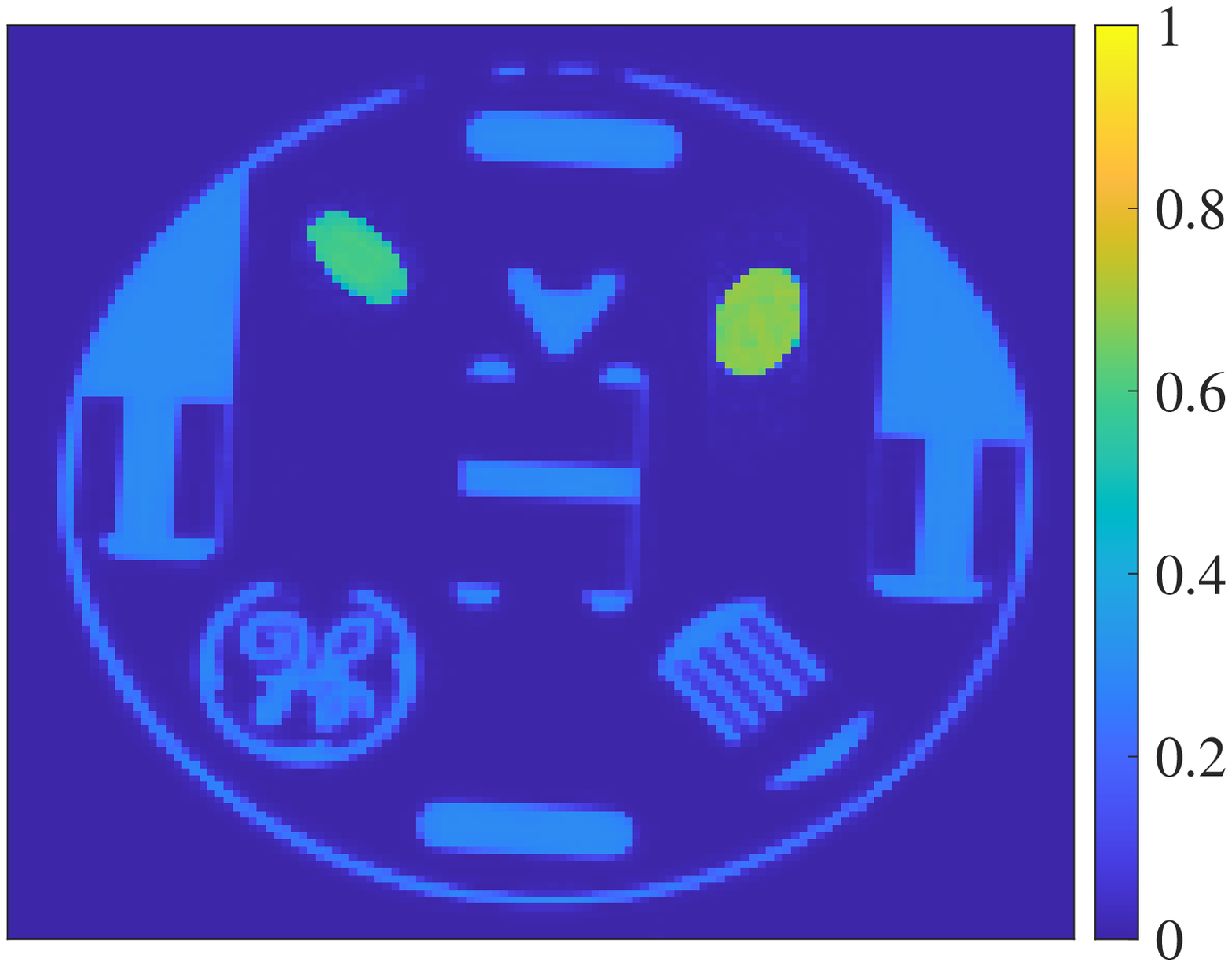}
    \caption{Jointly recovered 3,\\ Bayesian} 
    \label{fig:MRI_JHBL3} 
    \end{subfigure}
    \caption{
    Three temporal phantom images by a GE HTXT 1.5T clinical MRI scanner \cite{Lalwanietal} (first row) and their joint reconstructions from noisy and under-sampled Fourier data using the deterministic weighted $\ell^1$-regularized inverse problems \eqref{eq:joint_RIP} (second row) and the proposed JHBL algorithm (third row) 
    }
    \label{fig:MRI_joint}
\end{figure}

Figure \ref{fig:MRI_joint} illustrates the corresponding jointly recovered images. 
Specifically, Figures \ref{fig:MRI_ADMM_joint1}, \ref{fig:MRI_ADMM_joint2}, and \ref{fig:MRI_ADMM_joint3} visualizes the jointly recovered images using ADMM to solve the joint $\ell^1$-regularized inverse problem \eqref{eq:joint_RIP} as proposed in \cite{xiao2022sequential}, which we use as a benchmark. 
Figures \ref{fig:MRI_JHBL1}, \ref{fig:MRI_JHBL2}, and \ref{fig:MRI_JHBL3} show the jointly recovered images using our JHBL algorithm (Algorithm \ref{algo:JHBL}). 
Following the discussion in \S\ref{sec:model}, we chose the hyper-parameters as $\eta_{\alpha} = \eta_{\beta} = 1$, $\eta_{\gamma} = 2$, and $\theta_{\alpha} = \theta_{\beta} = \theta_{\gamma} = 10^{-3}$. 

\begin{remark}\label{rem:gamma}
	Following \cite{glaubitz2022generalized}, $\eta_{\alpha} = \eta_{\beta} = 1$ and $\theta_{\alpha} = \theta_{\beta} = 10^{-3}$ are usual choices for the GSBL algorithm. 
	Initially, we also tried to use $\eta_{\gamma} = 1$ and $\theta_{\gamma} = 10^{-3}$ for the parameters of the inter-image hyper-prior, but observed that the inter-image coupling was sometimes too strong. 
	We thus used $\eta_{\gamma} = 2$ and $\theta_{\gamma} = 10^{-3}$ instead. 
	The heuristic behind increasing the shape parameter to $\eta_{\gamma}$ is as follows: 
	We assume that every consecutive pair of images contains more no-change regions than change regions. 
We thus expect $\gamma^{(j-1,j)}_n \gg 0$ for most of the $\gamma^{(j-1,j)}_n$'s in \eqref{eq:inter_prior_joint}, indicating no change, and $\gamma^{(j-1,j)}_n \approx 0$ for only a few of the $\gamma^{(j-1,j)}_n$'s. 
	We increasingly promote this behavior for the $\gamma^{(j-1,j)}_n$'s by increasing the shape parameter $\eta_{\gamma}$ in \eqref{eq:hyperprior_gamma}. 
	We further investigate the influence of the shape and rate parameter $\eta_{\gamma}$ and $\theta_{\gamma}$ on the inter-image coupling in \S \ref{sub:param_investigation}.
\end{remark}

Both joint methods yield more accurate recovered images than the respective separate method. 
At the same time, our JHBL algorithm provides notably more accurate recovered images than the deterministic joint method proposed in \cite{xiao2022sequential}. 

\begin{figure}[tb]
    \centering
    \begin{subfigure}[b]{.45\textwidth}
    \includegraphics[width=\textwidth]{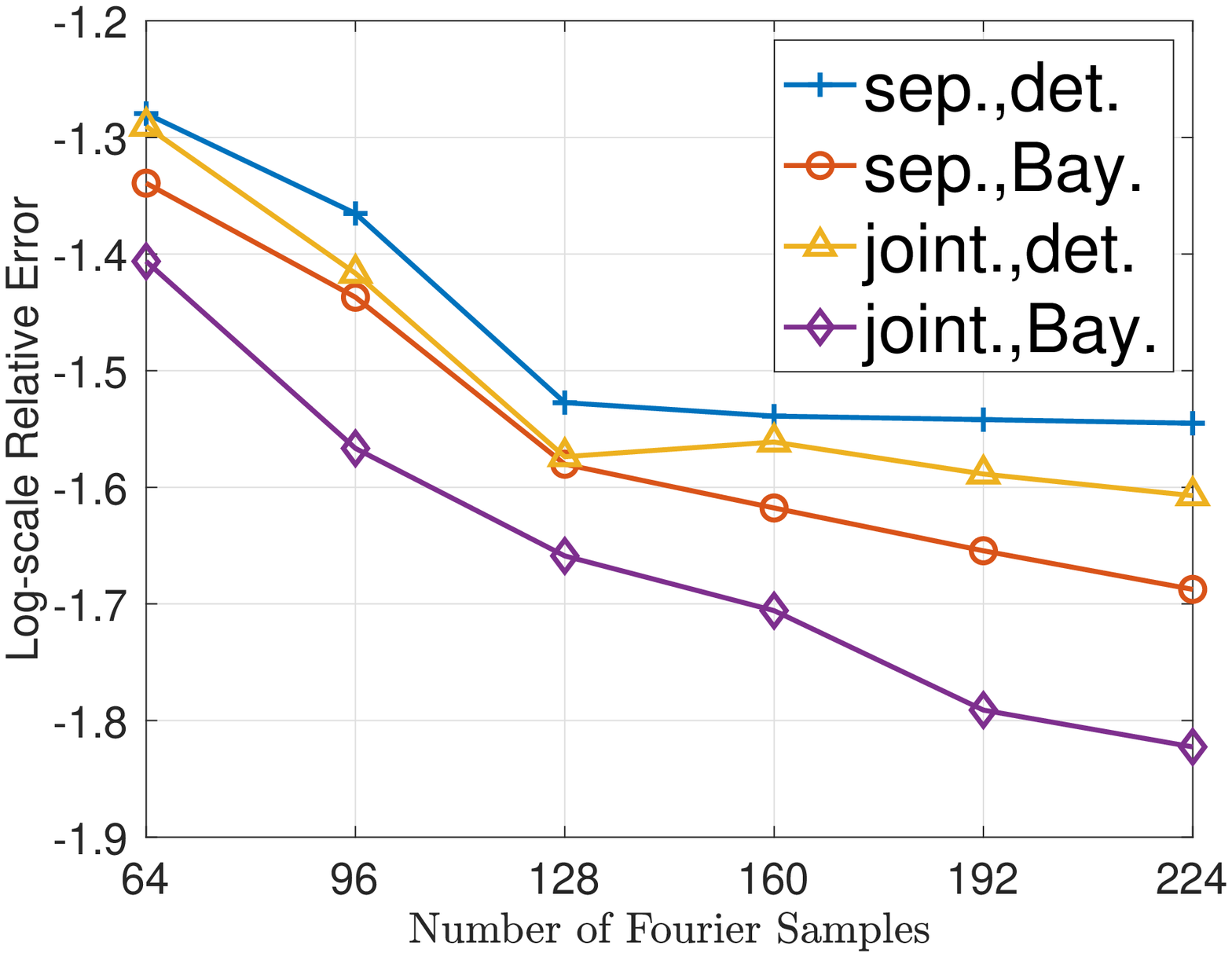} 
    \caption{Different numbers of Fourier samples}
    \label{fig:MRI_conv_samples}
    \end{subfigure}
    \begin{subfigure}[b]{.45\textwidth}
    \includegraphics[width=\textwidth]{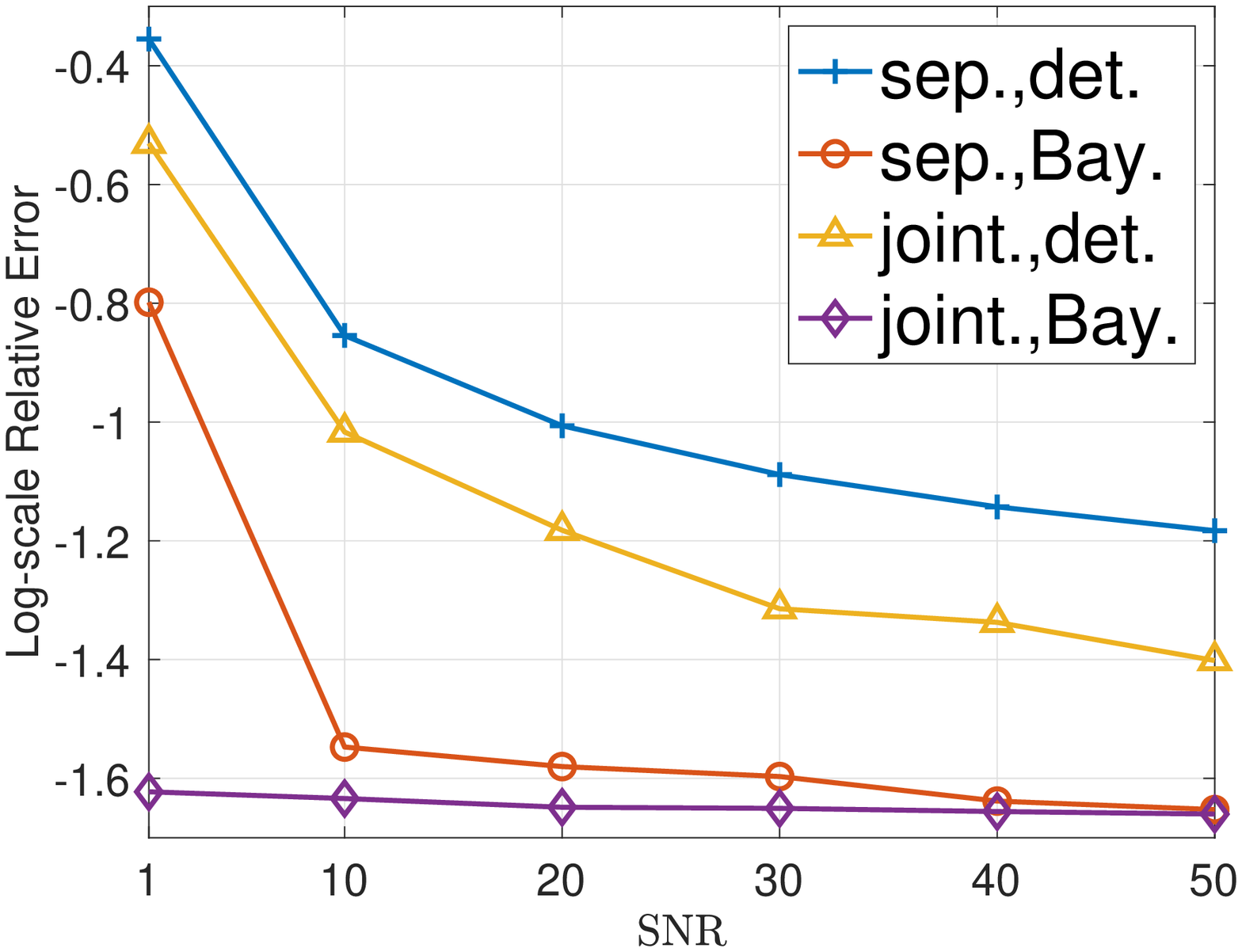} 
    \caption{Different SNRs}
    \label{fig:MRI_conv_SNR}
    \end{subfigure}
    \caption{
    Average relative log-errors for the separate deterministic (blue strokes), the separate Bayesian (orange circles), the joint deterministic (yellow triangles), and our joint Bayesian (purple diamonds) algorithm. 
    In Figure \ref{fig:MRI_conv_samples}, the SNR is $2$. 
    In Figure \ref{fig:MRI_conv_SNR}, we started with $128 \times 128$ Fourier samples and then removed the bands in \eqref{eq:MRI_bands_removed}. 
    }
    \label{fig:MRI_conv}
\end{figure}

We observed the proposed JHBL method to yield the most accurate recovered images for all combinations of Fourier samples and SNRs we considered. Figures \ref{fig:MRI_conv_samples} and \ref{fig:MRI_conv_SNR} show the average relative log-error for different numbers of Fourier samples and SNRs. 
The relative log-error of the $j$th image is 
\begin{equation}
    E_{\log}^{(j)} = \log_{10}\left( \frac{\| \mathbf{x}^{(j)}_{\rm ref} - \mathbf{x}^{(j)} \|_2}{\|\mathbf{x}^{(j)}_{\rm ref}\|_2} \right), 
    \quad j=1,\dots,J,
\end{equation}
where $\mathbf{x}^{(j)}_{\rm ref}$ and $\mathbf{x}^{(j)}$ are the reference and recovered image, respectively. 
The average relative log-error is $(E_{\log}^{(1)}+\dots+E_{\log}^{(J)})/J$.

\begin{figure}[tb]
    \centering
    \begin{subfigure}[b]{.45\textwidth}
    \includegraphics[width=\textwidth]{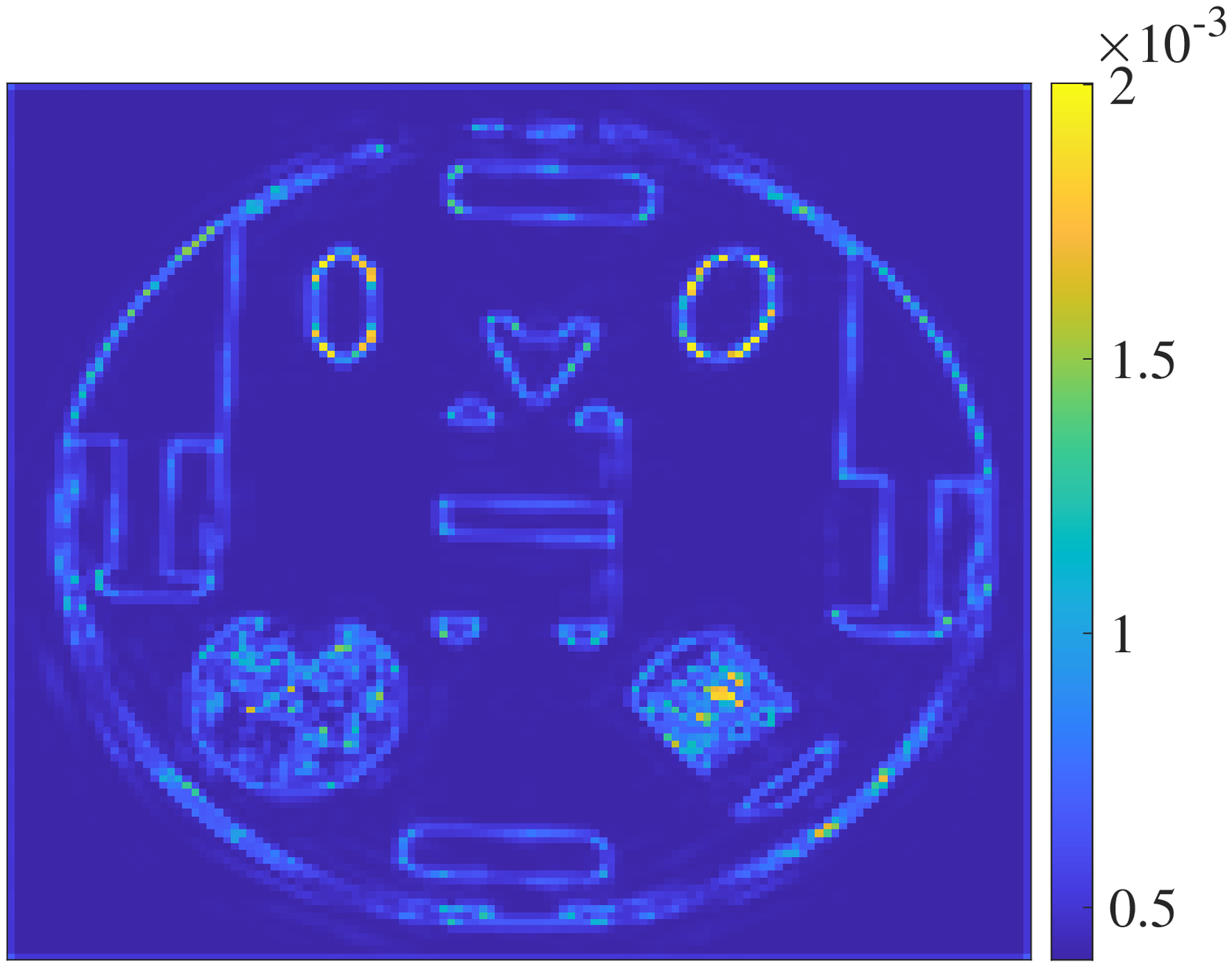}
    \caption{Separately recovered}
    \label{fig:MRI_UQ_sep}
    \end{subfigure}
    \begin{subfigure}[b]{.45\textwidth}
    \includegraphics[width=\textwidth]{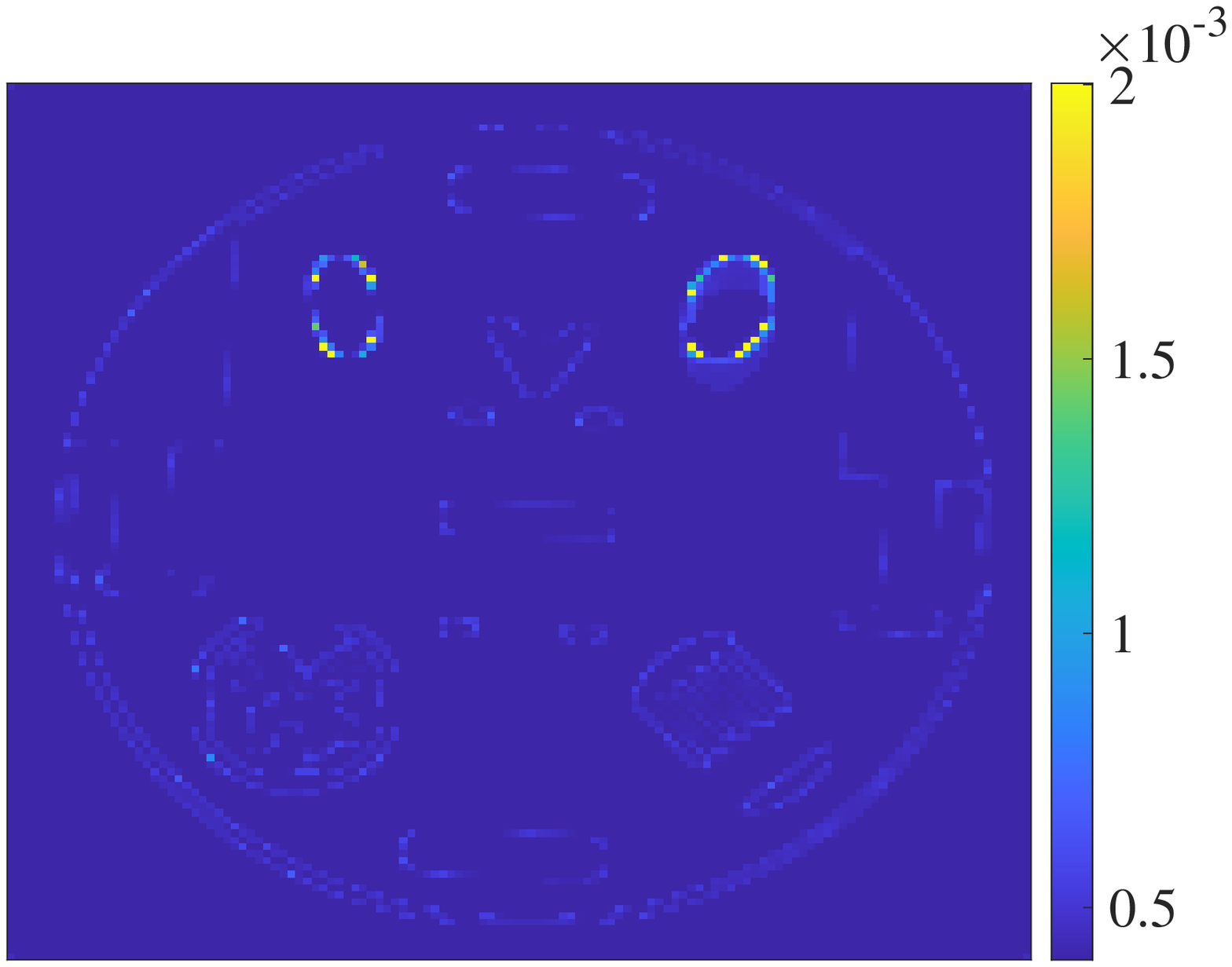}
    \caption{Jointly recovered}
    \label{fig:MRI_UQ_joint}
    \end{subfigure}
    \caption{
    Pixelwise variances of the recovered first image. 
    Jointly recovering the images by ''borrowing" missing information from the other images reduces uncertainty in no-change regions. 
    }
    \label{fig:MRI_UQ}
\end{figure}

Another advantage of our JHBL method is that it allows us to quantify uncertainty. 
Indeed, we recover a full Gaussian distribution, $\mathcal{N}(\boldsymbol{\mu}^{(j)}, \Sigma^{(j)})$, for every individual image, which is conditioned on the observed data sets and the other estimated parameters. 
To demonstrate this, Figures \ref{fig:MRI_UQ_sep} and \ref{fig:MRI_UQ_joint} illustrate the pixelwise variance of the separately and jointly recovered first image using the GSBL and our JHBL algorithm. 
The pixelwise variance corresponds to the diagonal elements of the covariance matrices \eqref{eq:covariances}. 
Comparing Figures \ref{fig:MRI_UQ_sep} and \ref{fig:MRI_UQ_joint}, we observe reduced uncertainty in the jointly recovered first image, except for the change regions around the two ellipses. 
The reduced uncertainty of the recovered first image by our JHBL method away from the change regions is due to the method ``borrowing" information from the neighboring images. 

\begin{figure}[tb]
    \centering
    \begin{subfigure}[t]{.32\textwidth}
    \includegraphics[width=\textwidth]{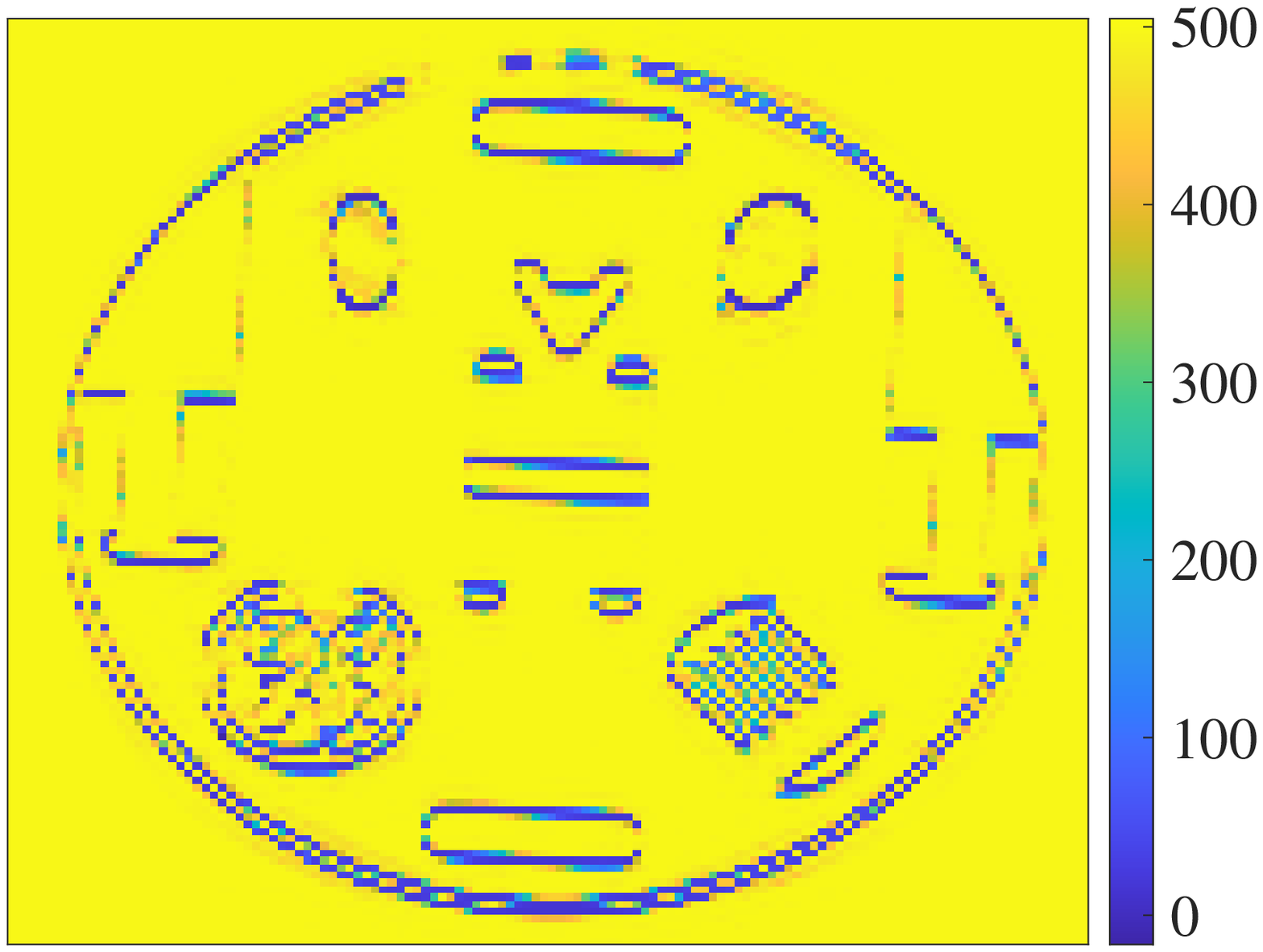}
    \caption{Vertical direction} 
    \label{fig:MRI_edges_vertical}
    \end{subfigure}
    \begin{subfigure}[t]{.32\textwidth}
    \includegraphics[width=\textwidth]{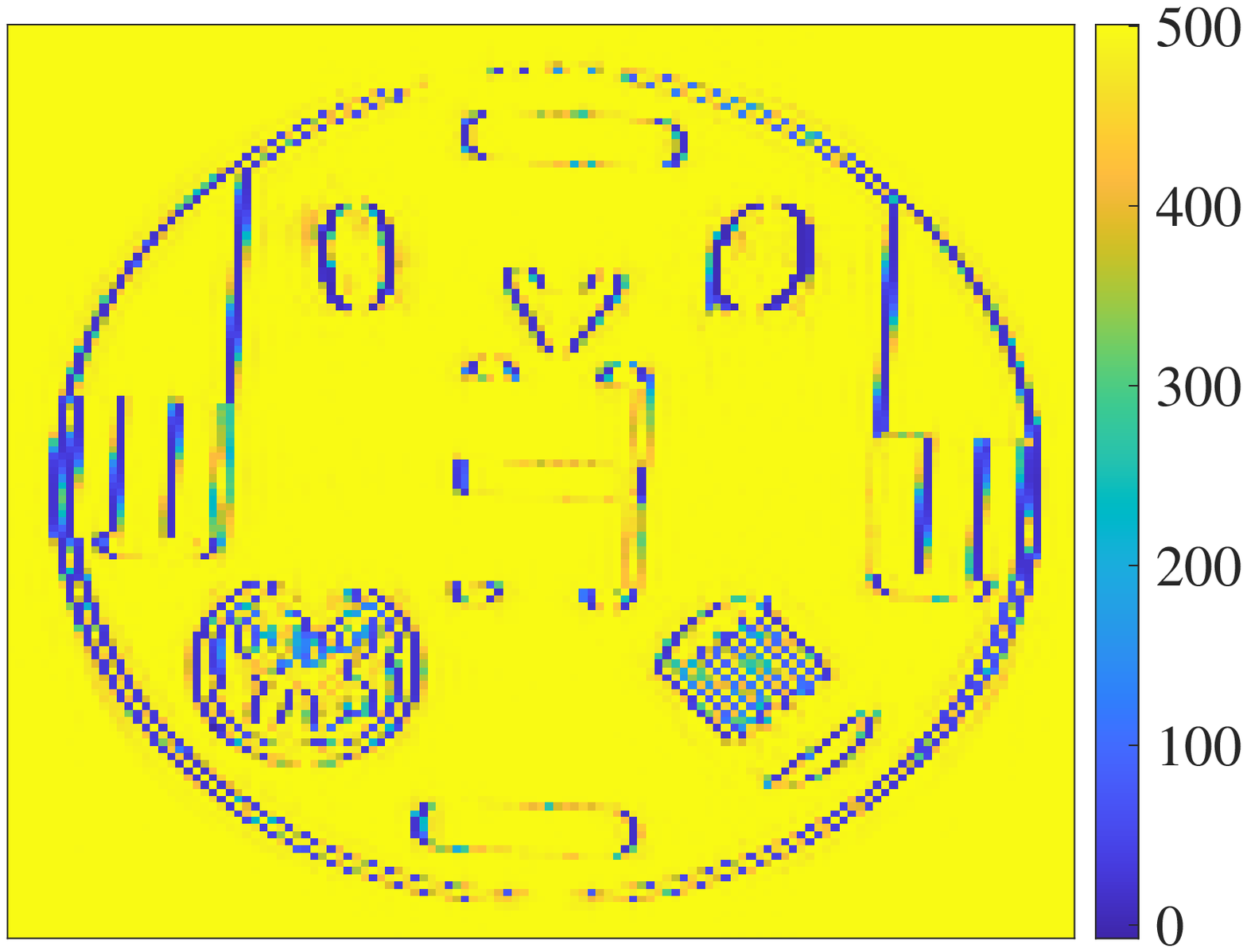}
    \caption{Horizontal direction} 
    \label{fig:MRI_edges_horizontal}
    \end{subfigure}
    \begin{subfigure}[t]{.32\textwidth}
    \includegraphics[width=\textwidth]{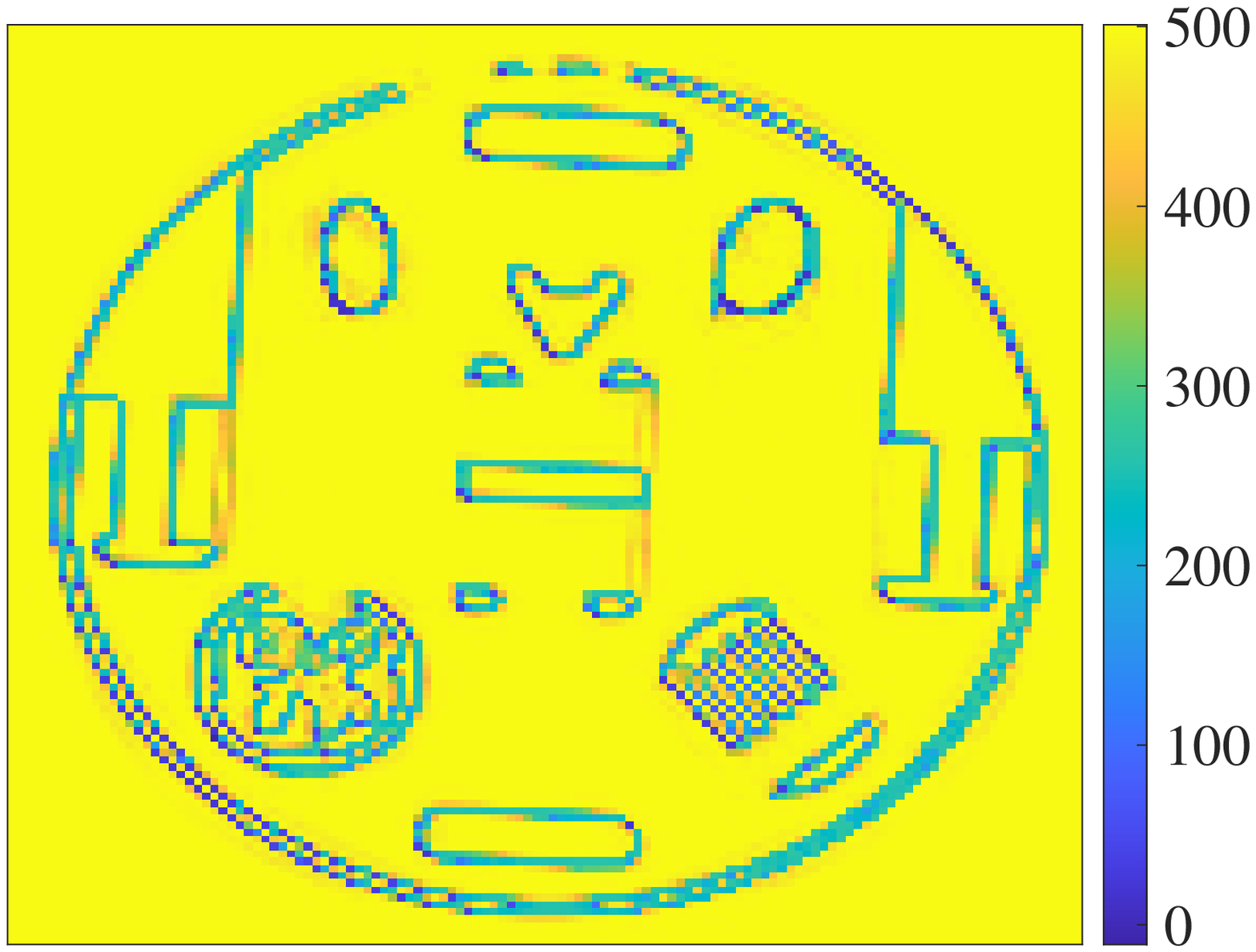}
    \caption{Pixelwise average value} 
    \label{fig:MRI_edges_combined}
    \end{subfigure}
    \caption{
    The final estimates for the intra-image regularization parameters in the vertical and horizontal direction for the first image in Figure \ref{fig:MRI_joint} and their pixelwise average 
    }
    \label{fig:MRI_edges}
\end{figure}

\begin{figure}[tb]
    \centering 
    \begin{subfigure}[b]{.45\textwidth}
    \includegraphics[width=\textwidth]{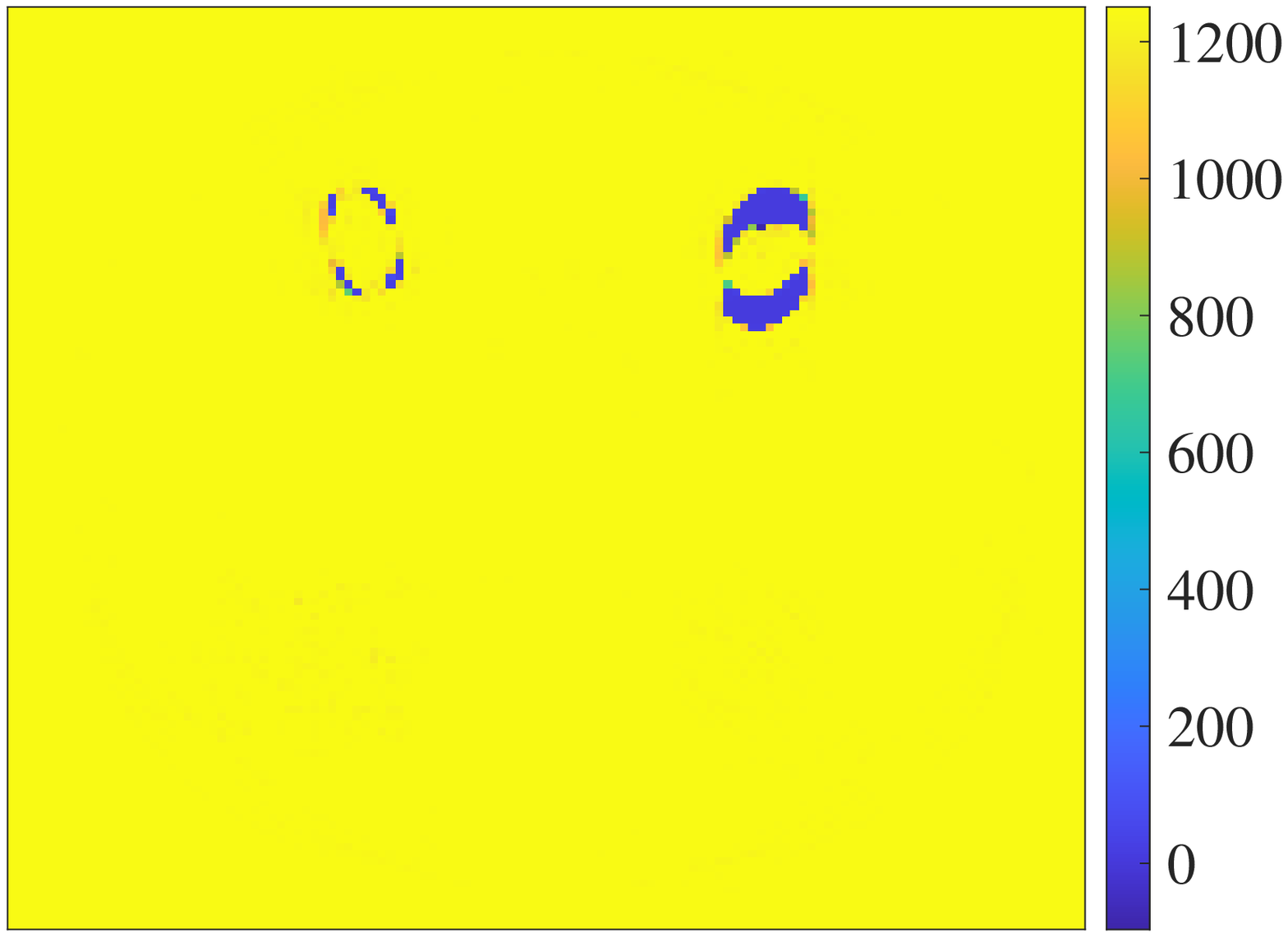}
    \caption{$1$st Bayesian change mask $C^{(1,2)}$} 
    \label{fig:MRI_change_Bay1}
    \end{subfigure}
    \begin{subfigure}[b]{.45\textwidth}
    \includegraphics[width=\textwidth]{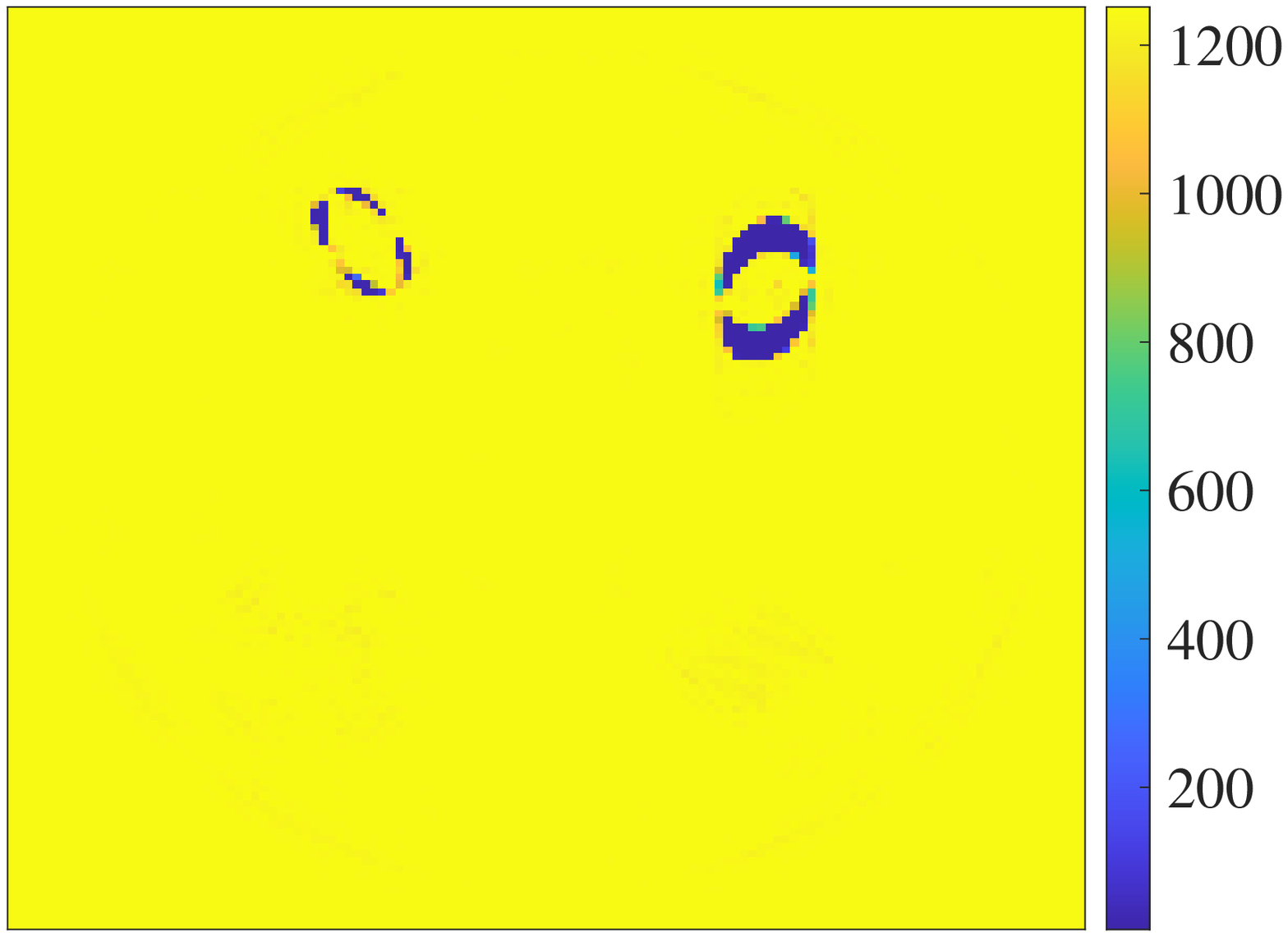}
    \caption{$2$nd Bayesian change mask $C^{(2,3)}$}
    \label{fig:MRI_change_Bay2}
    \end{subfigure}
    \\ 
    \begin{subfigure}[b]{.45\textwidth}
    \includegraphics[width=\textwidth]{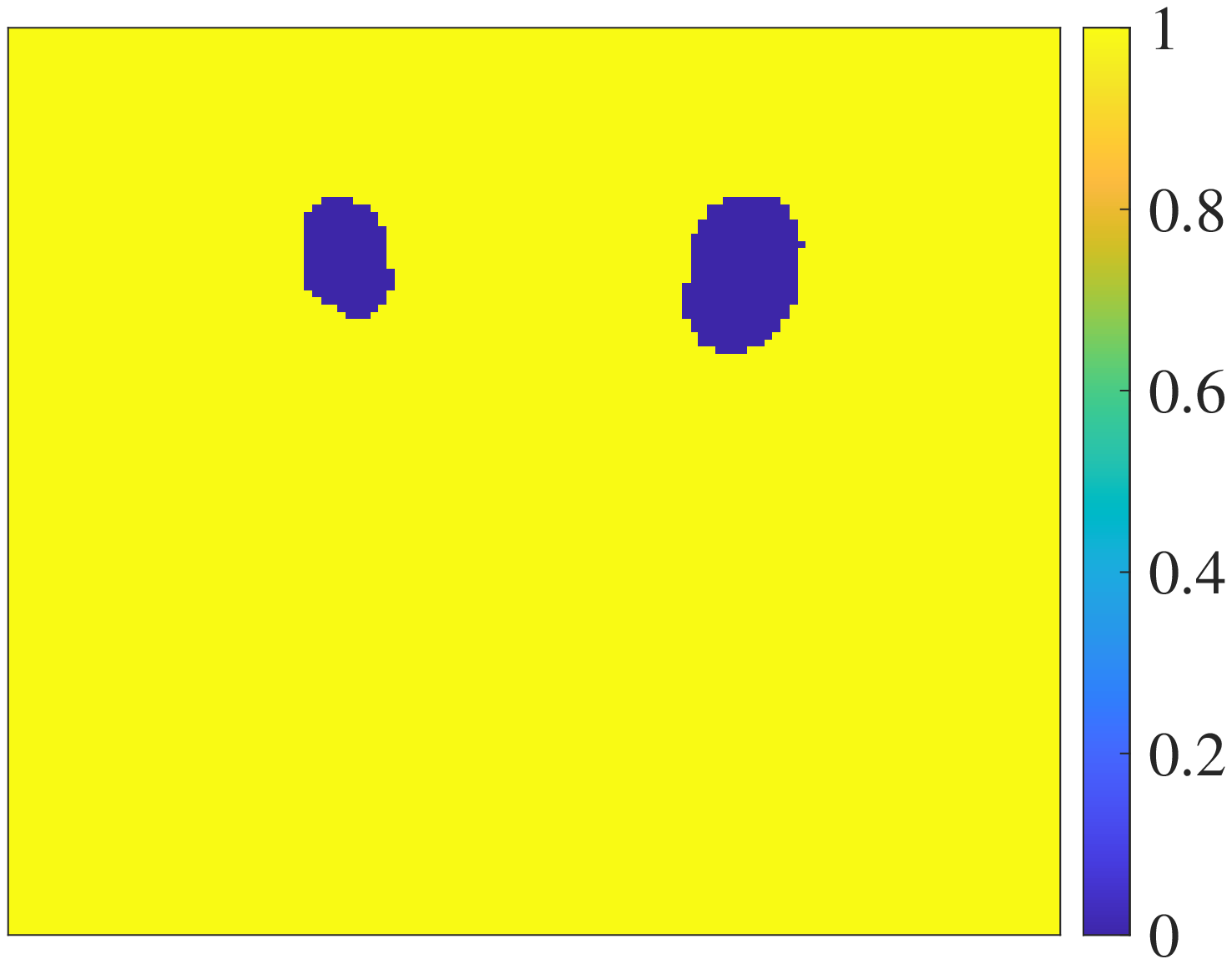}
    \caption{$1$st pre-computed binary change mask $C^{(1,2)}$}
    \label{fig:MRI_change_det1}
    \end{subfigure}
    \begin{subfigure}[b]{.45\textwidth}
    \includegraphics[width=\textwidth]{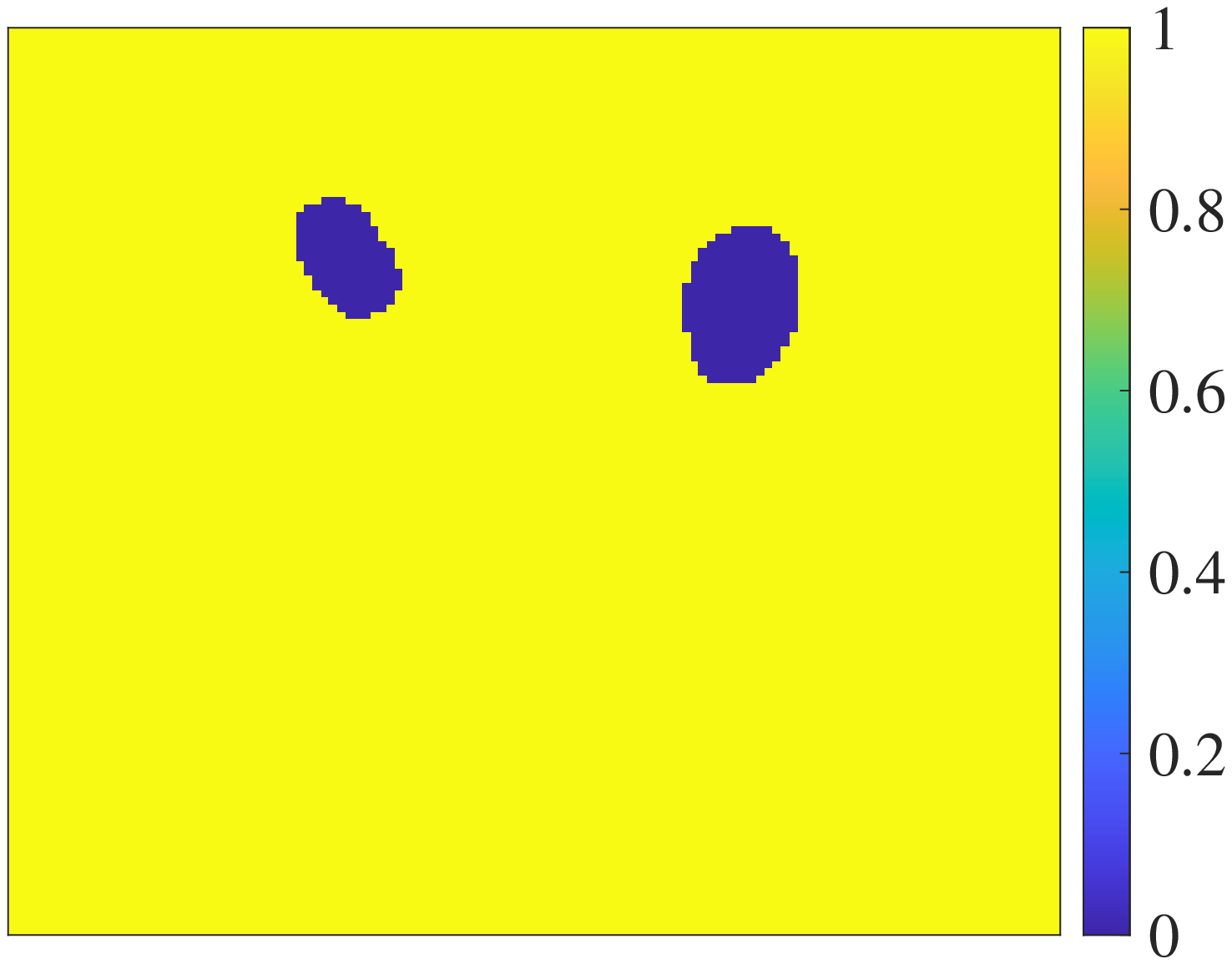}
    \caption{$2$nd pre-computed binary change mask $C^{(2,3)}$}
    \label{fig:MRI_change_det2}
    \end{subfigure}
    \caption{
    The final estimates for the Bayesian change masks (top row) and pre-computed binary change masks (bottom row) used by the Bayesian and deterministic method to jointly recover the temporal image sequence in Figure \ref{fig:MRI_joint} 
    }
    \label{fig:MRI_change}
\end{figure}

Another convenient by-product of our JHBL method is that it allows for edge and change detection, which we illustrate in Figures \ref{fig:MRI_edges} and \ref{fig:MRI_change}.
Figures \ref{fig:MRI_edges_vertical} and \ref{fig:MRI_edges_horizontal} respectively visualize the final estimate of the first and second half of the intra-image regularization parameter $\boldsymbol{\beta}^{(1)}$ of our JHBL method for the first recovered image. 
Since we used an anisotropic first-order TV operator \eqref{eq:MRI_reg}, the intra-image regularization parameter values indicate edges in the vertical and the horizontal direction.
We can combine them by considering the pixelwise average of the images in Figures \ref{fig:MRI_edges_vertical} and \ref{fig:MRI_edges_horizontal} to obtain the edge profile in Figure \ref{fig:MRI_edges_combined}. 

Figures \ref{fig:MRI_change_Bay1} and \ref{fig:MRI_change_Bay2} visualize the final estimate of the conditional change mask used by our JHBL method for the first and second pair of images, while Figures \ref{fig:MRI_change_det1} and \ref{fig:MRI_change_det2} illustrates the corresponding pre-computed binary change masks used in \cite{xiao2022sequential}. 
The pre-computed binary change masks rely on the Fourier data sets and sequential images containing only objects with closed boundaries. 
By contrast, the change masks used in our JHBL method neither rely on Fourier data nor the sequential images containing only objects with closed boundaries.

\subsection{Sequential image deblurring} 
\label{sub:tests_deblur}

\begin{figure}[tb]
    \centering
    \begin{subfigure}[b]{.32\textwidth}
    \includegraphics[width=\textwidth]{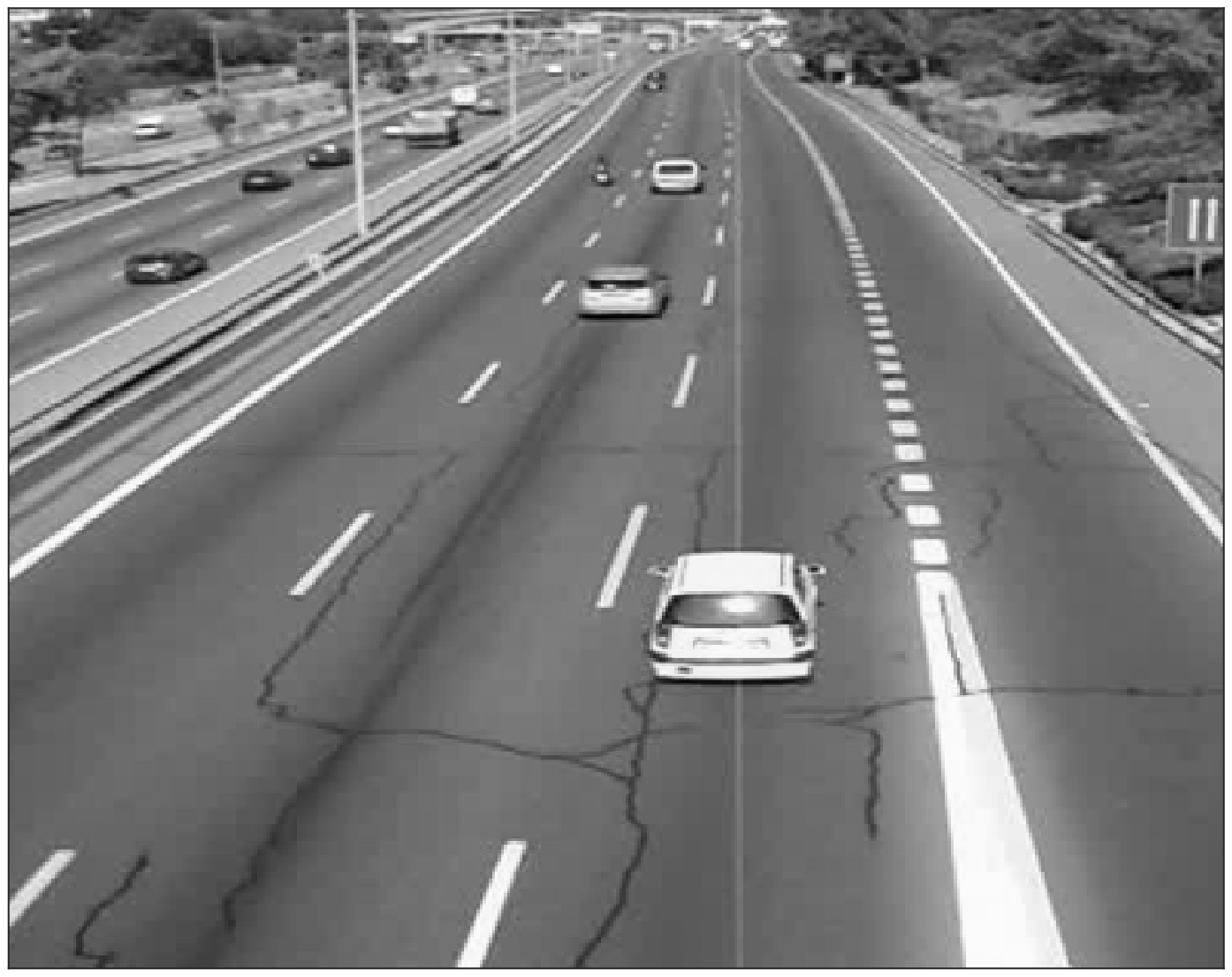}
    \caption{Reference image $1$}
    \label{fig:deblur_images_ref1}
    \end{subfigure}
    \begin{subfigure}[b]{.32\textwidth}
    \includegraphics[width=\textwidth]{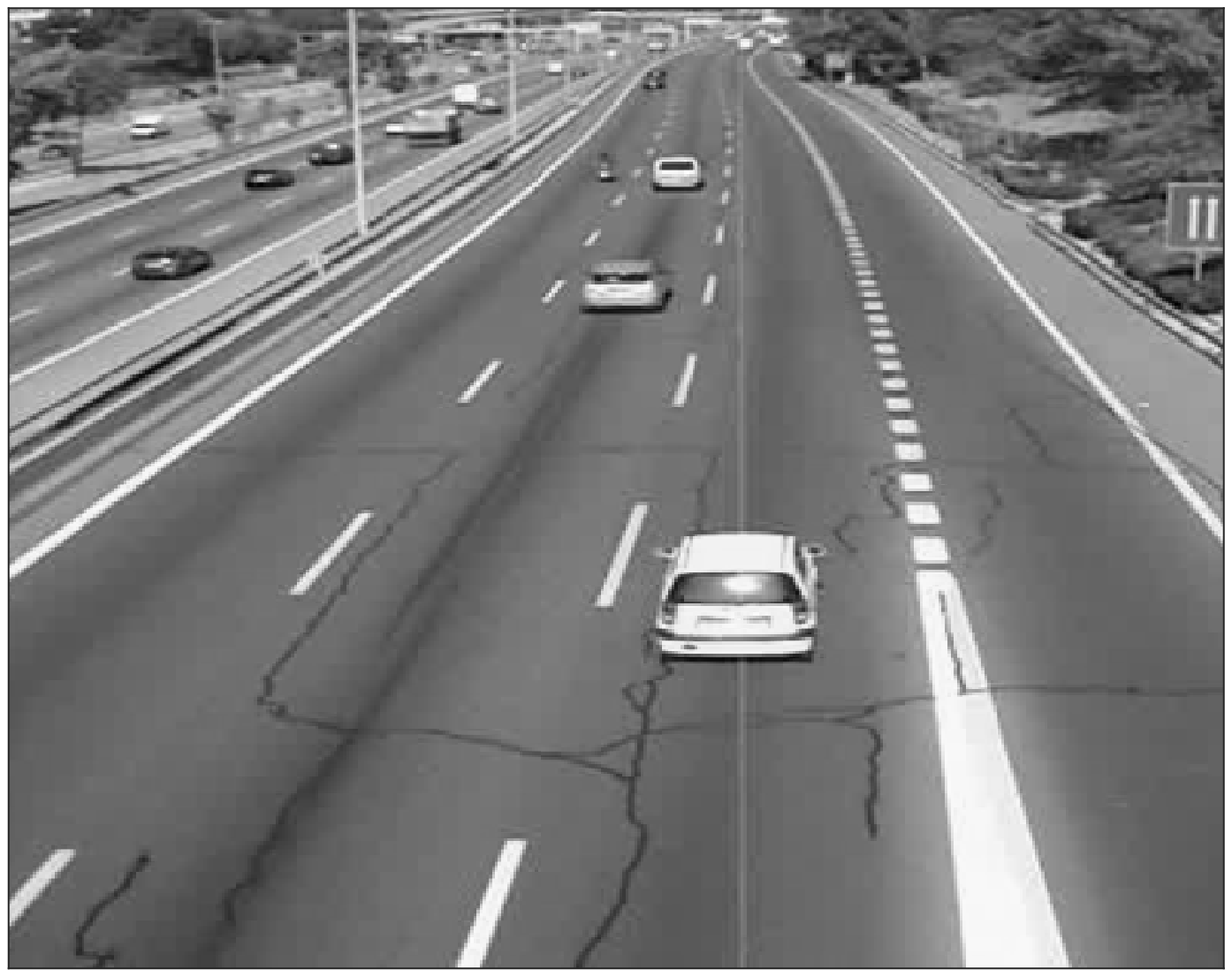}
    \caption{Reference image $2$}
    \label{fig:deblur_images_ref2}
    \end{subfigure}
    \begin{subfigure}[b]{.32\textwidth}
    \includegraphics[width=\textwidth]{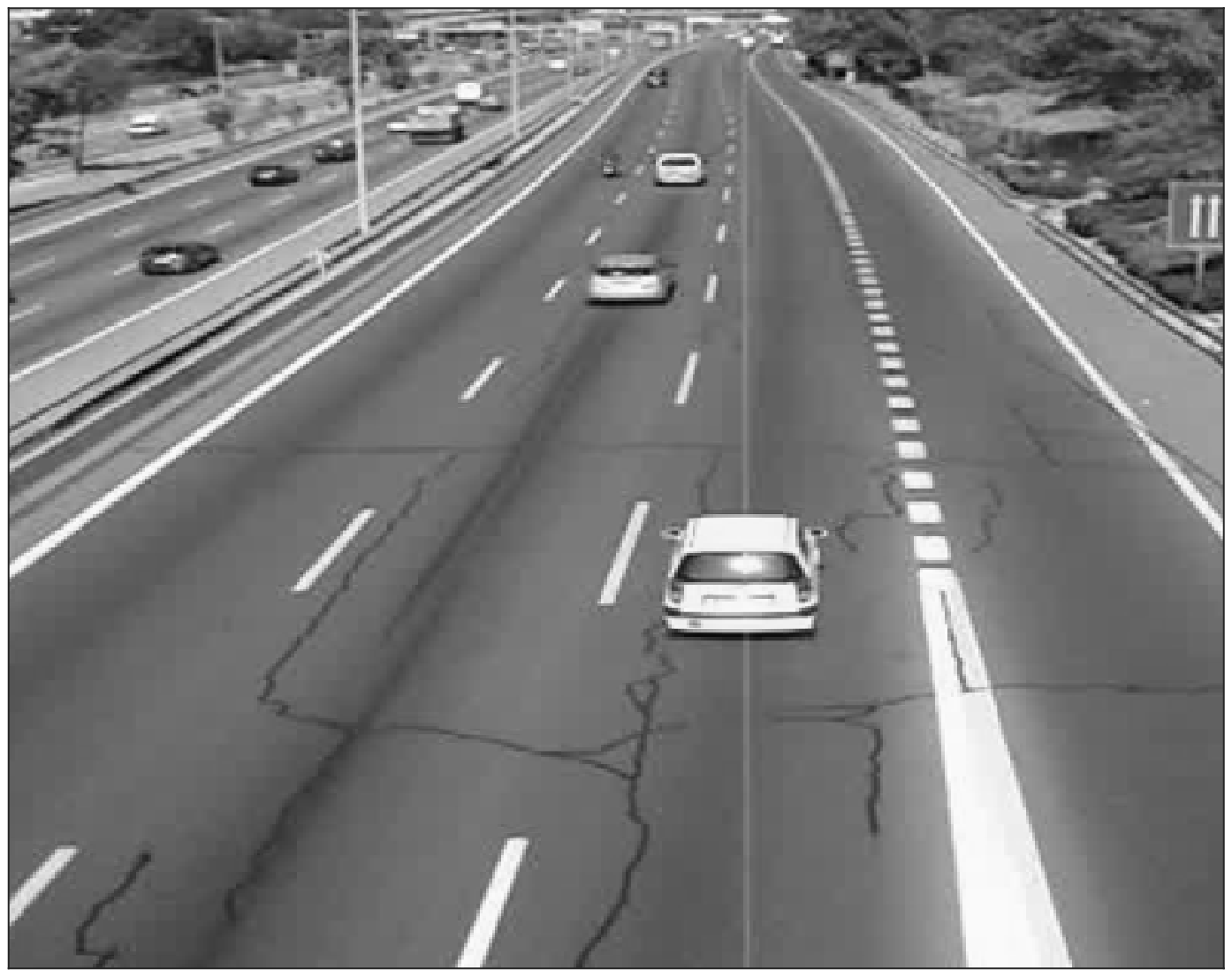}
    \caption{Reference image $3$}
    \label{fig:deblur_images_ref3}
    \end{subfigure}
    \\
    \begin{subfigure}[b]{.32\textwidth}
    \includegraphics[width=\textwidth]{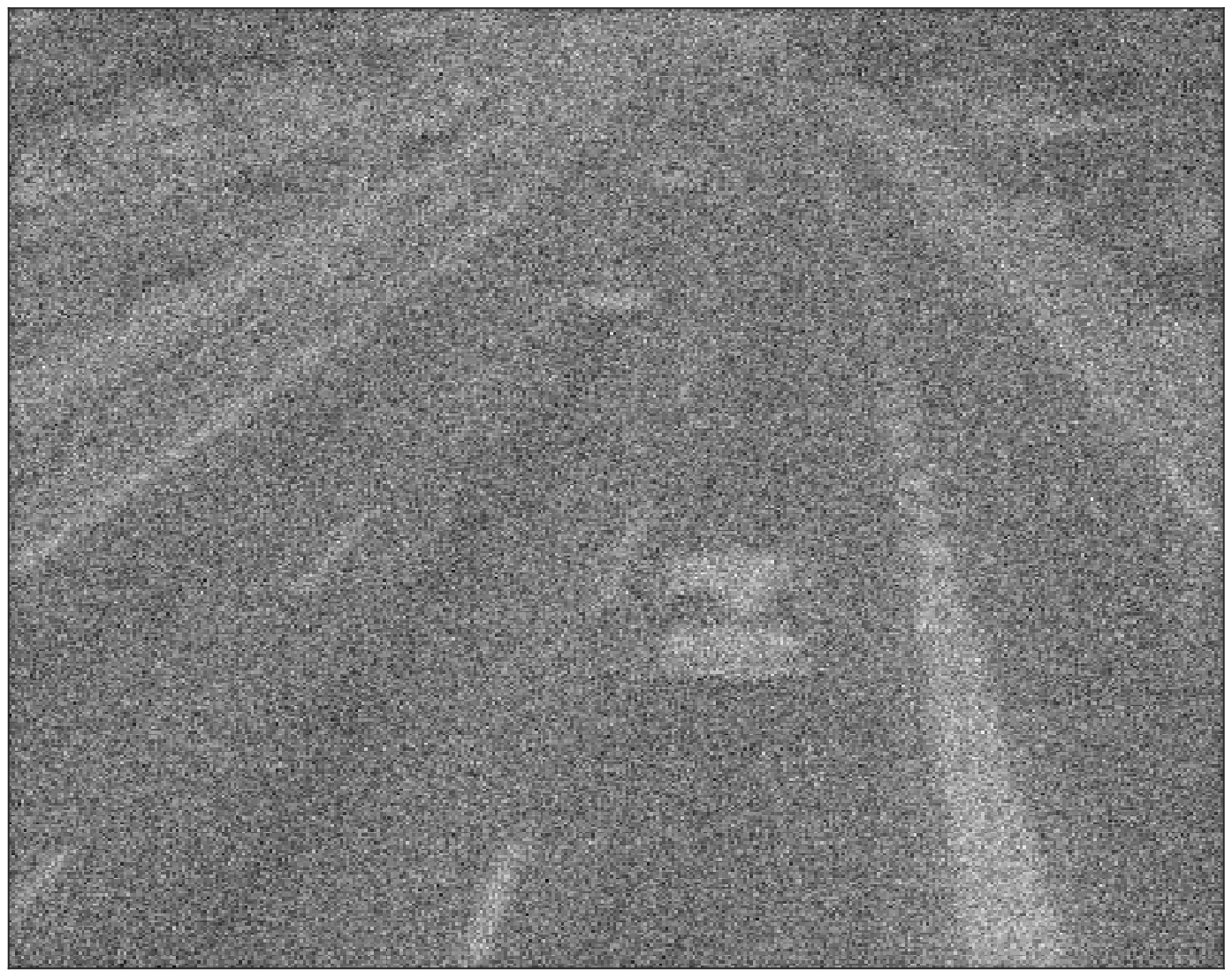}
    \caption{Noisy blurred image $1$}
    \label{fig:deblur_images_blurred1}
    \end{subfigure}
    \begin{subfigure}[b]{.32\textwidth}
    \includegraphics[width=\textwidth]{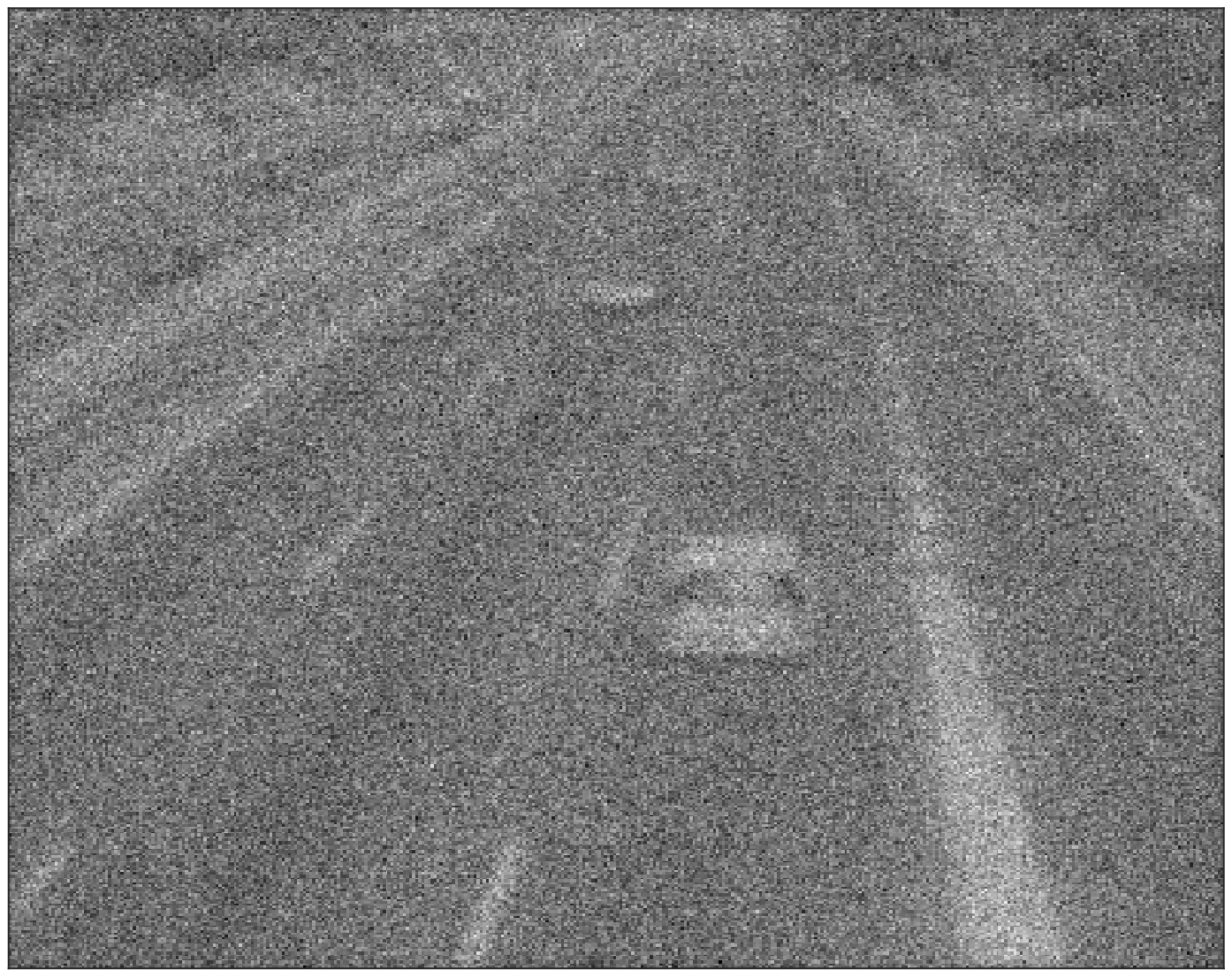}
    \caption{Noisy blurred image $2$}
    \label{fig:deblur_images_blurred2}
    \end{subfigure}
    \begin{subfigure}[b]{.32\textwidth}
    \includegraphics[width=\textwidth]{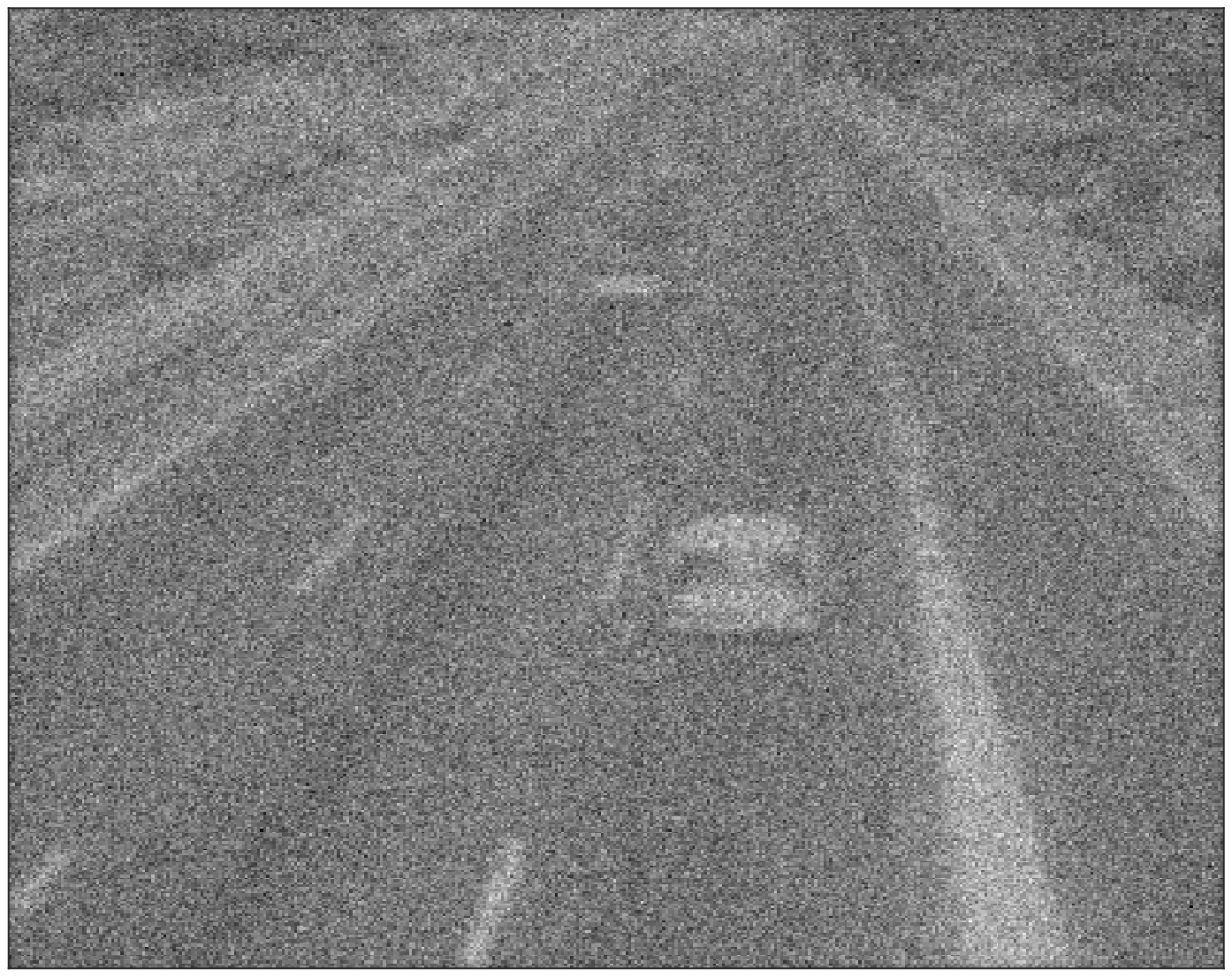}
    \caption{Noisy blurred image $3$}
    \label{fig:deblur_images_blurred3}
    \end{subfigure}
    \caption{The first three reference images of a temporal image sequence coming from the GRAM road-traffic monitoring data set \cite{guerrero2013iwinac} (first row) and their noisy blurred versions (second row)
    }
    \label{fig:deblur_images}
\end{figure}

We next consider the deconvolution of a temporal sequence of six $400 \times 400$ images from the GRAM road-traffic monitoring data set \cite{guerrero2013iwinac}.
Figure \ref{fig:deblur_images} illustrates the first three reference images and their noisy blurred versions. 
The i.\,i.\,d.\ normal noise $\mathbf{e}^{(j)}$ in the corresponding linear data model \eqref{eq:data_model} has ${\rm SNR} = 2+j$, $j=1,\dots,J$.  
The forward operator $F$ is obtained by applying the tensor-product midpoint quadrature to the convolution equations 
\begin{equation} 
	y^{(j)}(s,t) = \int_0^1 \int_0^1 k(s-s',t-t') x^{(j)}(s,t) \intd s' \intd t', \quad j=1,\dots,J, 
\end{equation} 
where $x^{(j)}$ is a function describing the $j$th image.
We assume a Gaussian convolution kernel 
\begin{equation} 
	k(s,t)=\frac{1}{2\pi\gamma^2}\exp{\left(-\frac{s^2+t^2}{2\gamma^2}\right)}
\end{equation} 
with blurring parameter $\gamma = 5 \cdot 10^{-3}$, which makes $F$ highly ill-conditioned. 
We use an anisotropic second-order TV regularization operator 
\begin{equation}\label{eq:R_deconvolution}
    R = \begin{bmatrix} I \otimes D \\ D \otimes I \end{bmatrix} 
    \quad \text{with} \quad 
    D = 
    \begin{bmatrix}
        -1 & 2 & -1 & & \\ 
         & \ddots & \ddots & \ddots & \\ 
         & & -1 & 2 & -1
    \end{bmatrix} 
    \in \R^{(N_1-2) \times N_1}, 
\end{equation}
where $N_1 = 400$ is the number of pixels in each direction to promote the images being piecewise smooth---but not necessarily piecewise constant. 

\begin{figure}[tb]
    \centering
    \begin{subfigure}[b]{.32\textwidth}
    \includegraphics[width=\textwidth]{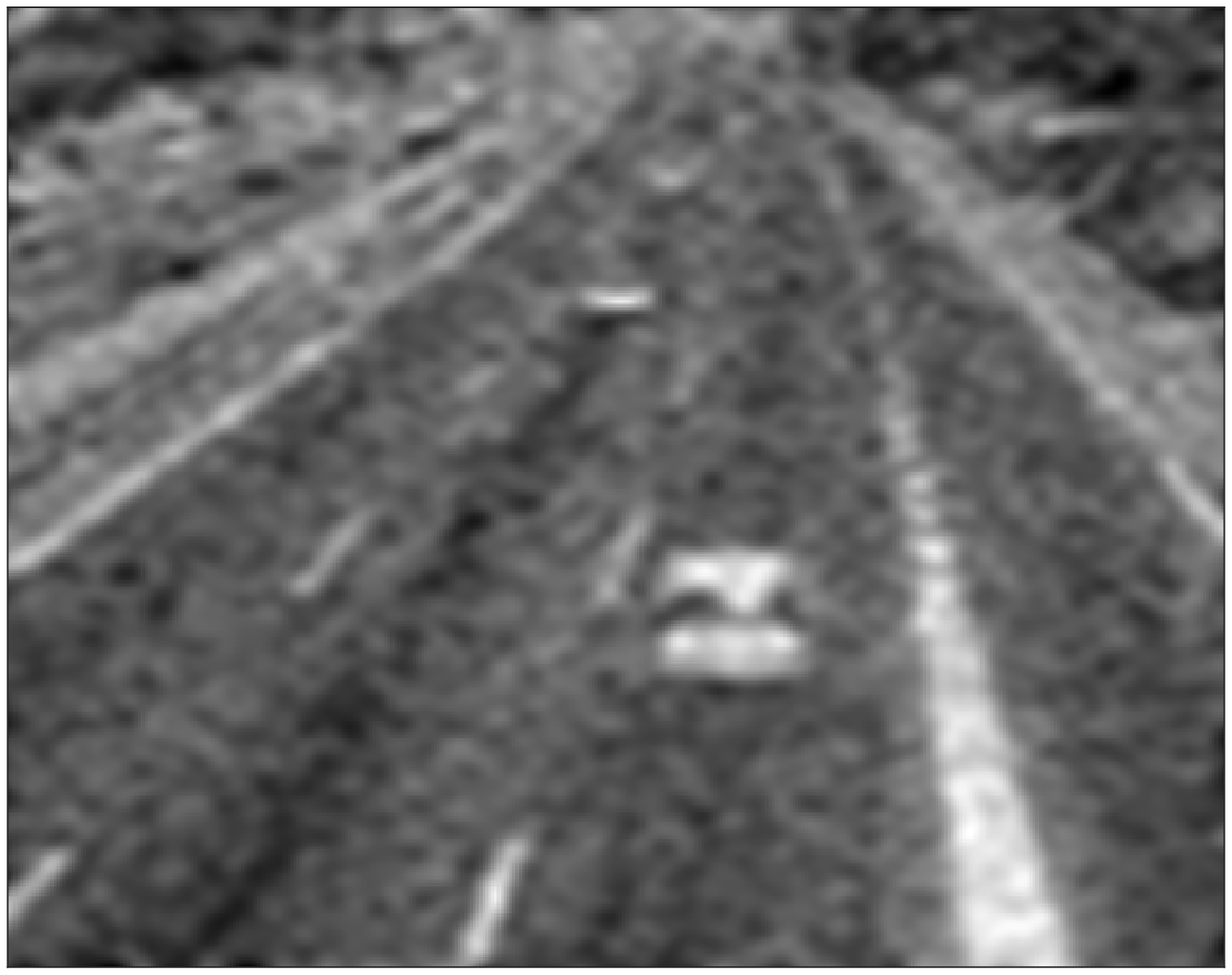}
    \caption{Separately recovered $1$} 
    \label{fig:deblur_recovered_sep1}
    \end{subfigure}
    \begin{subfigure}[b]{.32\textwidth}
    \includegraphics[width=\textwidth]{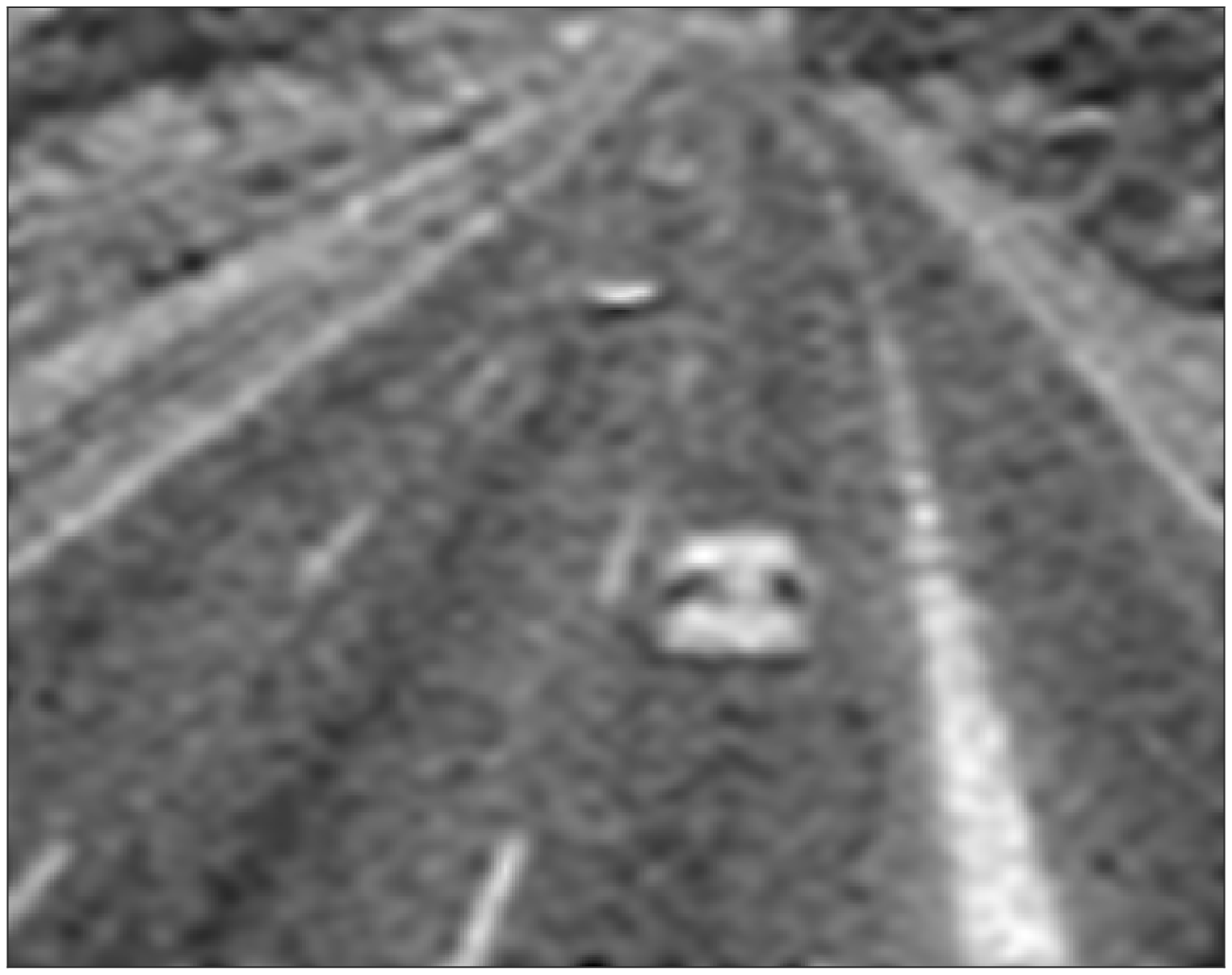}
    \caption{Separately recovered $2$}
    \label{fig:deblur_recovered_sep2}
    \end{subfigure}
    \begin{subfigure}[b]{.32\textwidth}
    \includegraphics[width=\textwidth]{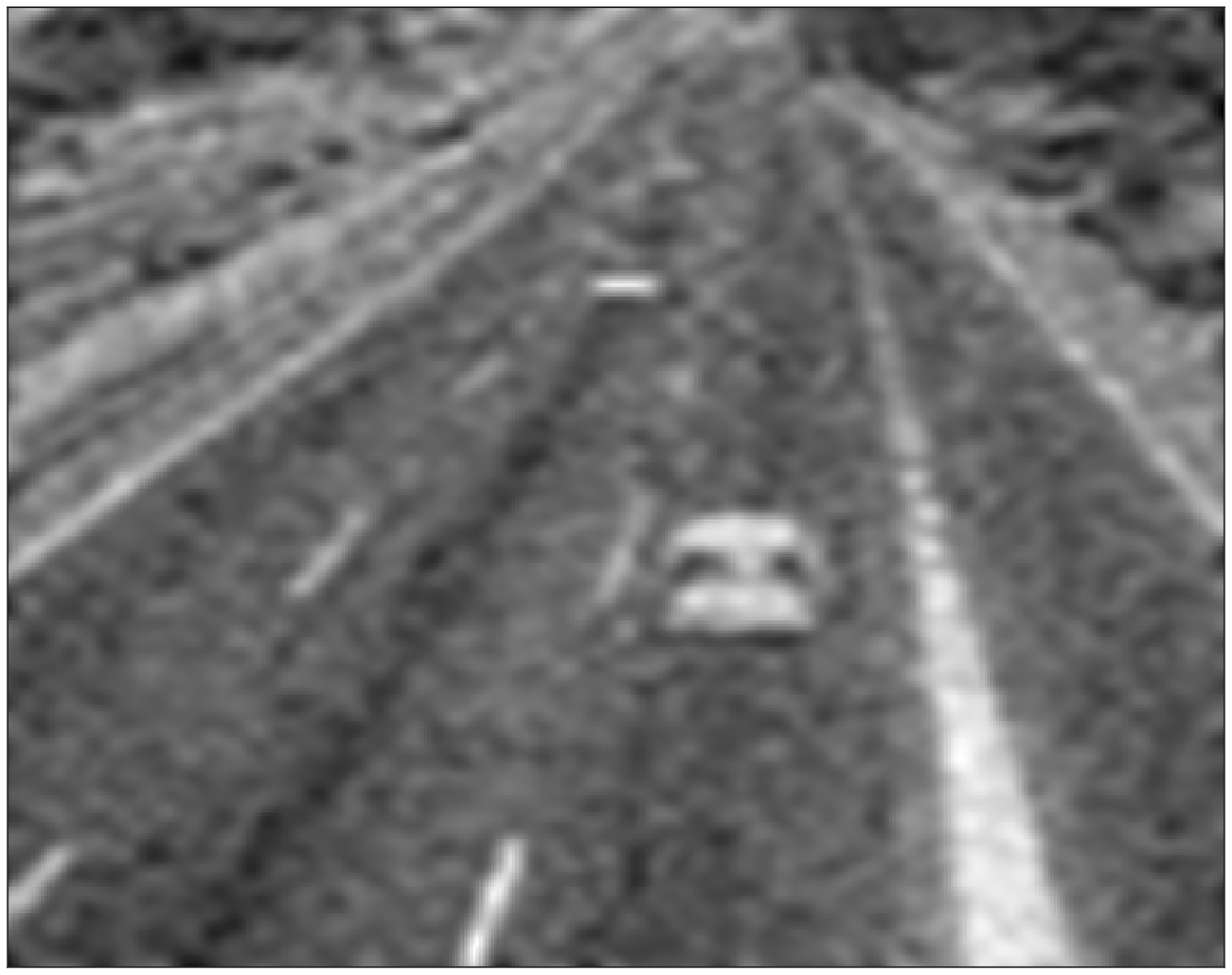}
    \caption{Separately recovered $3$}
    \label{fig:deblur_recovered_sep3}
    \end{subfigure}
    \\
    \begin{subfigure}[b]{.32\textwidth}
    \includegraphics[width=\textwidth]{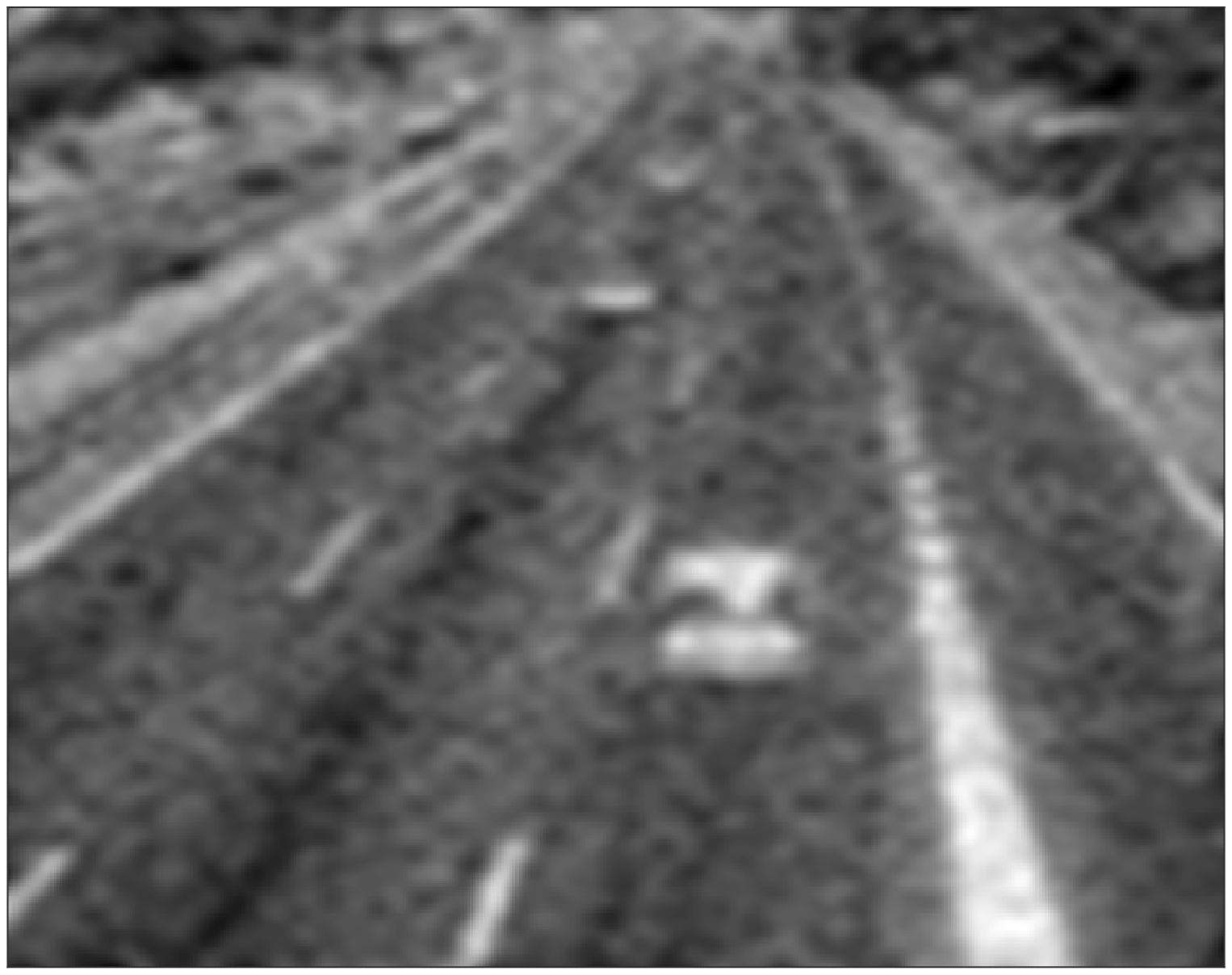}
    \caption{Jointly recovered $1$}
    \label{fig:deblur_recovered_joint1}
    \end{subfigure}
    \begin{subfigure}[b]{.32\textwidth}
    \includegraphics[width=\textwidth]{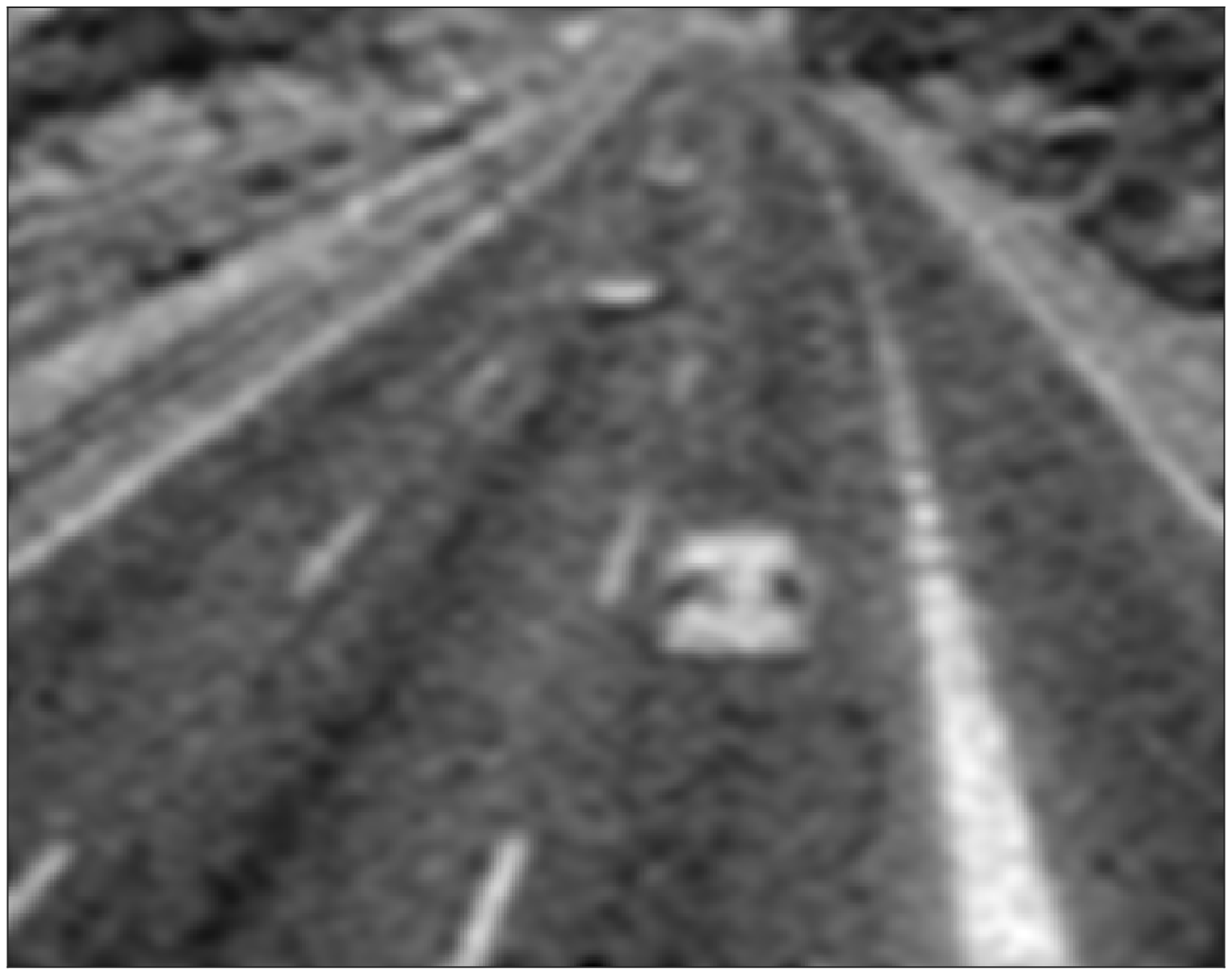}
    \caption{Jointly recovered $2$}
    \label{fig:deblur_recovered_joint2}
    \end{subfigure}
    \begin{subfigure}[b]{.32\textwidth}
    \includegraphics[width=\textwidth]{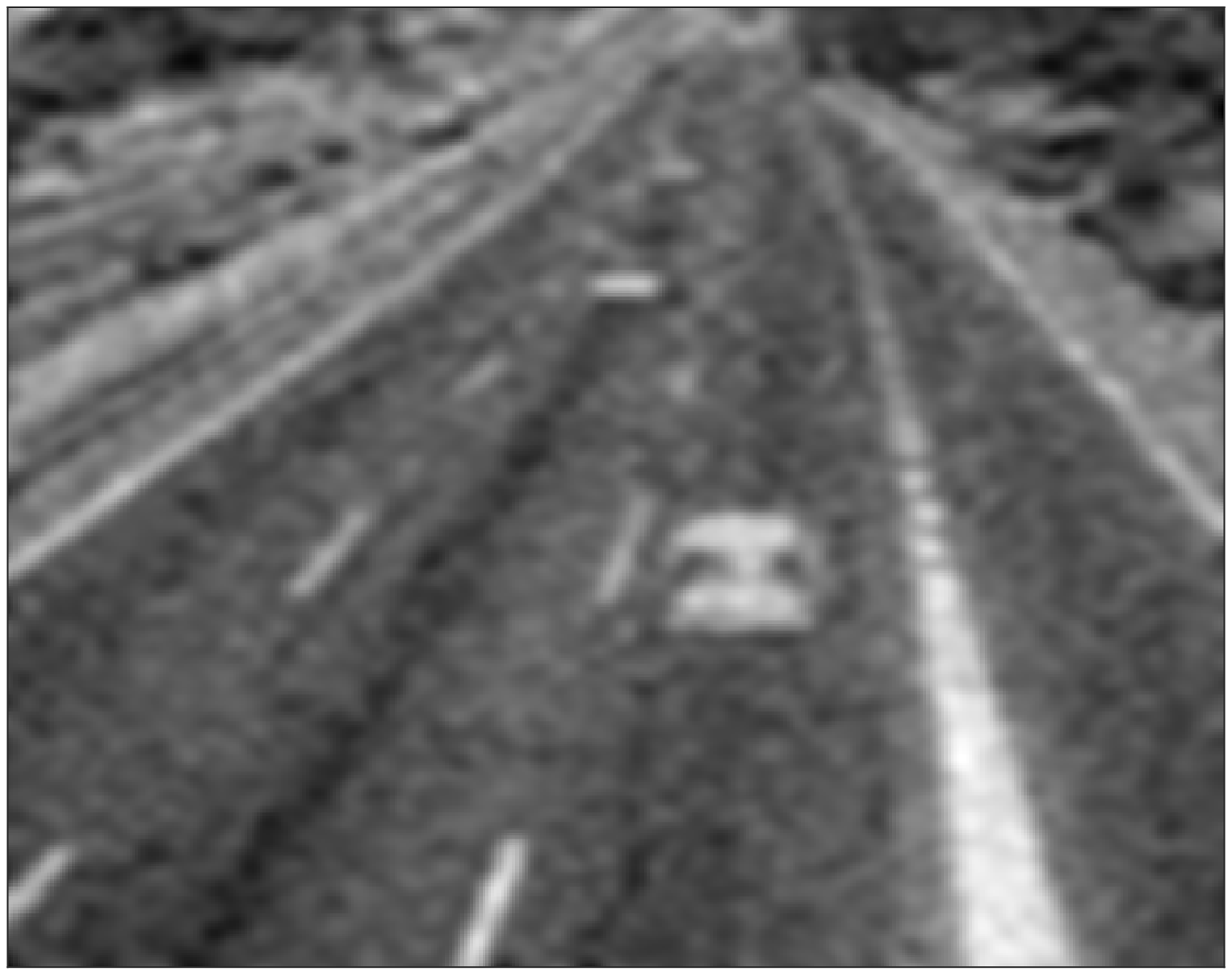}
    \caption{Jointly recovered $3$}
    \label{fig:deblur_recovered_joint3}
    \end{subfigure}
    \caption{ 
    Separately recovered images using GSBL (top row) and the jointly recovered images using our JHBL algorithm (bottom row) 
    }
    \label{fig:deblur_recovered}
\end{figure}

We can no longer use the deterministic method \cite{xiao2022sequential} to pre-compute binary change masks directly from the indirect data sets. 
However, we can still use our JHBL method to jointly recover the sequential images in a Bayesian setting. 
Figure \ref{fig:deblur_recovered} illustrates the separately (top row) and jointly (bottom row) recovered images from the noisy blurred images in Figures \ref{fig:deblur_images_blurred1}, \ref{fig:deblur_images_blurred2}, and \ref{fig:deblur_images_blurred3} using the GSBL and our JHBL algorithm, respectively. 
The hyper-parameters are $\eta_{\alpha} = \eta_{\beta} = 1$, $\eta_{\gamma} = 1$, $\theta_{\alpha} = \theta_{\beta} = 10^{-3}$, and $\theta_{\gamma}=10^{-1}$. 
The jointly recovered images using our JHBL algorithm are more accurate than the separately recovered images. 

\begin{figure}[tb]
    \centering
    \begin{subfigure}[b]{.45\textwidth}
    \includegraphics[width=\textwidth]{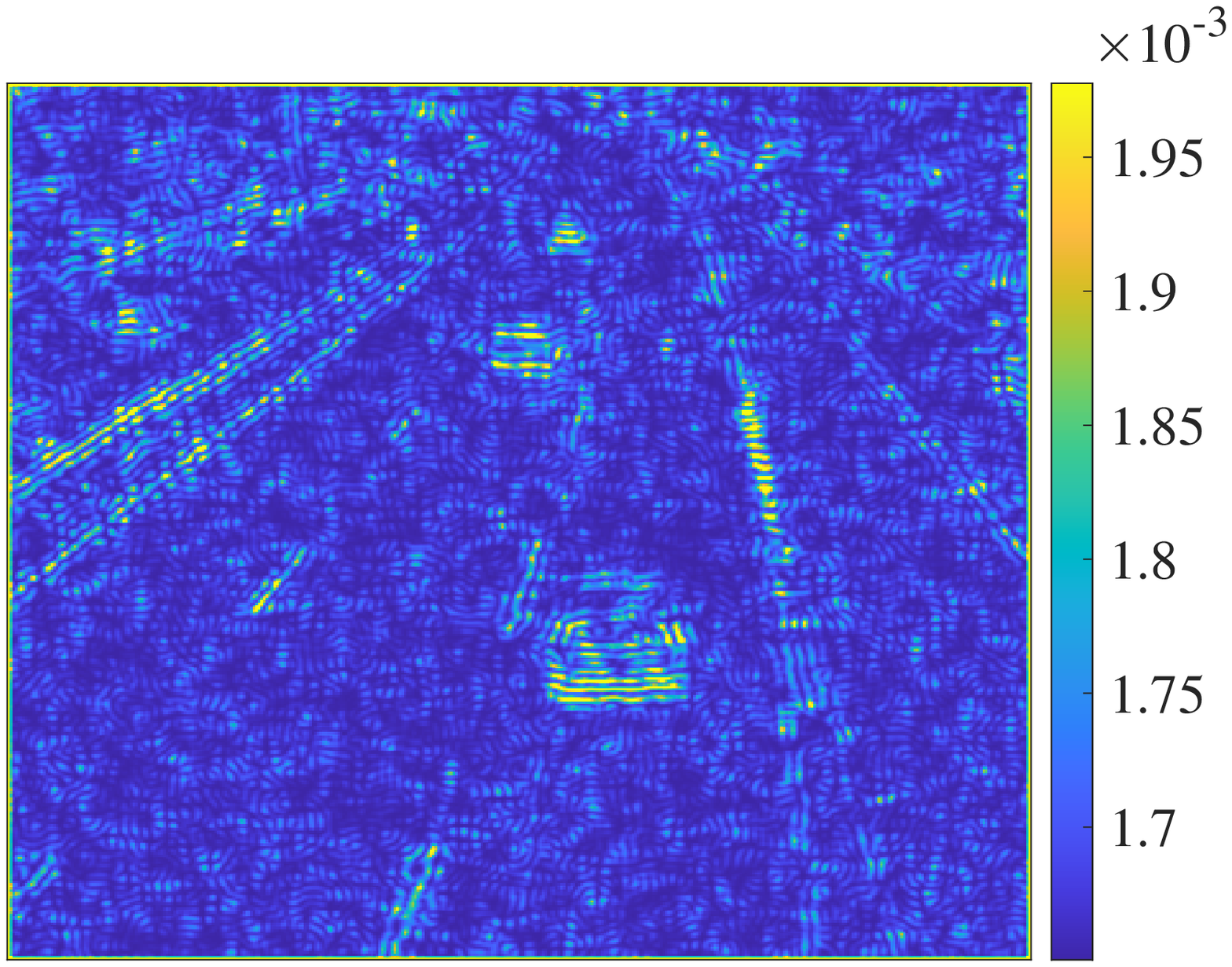}
    \caption{Separately recovered}
    \end{subfigure}
    \begin{subfigure}[b]{.45\textwidth}
    \includegraphics[width=\textwidth]{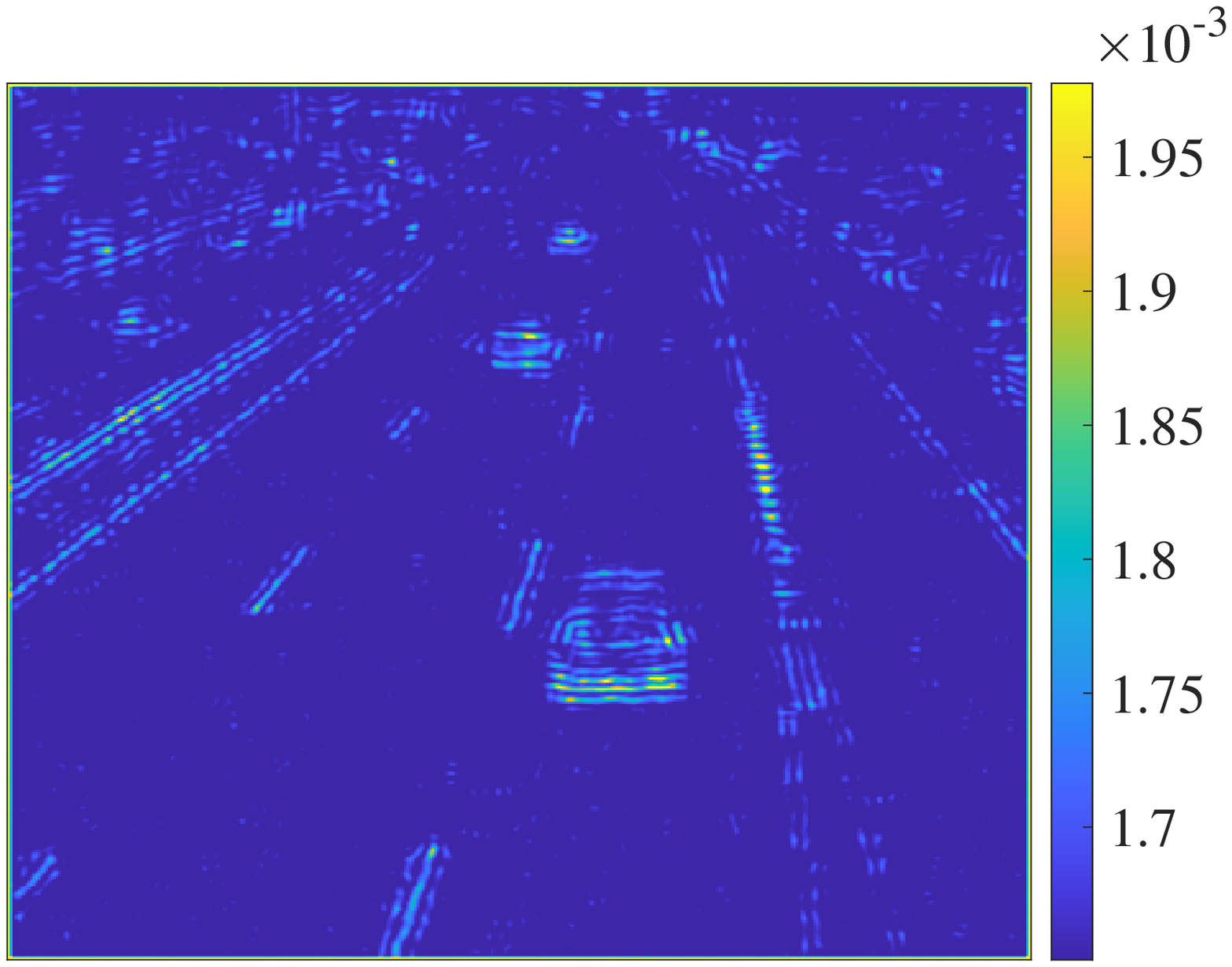}
    \caption{Jointly recovered}
    \end{subfigure}
    \caption{
    Pixelwise variances of the recovered first image in Figure \ref{fig:deblur_recovered}. 
    Jointly recovering the images by ``borrowing" missing information from the other images reduces uncertainty.
    }
    \label{fig:deblur_UQ}
\end{figure}

As mentioned, the proposed JHBL method can quantify uncertainty in the recovered images with increased reliability, which is often desirable in applications with no reference images. 
We demonstrate this in Figure \ref{fig:deblur_UQ}, which visualizes the pixelwise variances of the recovered first image in Figure \ref{fig:deblur_recovered}. 
We again see that jointly recovering the images by ``borrowing" missing information from the other images reduces uncertainty. 

\begin{figure}[tb]
    \centering
    \begin{subfigure}[t]{.32\textwidth}
    \includegraphics[width=\textwidth]{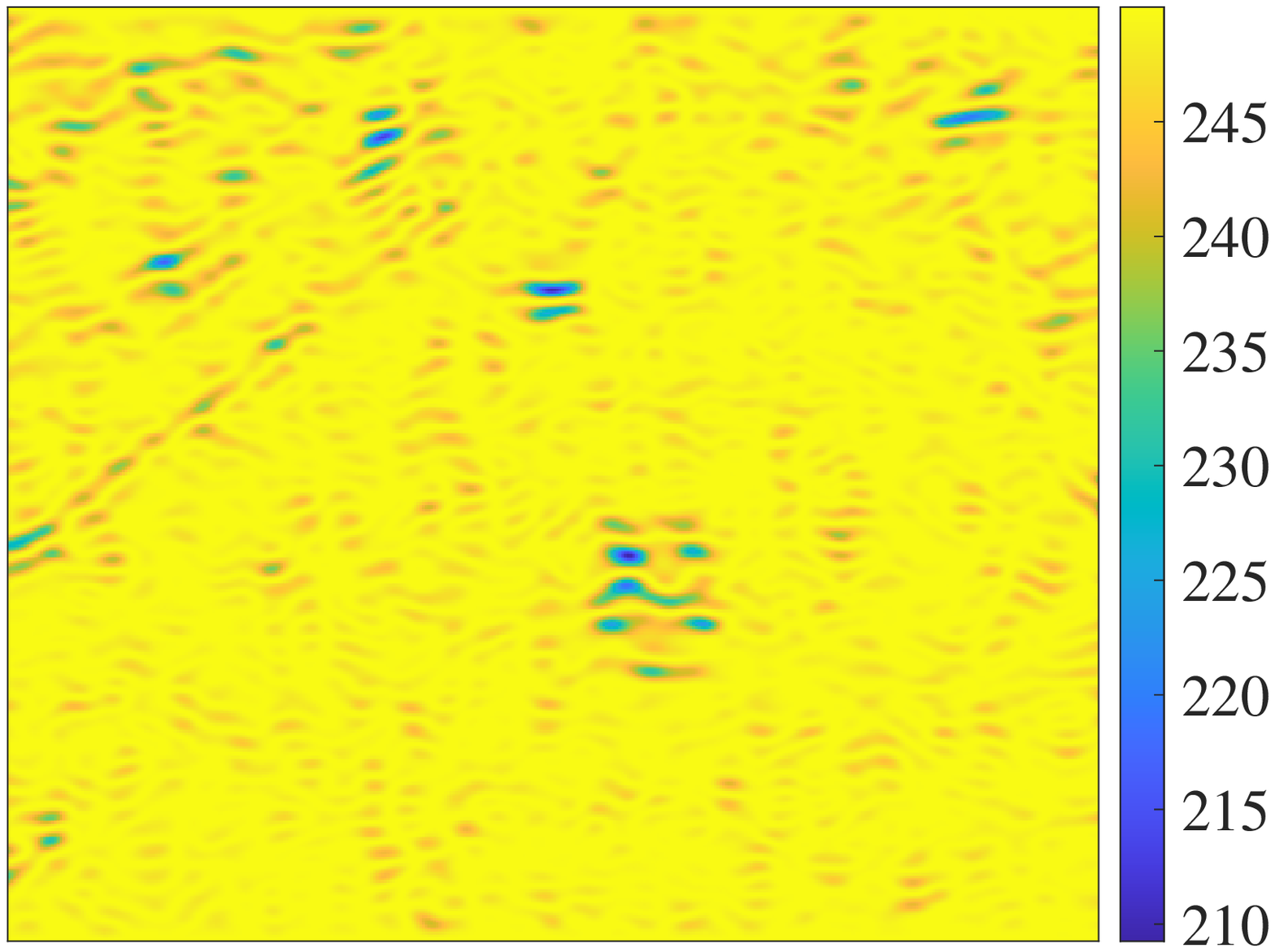}
    \caption{Vertical direction}
    \label{fig:deblur_edges_vertical}
    \end{subfigure}
    \begin{subfigure}[t]{.32\textwidth}
    \includegraphics[width=\textwidth]{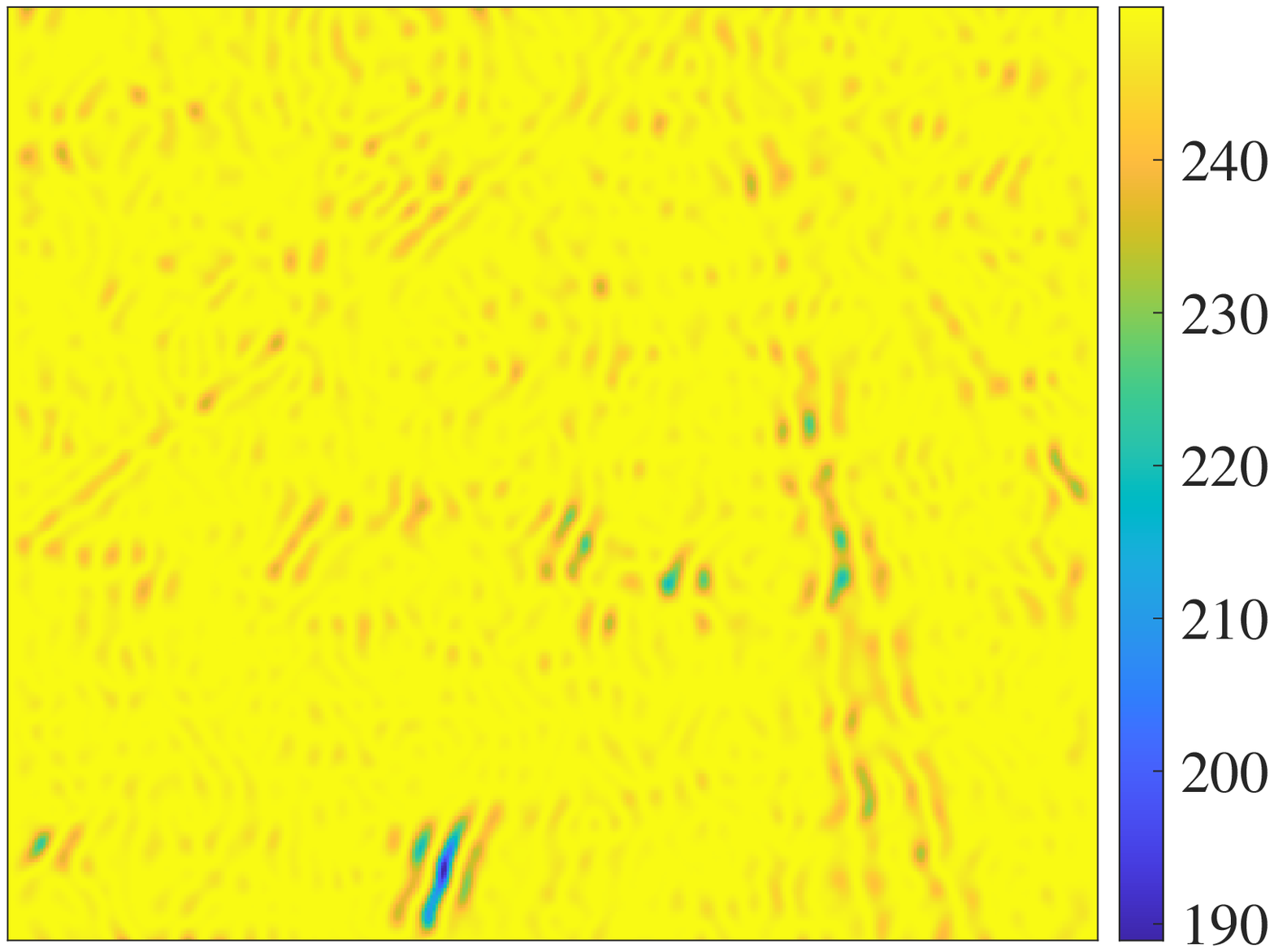}
    \caption{Horizontal direction}
    \label{fig:deblur_edges_horizontal}
    \end{subfigure}
    \begin{subfigure}[t]{.32\textwidth}
    \includegraphics[width=\textwidth]{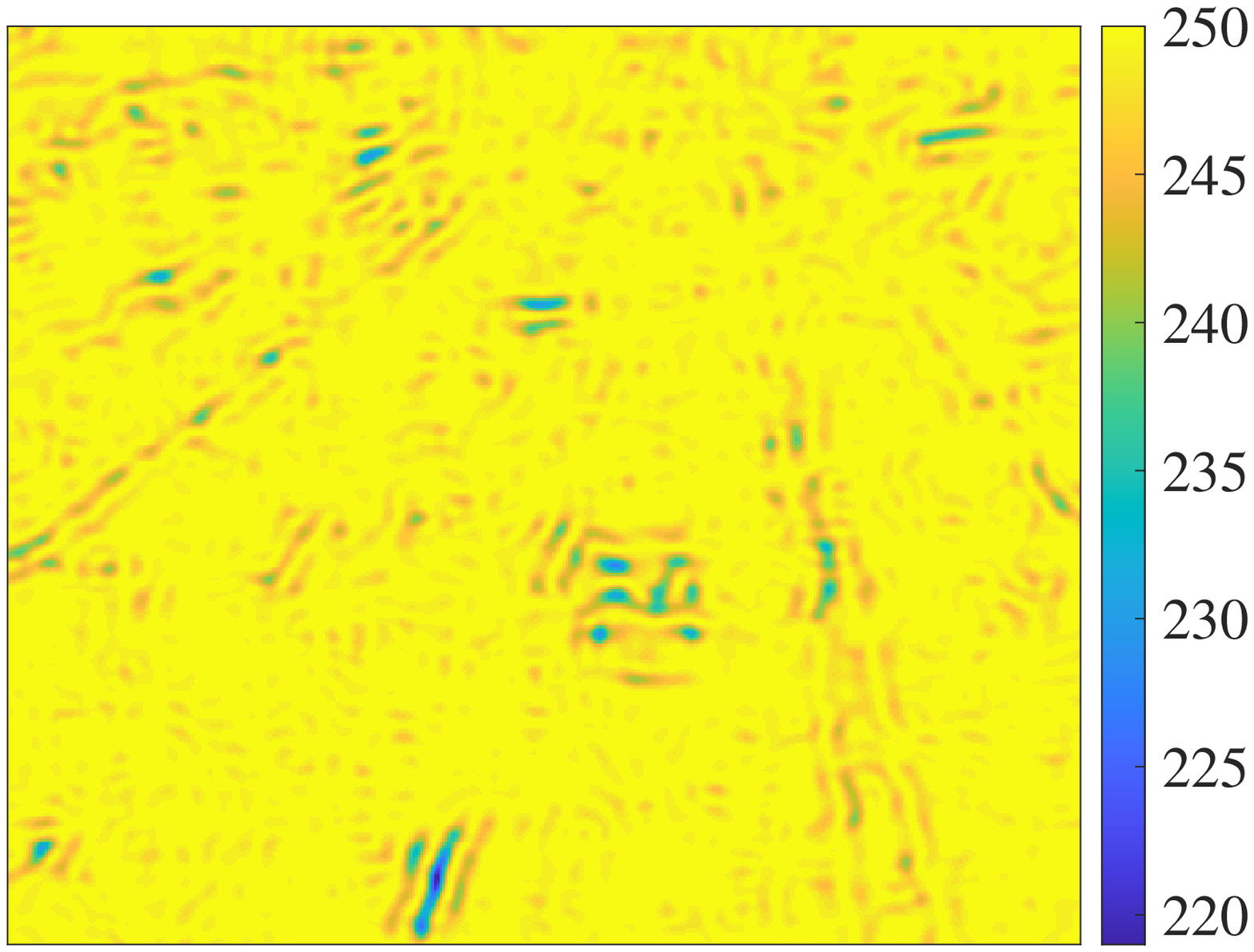}
    \caption{Pixelwise averege value}
    \label{fig:deblur_edges_combined}
    \end{subfigure}
    \caption{
    The final estimates for the intra-image regularization parameters in the vertical and
horizontal direction for the first recovered image in Figure \ref{fig:deblur_recovered} using our JHBL algorithm and their pixelwise average 
    }
    \label{fig:deblur_edges}
\end{figure}

Moreover, Figures \ref{fig:deblur_edges_vertical} and \ref{fig:deblur_edges_horizontal} illustrate the final estimate of the first and second half of the intra-image regularization parameter $\boldsymbol{\beta}^{(1)}$ of our JHBL method for the first recovered image in Figure \ref{fig:deblur_recovered_joint1}. 
Since we used an anisotropic second-order TV regularization operator \eqref{eq:R_deconvolution}, the
intra-image regularization parameter values indicate edges in the vertical and horizontal direction.
We can again combine them, e.\,g., by considering the pixelwise average of the images in Figures \ref{fig:deblur_edges_vertical} and \ref{fig:deblur_edges_horizontal} to obtain the edge profile in Figure \ref{fig:deblur_edges_combined}. 

\begin{figure}[tb]
    \centering
    \begin{subfigure}[b]{.45\textwidth}
    \includegraphics[width=\textwidth]{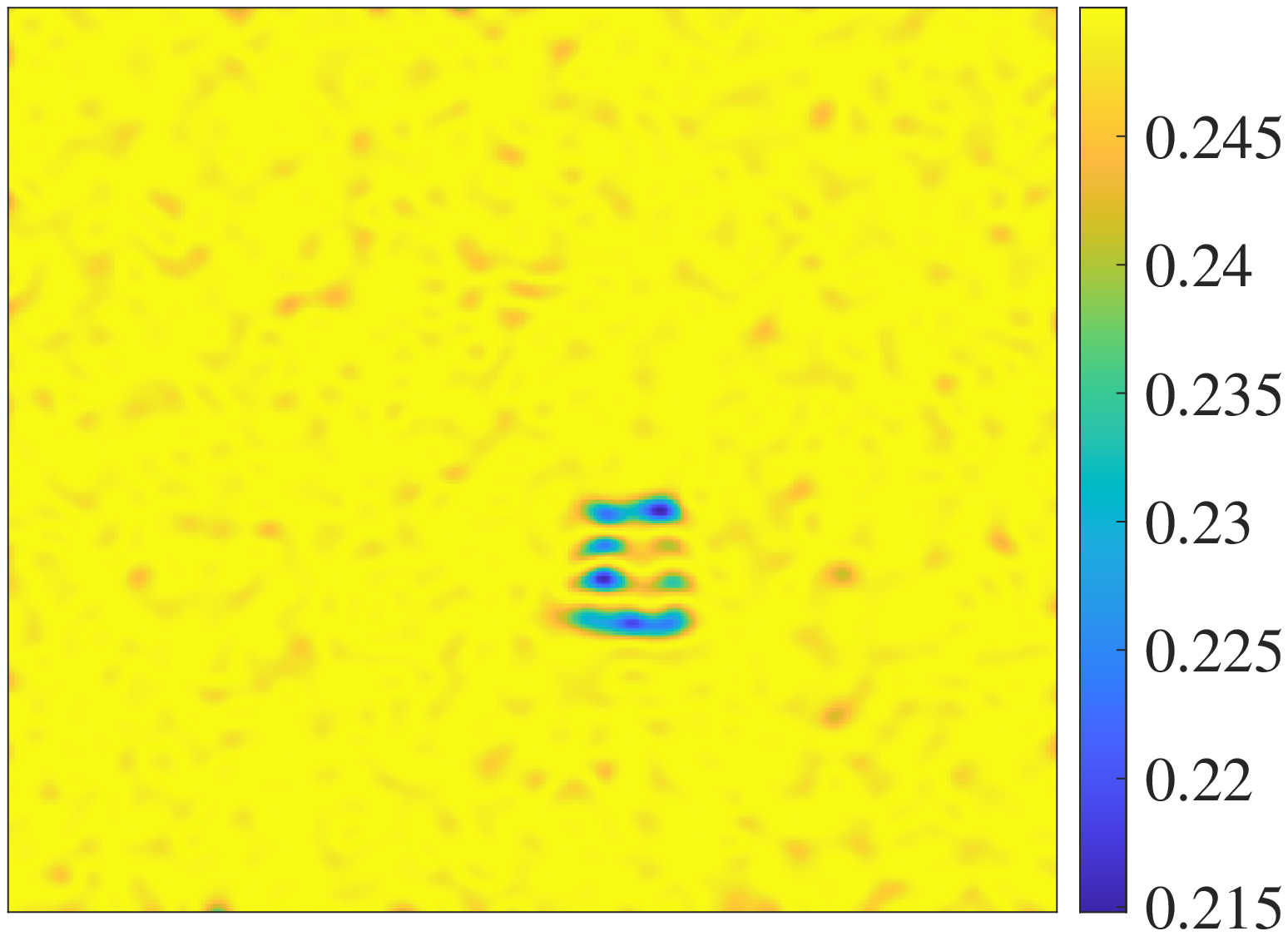}
    \caption{$1$st Bayesian change mask $C^{(1,2)}$}
    \end{subfigure}
    \begin{subfigure}[b]{.45\textwidth}
    \includegraphics[width=\textwidth]{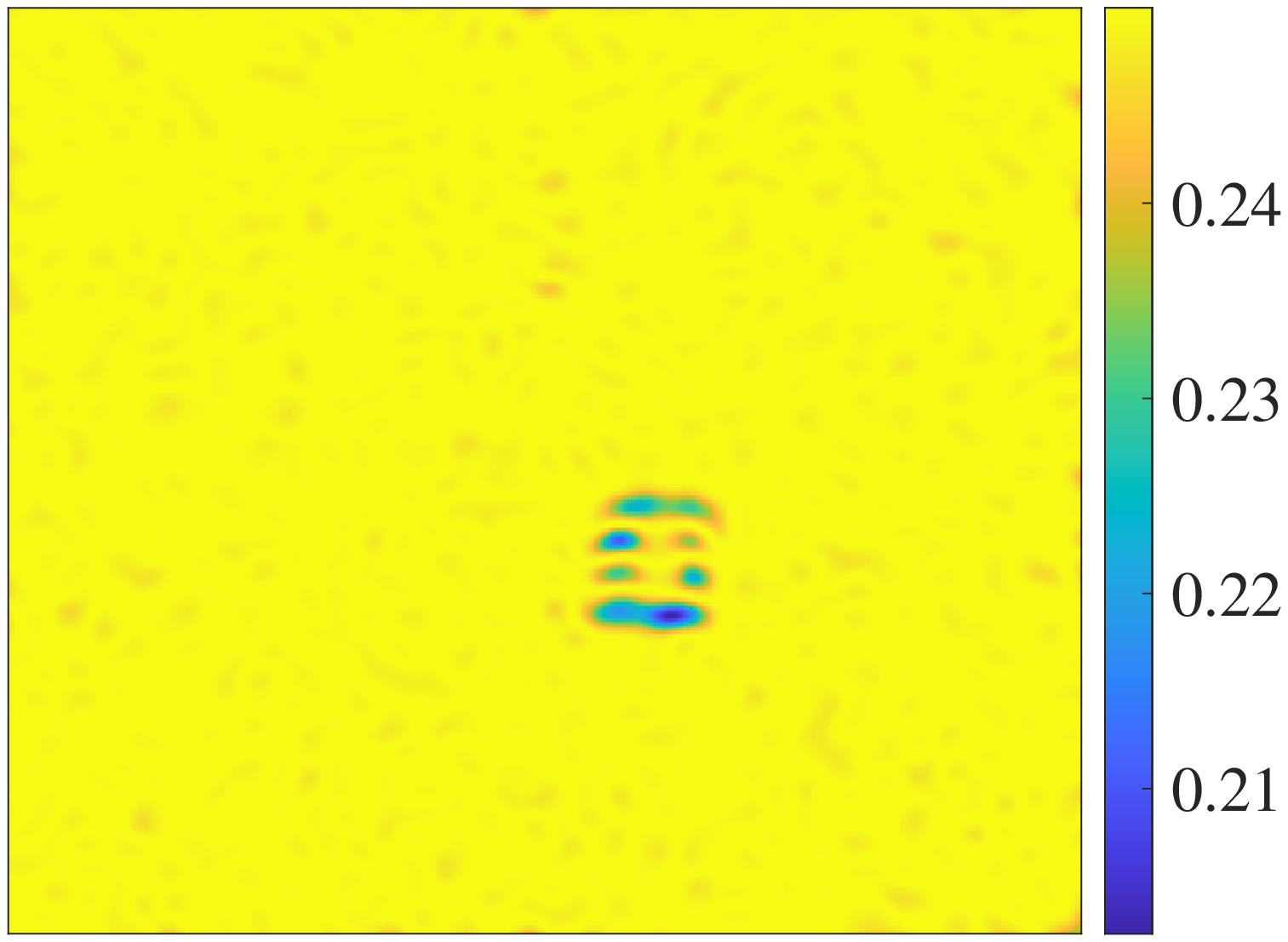}
    \caption{$2$nd Bayesian change mask $C^{(2,3)}$}
    \end{subfigure}
    \caption{
    The final estimates for the Bayesian change masks for the first and second pair of images in Figure \ref{fig:deblur_recovered} using our JHBL algorithm 
    }
    \label{fig:deblur_change}
\end{figure} 

Finally, Figure \ref{fig:deblur_change} provides the final estimates for the Bayesian change mask for the first and second pair of images in Figure \ref{fig:deblur_recovered} using our JHBL algorithm only.

\subsection{Investigating the influence of the shape and rate parameter} 
\label{sub:param_investigation} 

We end this section by briefly demonstrating how different choices for the shape and rate parameter, $\eta_{\gamma}$ and $\theta_{\gamma}$, of the inter-image hyper-prior (see \S \ref{sub:prior_inter}) influence the jointly recovered image. 
Recall (,e.g., from Remark \ref{rem:gamma}) that we expect the coupling between images to increase/decrease as smaller/larger values for the inter-image hyper-parameters $\gamma_n^{(j-1,j)}$ become more likely. 
At the same time, the expected value and variance of the inter-image hyper-prior \eqref{eq:hyperprior_gamma} are $\eta_{\gamma}/\theta_{\gamma}$ and $\eta_{\gamma}/\theta_{\gamma}^2$, respectively. 
Hence, if $\eta_{\gamma}$ is increased, we expect larger values for $\gamma_n^{(j-1,j)}$ to become more likely and the inter-image coupling to decrease. 
On the other hand, if $\theta_{\gamma}$ is increased, we expect smaller values for $\gamma_n^{(j-1,j)}$ to become more likely and the inter-image coupling to increase.

\begin{figure}[tb]
    \centering
    \begin{subfigure}[b]{.32\textwidth}
    \includegraphics[width=\textwidth]{figures/gBCD_x1_GE.eps}
    \caption{Separately recovered}
    \label{fig:MRI_parameters_eta_sep}
    \end{subfigure}
    \begin{subfigure}[b]{.32\textwidth}
    \includegraphics[width=\textwidth]{figures/cBCDs_x1_GE.eps}
    \caption{Jointly recovered}
    \label{fig:MRI_parameters_joint}
    \end{subfigure}
    \begin{subfigure}[b]{.32\textwidth}
    \includegraphics[width=\textwidth]{figures/true_x2_GE.eps}
    \caption{Reference image}
    \end{subfigure}
    \\ 
    \begin{subfigure}[b]{.32\textwidth}
    \includegraphics[width=\textwidth]{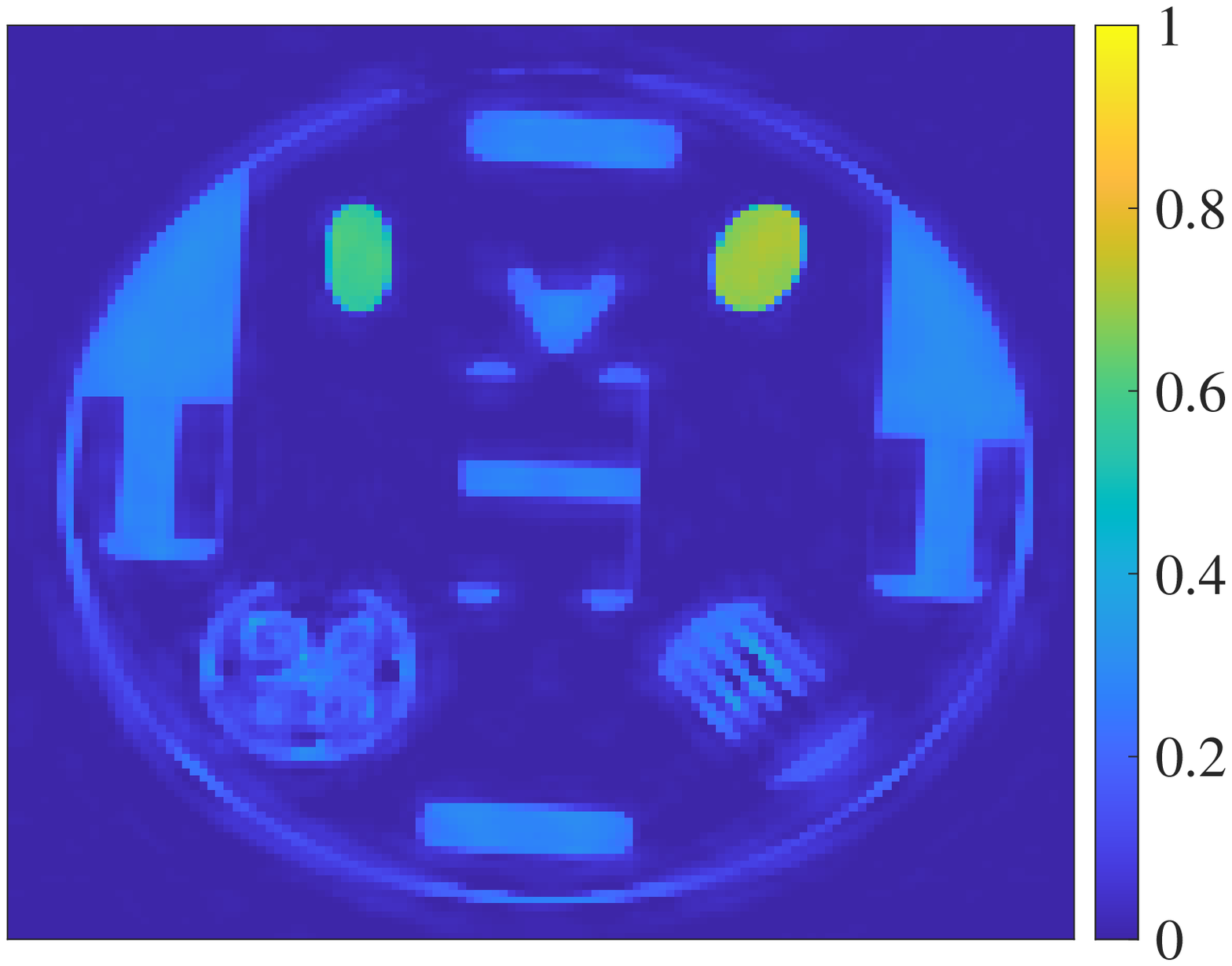}
    \caption{$\eta_{\gamma} = 0.8$}
    \label{fig:MRI_parameters_eta08}
    \end{subfigure}
    \begin{subfigure}[b]{.32\textwidth}
    \includegraphics[width=\textwidth]{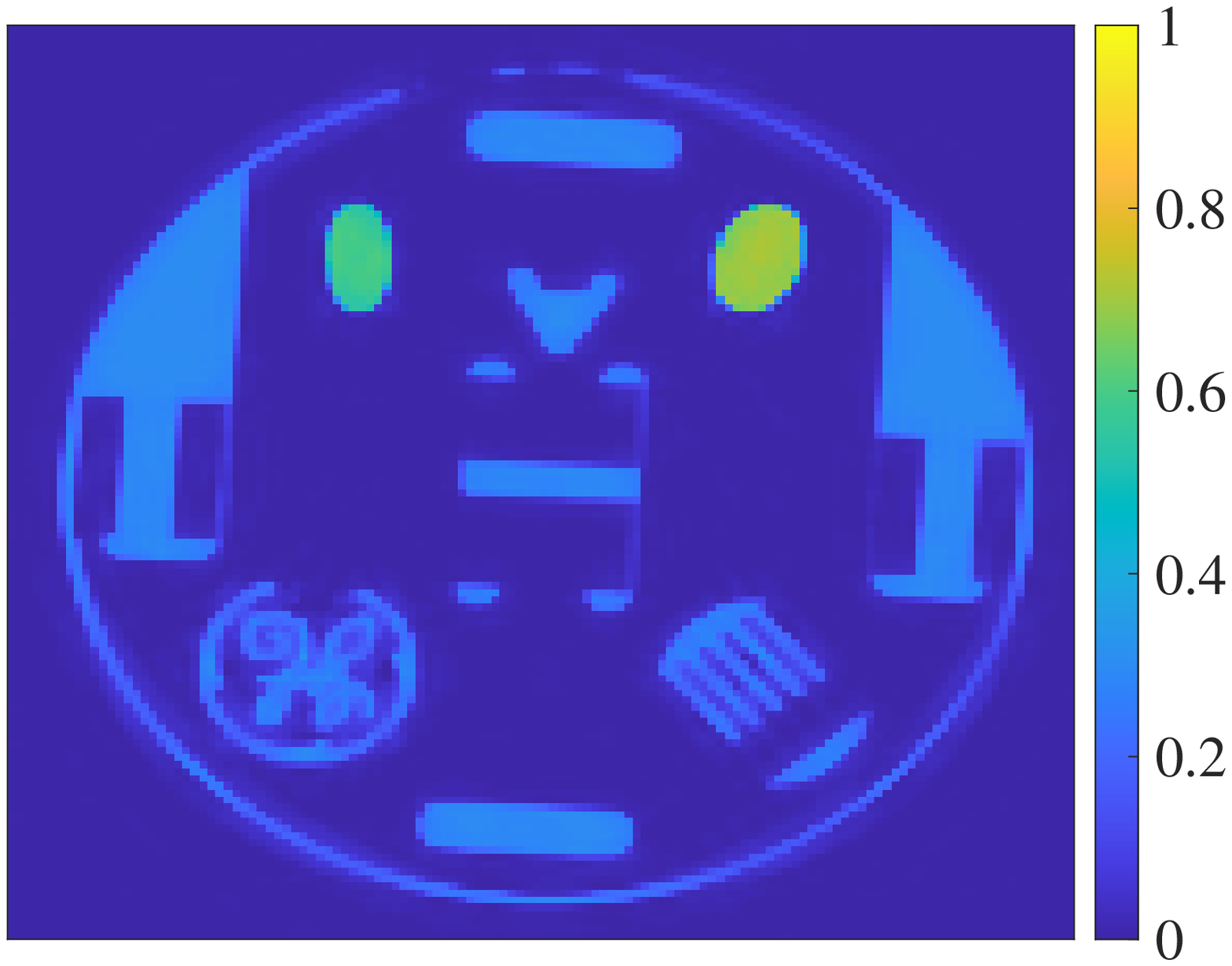}
    \caption{$\eta_{\gamma} = 1$}
    \label{fig:MRI_parameters_eta1}
    \end{subfigure}
    \begin{subfigure}[b]{.32\textwidth}
    \includegraphics[width=\textwidth]{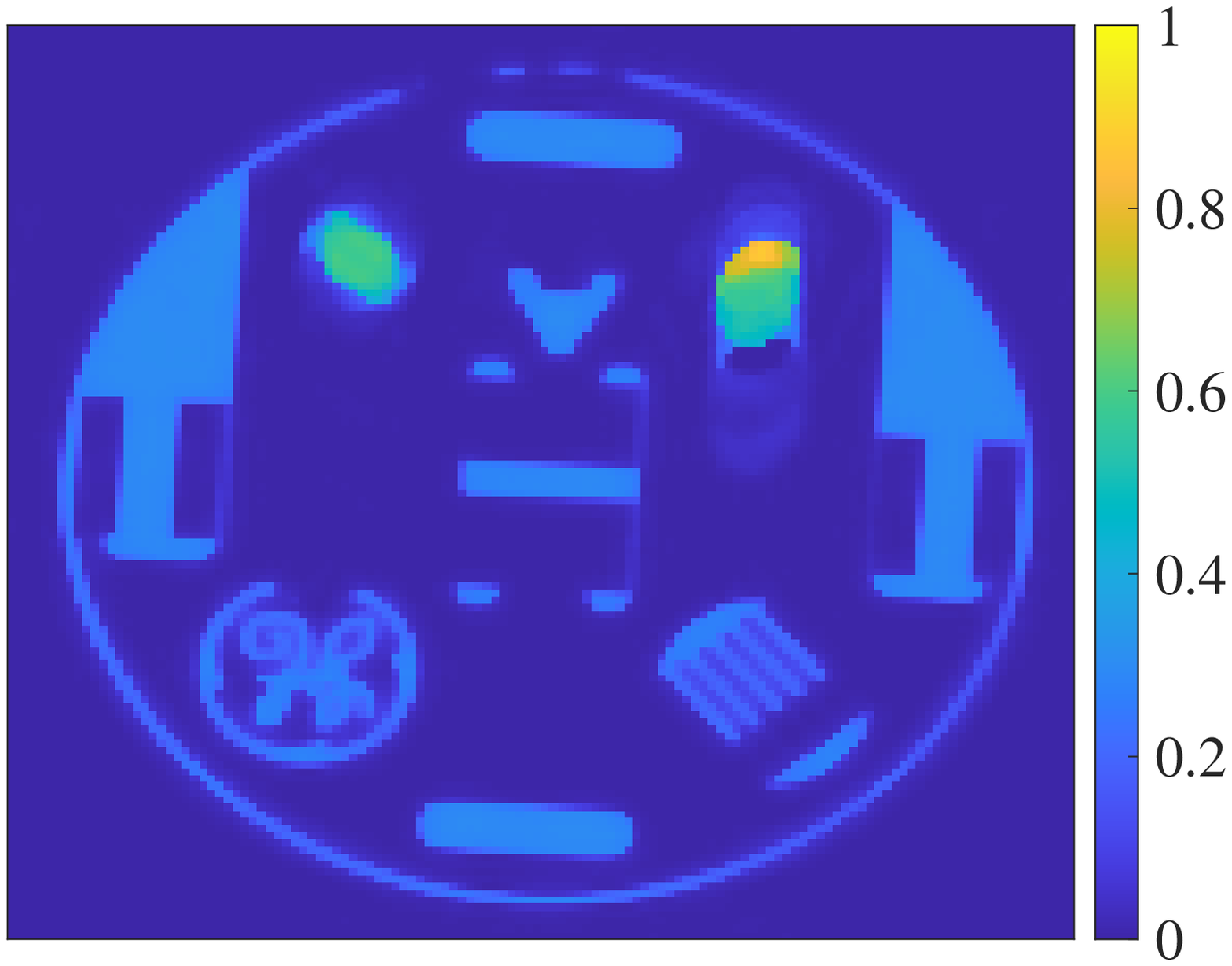}
    \caption{$\eta_{\gamma} = 4$}
    \label{fig:MRI_parameters_eta4}
    \end{subfigure}
    \\
    \begin{subfigure}[b]{.32\textwidth}
    \includegraphics[width=\textwidth]{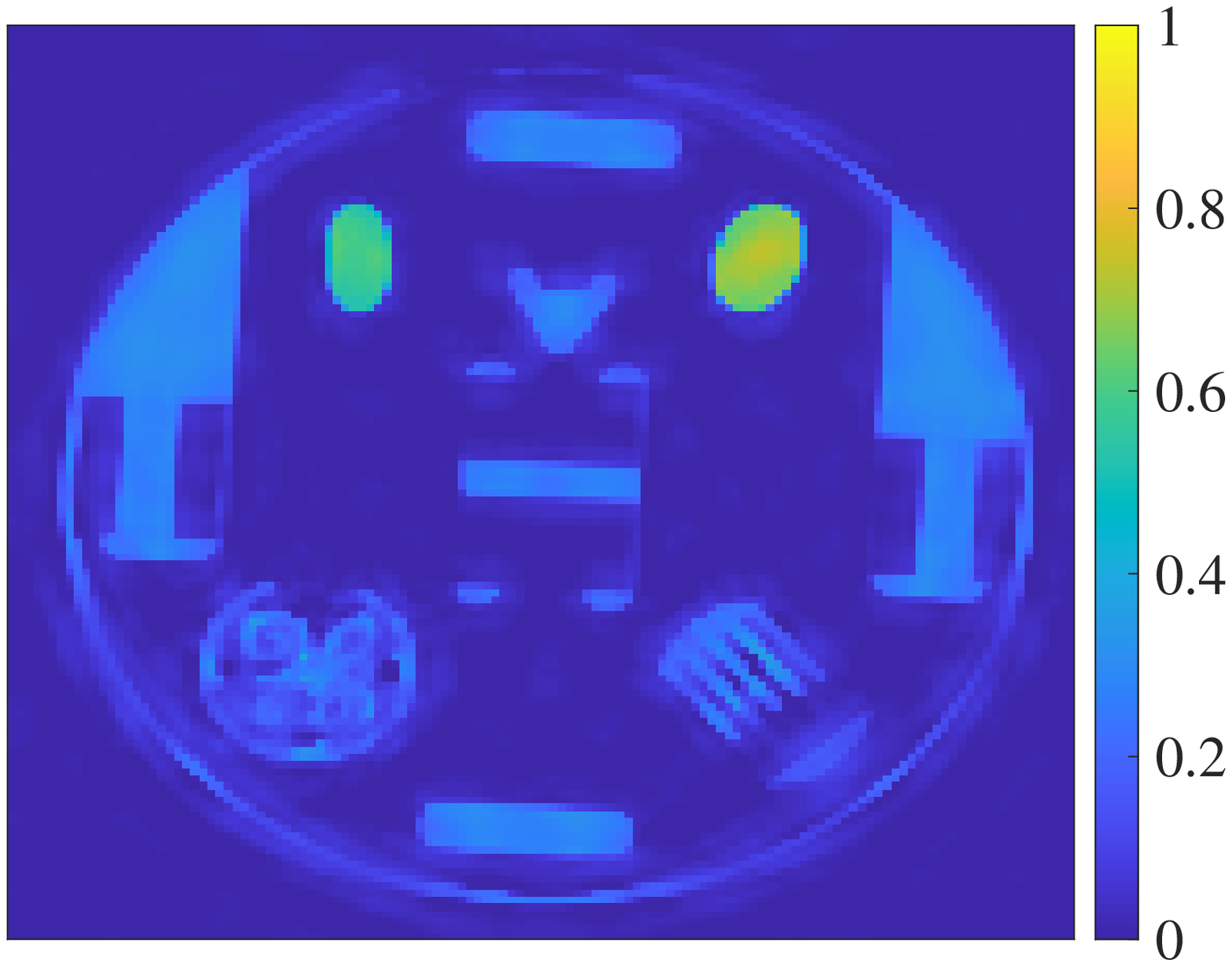}
    \caption{$\theta_{\gamma} = 10^{-1}$}
    \label{fig:MRI_parameters_theta_1}
    \end{subfigure}
    \begin{subfigure}[b]{.32\textwidth}
    \includegraphics[width=\textwidth]{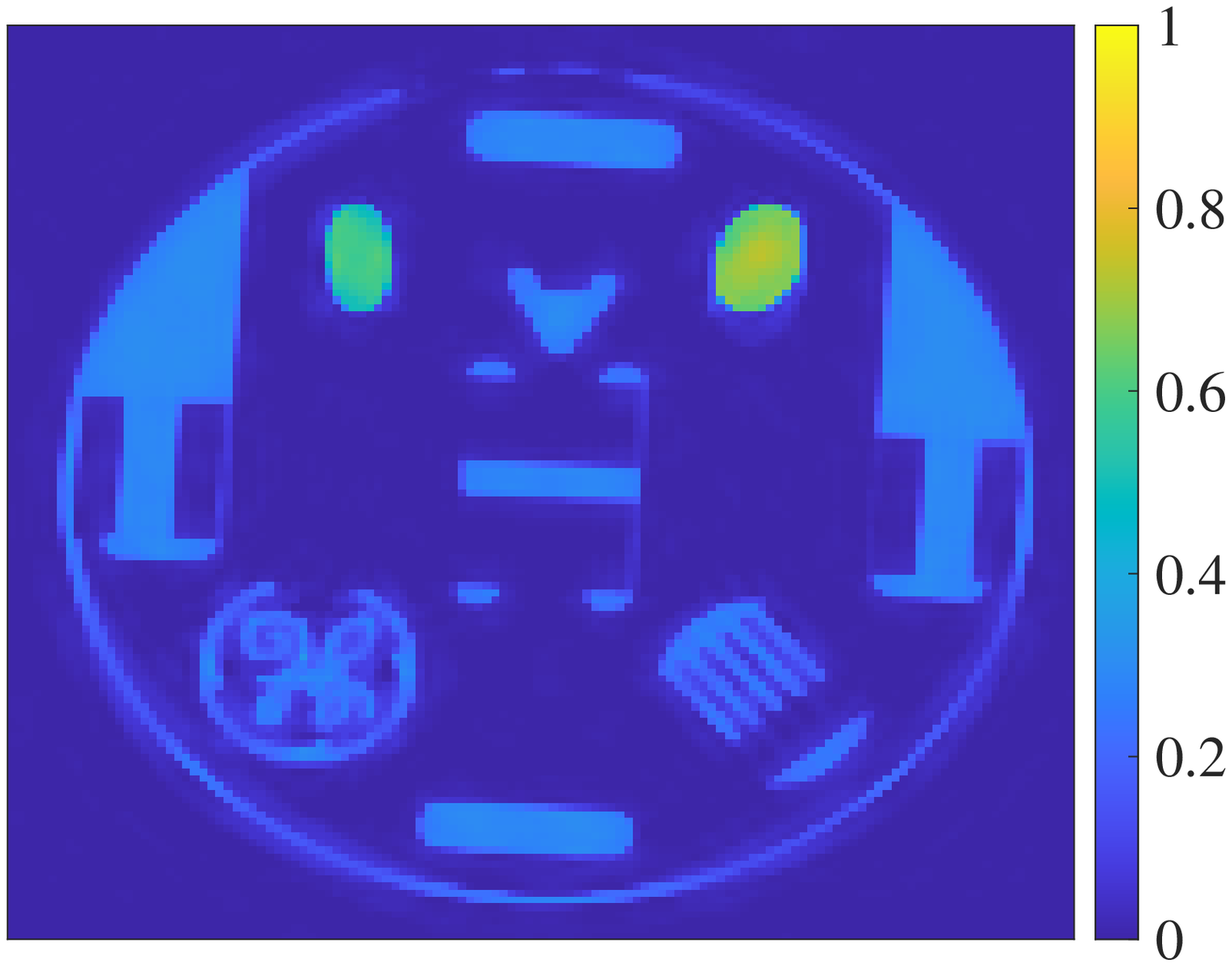}
    \caption{$\theta_{\gamma} = 10^{-2}$}
    \label{fig:MRI_parameters_theta_2}
    \end{subfigure}
    \begin{subfigure}[b]{.32\textwidth}
    \includegraphics[width=\textwidth]{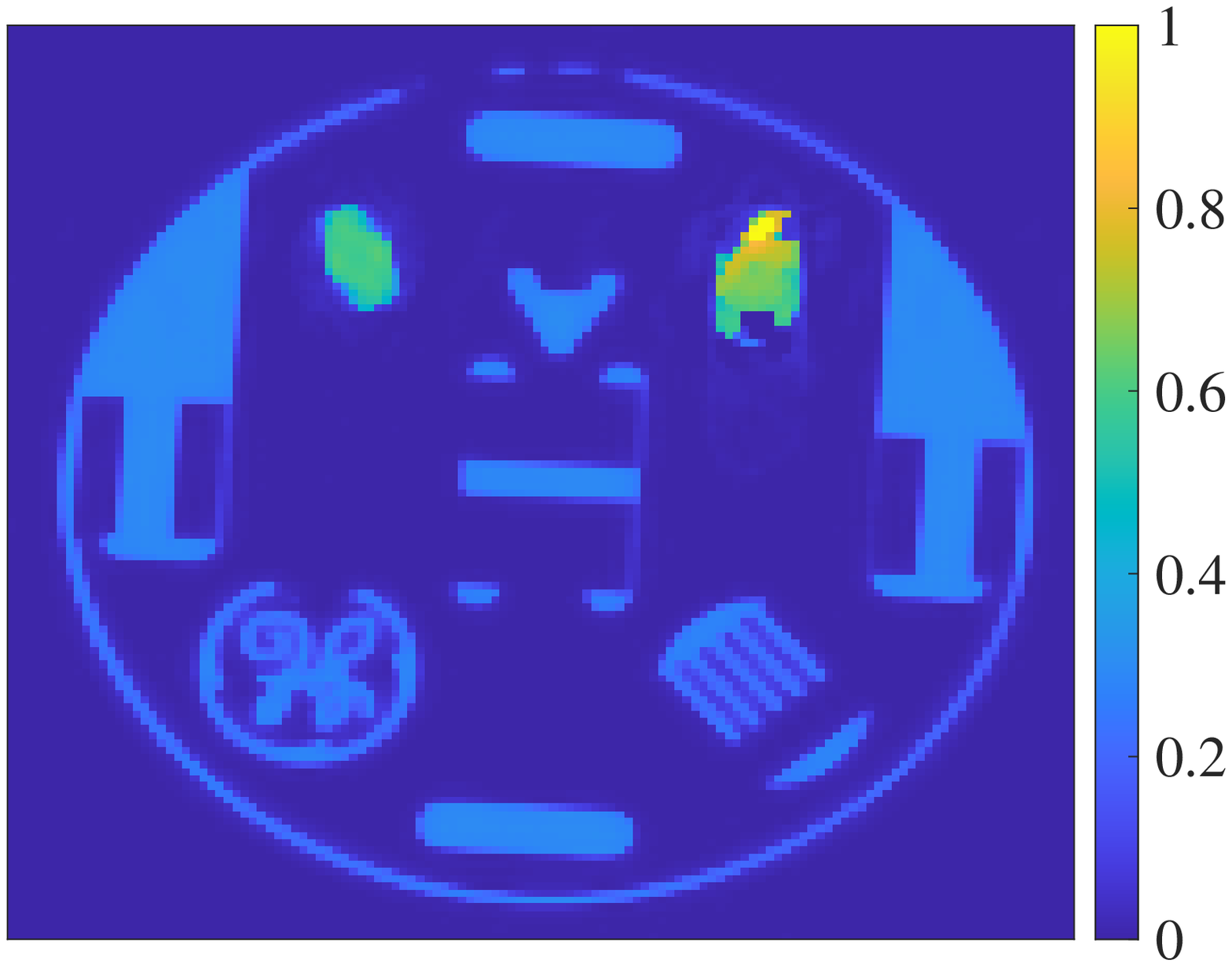}
    \caption{$\theta_{\gamma} = 10^{-3.5}$}
    \label{fig:MRI_parameters_theta_35}
    \end{subfigure}
    \caption{ 
    Demonstrating the influence of the shape and rate parameter, $\eta_{\gamma}$ and $\vartheta_{\gamma}$, on the inter-image coupling. 
    First row:  First phantom reference image, its joint reconstruction using the JHBL algorithm with $\eta_{\gamma} = 2$ and $\eta_{\gamma} = 10^{-3}$,  and its separate reconstruction using the GSBL algorithm. 
    Second row: Joint reconstructions for fixed $\eta_{\gamma} = 10^{-3}$ and varying $\eta_{\gamma}$.
    Third row: Joint reconstructions for fixed $\theta_{\gamma} = 2$ and varying $\theta_{\gamma}$.
    }
    \label{fig:MRI_parameters}
\end{figure}

Figure \ref{fig:MRI_parameters} illustrates the connection between $\eta_{\gamma}, \theta_{\gamma}$ and the strength of the inter-image coupling for the first phantom image from the sequential MRI test case previously discussed in \S \ref{sub:tests_MRI}.
Specifically, the second row of Figure \ref{fig:MRI_parameters} shows that the jointly recovered image is visibly close to the separately recovered image for $\eta_{\gamma} = 0.8$. 
At the same time, the inter-image coupling becomes so strong that some spurious artifacts from the subsequent images are introduced in change regions for $\eta_{\gamma} = 4$. 
The third row of Figure \ref{fig:MRI_parameters} demonstrates a similar behavior for fixed $\eta_{\gamma} = 2$ and decreasing $\theta_{\gamma}$; 
The jointly recovered image is visibly close to the separately recovered image for $\theta_{\gamma} = 10^{-1}$, while the inter-image coupling becomes so strong that some spurious artifacts are introduced in change regions for $\theta_{\gamma} = 10^{-3.5}$.
Future work will optimize the shape and rate parameter selection of the inter-image hyper-prior to further increase the advantage of inter-image coupling between sequential images. 
\section{Summary} 
\label{sec:summary}

We presented a new method to jointly recover temporal image sequences by ``borrowing" missing information in each image from the other images. 
Our JHBL method is simple to implement, easily parallelized, and efficient. 
We found our method to yield more accurate recovered images than both, separately recovering the images using the Bayesian GSBL algorithm and jointly recovering them using the deterministic method from \cite{xiao2022sequential}. 
In addition, our method avoids exhaustive parameter fine-tuning. 
Moreover, the deterministic method proposed in \cite{xiao2022sequential} is limited to Fourier data sets and images only containing objects with closed boundaries to pre-compute a binary change mask. 
By contrast, we treat the change mask that steers the coupling between neighboring images as a random variable, which we estimate together with the images and other parameters, making our method applicable to general modalities. 
We demonstrated this by considering a sequential image deblurring problem based on the GRAM road-traffic monitoring data set. 
Another distinct advantage of our method is that it allows us to quantify uncertainty, which is vital in applications without reference images. 
An additional valuable by-product is that our method allows for edge and change detection. 

Future work will address the selection of hyper-parameters. 
Future efforts will also investigate the structural properties of the cost function and convergence of the proposed JHBL algorithm. 
Finally, a comparison or combination of the proposed BCD algorithm for Bayesian inference with other existing methods would be of interest. 

\section*{Acknowledgements} 
We thank Guohui Song and the anonymous reviewers for helpful feedback on an earlier version of this manuscript. 
JG acknowledges support from AFOSR \#F9550-18-1-0316, ONR \#N00014-20-1-2595, and the US Department of Energy, SciDAC program, under grant DE-SC0012704.


\bibliographystyle{siamplain}
\bibliography{literature}

\end{document}